\documentclass{book}
\usepackage{graphicx} 
\usepackage[english]{babel}
\usepackage{csquotes}
\usepackage[backend=biber,style=alphabetic,maxnames=2,url=false,eprint=false]{biblatex}
\usepackage[toc,page]{appendix}
\usepackage{amsmath}
\usepackage{amsthm}
\usepackage{amsfonts}
\usepackage{booktabs}
\usepackage{arydshln}
\usepackage{algorithm}
\usepackage[noend]{algpseudocode}
\usepackage[dvipsnames]{xcolor}
\usepackage{tikz}
\usetikzlibrary{automata, arrows.meta, shapes, shapes.geometric, positioning}
\usepackage{pdfpages}

\theoremstyle{definition}
\newtheorem{definition}{Definition}
\newtheorem{theorem}{Theorem}
\newtheorem{observation}{Observation}

\newcommand{\MDP}{\mathcal{M}}
\newcommand{\fullMDP}{\MDP = (S, F, A, r, p, \gamma)}
\newcommand{\POMDP}{\mathcal{P}_\mathcal{O}}
\newcommand{\fullPOMDP}{\POMDP = (S, F, A, O, r, p, \omega, \gamma)}
\newcommand{\MDPRM}{\MDP_{\RM}}

\newcommand{\propsSet}{\mathcal{P}}
\newcommand{\RM}{\mathcal{R}_\propsSet}
\newcommand{\fullRM}{\RM = (U, u_0, T, \delta_u, \delta_r)}
\newcommand{\ab}[1]{\langle #1 \rangle} 
\newcommand{\eg}{\emph{e.g.: }} 
\newcommand{\ie}{\emph{i.e.,}\ } 
\newcommand{\iie}{\emph{i.i.e.}} 
\newcommand{\rmEdge}[1]{\langle #1 \rangle}
\newcommand{\ao}[1]{\textcolor{NavyBlue}{\textbf{#1}}}           
\newcommand{\rwP}[1]{\textcolor{Green}{\textbf{#1}}}             
\newcommand{\rwN}[1]{\textcolor{Red}{\textbf{#1}}}               
\newcommand{\I}{\ | \ }             

\makeatother
\makeatletter

\DeclareMathOperator*{\argmax}{argmax}
\newcommand{\bigO}{\mathcal{O}}
\DeclareMathOperator*{\E}{\mathbb{E}}

\algblockdefx{Local}{EndLocal}{\textbf{local variables:}}{}
\algblockdefx{Static}{EndStatic}{\textbf{static variables:}}{}

\newcommand{\hd}{\hdashline[0.5pt/5pt]}
\newcommand{\HD}{\hdashline[2.5pt/2.5pt]}

\title{Adversarial Attacks to Reward Machine-based Reinforcement Learning}
\author{Lorenzo Nodari}
\date{July 2023}

\begin{document}

\includepdf[pages={1}]{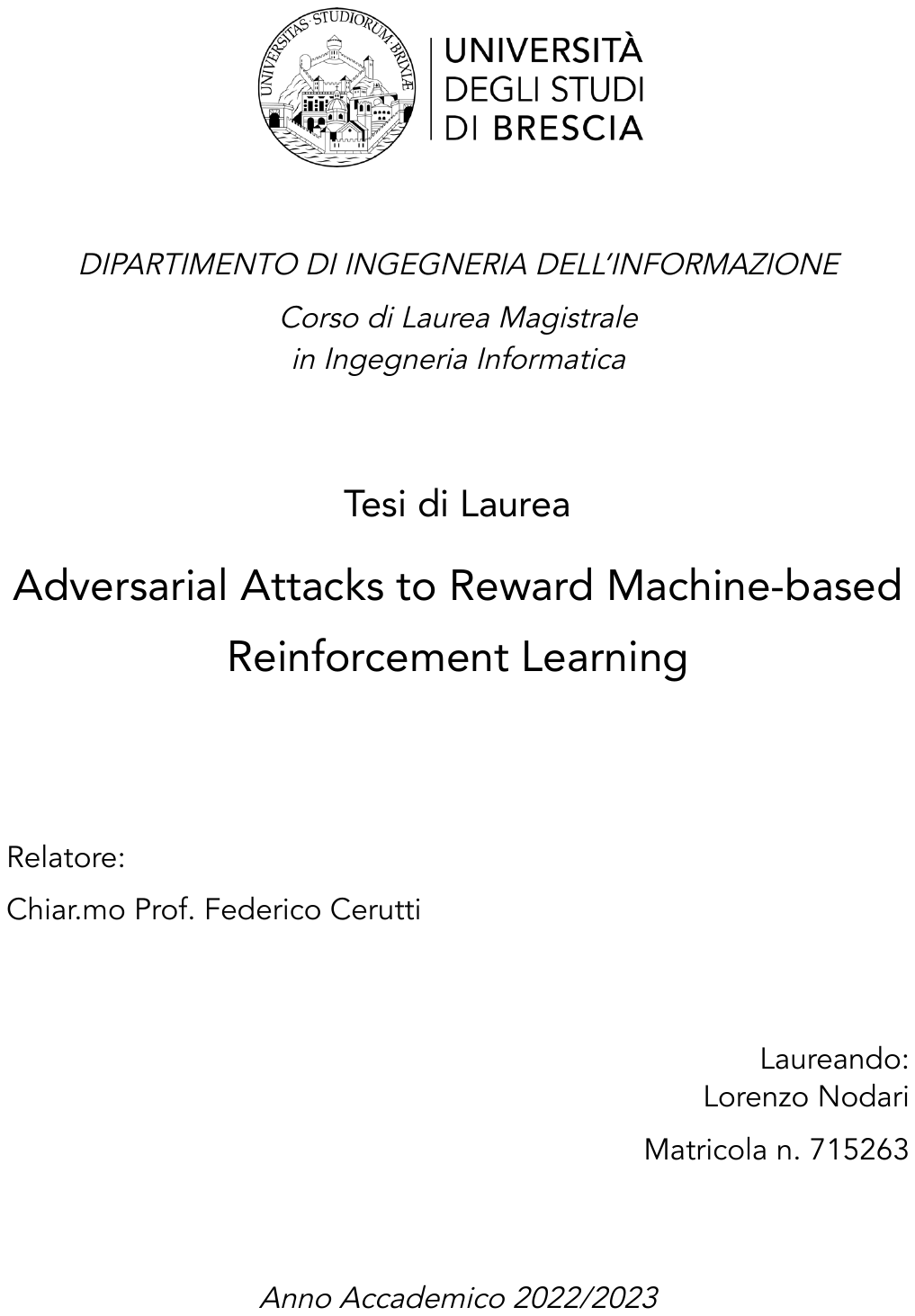}

\tableofcontents

\chapter*{Sommario}
Negli ultimi dieci anni, il campo dell'\textbf{Intelligenza Artificiale} (IA) ha subito un enorme sviluppo. Ciò è stato, ed è tuttora, possibile grazie agli ingenti investimenti nel settore da parte dei principali istituti di ricerca pubblici e privati, e dalle aziende leader del settore tecnologico. In particolare, uno dei principali fattori alla base di tale interesse verso l'IA può essere identificato nei risultati ottenuti grazie a nuove, promettenti tecniche, che stanno dimostrando un potenziale di innovazione senza precedenti in una miriade di ambiti, sia accademici che commerciali. Pertanto, in questo contesto di rapido progresso, la necessità di studiare le implicazioni, in termini di sicurezza, derivanti dall'uso di tali tecnologie è più alta che mai. Una mancanza di analisi in tal senso porterebbe infatti, negli anni a venire, ad enormi rischi di natura etica, personale ed economica e, nel caso peggiore, a danni inestimabili alla società e ai singoli individui.

Tra i vari sotto-ambiti dell'IA che stanno alimentando i sopracitati progressi, un esempio notevole è rappresentato dal \textbf{Reinforcement Learning} (RL)--- \ie apprendimento per rinforzo --- disciplina che studia i fondamenti teorici e le tecniche pratiche per la creazione di agenti artificiali in grado di apprendere autonomamente mediante la sperimentazione di nuovi comportamenti e la correzione di quelli non adeguati. Grazie a tale approccio, l'uso del reinforcement learning permette di affrontare problemi la cui soluzione risulterebbe impossibile a fronte dell'utilizzo di altre tecniche. In tale ambito, una promettente classe di tecniche correntemente studiate dalla comunità scientifica nasce dall'utilizzo delle cosiddette \textbf{Reward Machines} (RM), una rappresentazione strutturata, basata sul formalismo degli automi a stati finiti, dei segnali di rinforzo utilizzati dagli agenti per orientare il loro apprendimento. Uno dei principali vantaggi di tale approccio è la possibilità di migliorare le prestazioni degli agenti addestrati in ambienti altamente complessi e caratterizzati da parziale osservabilità. Nonostante la loro rilevanza, ad oggi, e al meglio della mia conoscenza, nella letteratura in materia non è mai stata proposta alcuna analisi delle implicazioni, in termini di sicurezza, derivanti dall'utilizzo di tali tecniche.

Alla luce di ciò, la mia tesi mira ad essere una prima analisi sulla sicurezza degli approcci di reiforcement elarning basati su reward machines, con la speranza di fungere da base per lo sviluppo di ulteriori studi in materia. Partendo da un'analisi teorica delle implicazioni di sicurezza derivanti dal loro utilizzo, questa tesi propone una nuova classe di attacchi mirati a degradare le prestazioni di agenti ottimi e discute un ampio sprettro di variazioni sull'approccio alla loro base. Infine, il mio lavoro si conclude con una presentazione dei risutati ottenuti mediante la valutazione sperimentale delle tecniche proposte in tre diversi ambienti di test parzialmente osservabili.
\chapter{Introduction}

\textbf{Reinforcement Learning (RL)} is a sub-field of artificial intelligence (AI) that deals with the training of autonomous agents that learn while being guided by a reward signal obtained using iterative interaction with their environment. In formal terms, the object of study of reinforcement learning is the optimal solution of Markov Decision Processes (MDPs) and their generalizations. Following this scheme, an RL agent aims at learning the best \emph{policy} for completing its task: a mapping from environment states to actions to be taken.

In the last decade, following the introduction and refinement of deep-learning-based algorithms designed explicitly for RL, the field has obtained astonishing successes in a variety of highly-complex tasks, with robotics \cite{li_reinforcement_2021}, game playing \cite{vinyals_grandmaster_2019}, and conversational AI\footnote{See \url{https://openai.com/blog/chatgpt}. URL visited on 26/07/2023} being just a few notable examples demonstrating the breadth of its impact on today's artificial intelligence research. Thus, building on the advances above, recent years' work has focused on solving some of the issues that still limit RL potential for many real-world applications, such as its low sample efficiency, the need for high-quality memory architectures to handle partial observability, the difficulty associated with the representation of highly-structured tasks and rewards, and the poor explainability of learned policies.

\textbf{Neuro-symbolic Reinforcement Learning} is an emerging paradigm that can tackle some of these problems by combining deep learning with symbolic artificial intelligence techniques such as logics-based knowledge representation and inference. Following this trend, \textbf{Reward Machines (RMs)} have stood out as a simple yet effective automata-based formalism for exposing and exploiting underlying task structure during training, allowing for improved sample efficiency and demonstrating the ability to converge in partially observable benchmark environments where state-of-the-art A3C, ACER and PPO algorithms equipped with LSTM-based memory could not make any progress. These results sparked the interest of the research community, leading to a high number of works refining the approach in various ways for both single-agent and multi-agent settings, \eg \cite{furelos-blanco_hierarchies_2022}, \cite{neary_reward_2021}, \cite{li_noisy_2022}, \cite{dohmen_inferring_2022}, among others.

Despite their relevance, \textbf{little to no attention has been directed to the study of the security and robustness of neuro-symbolic approaches}, likely due to their recent appearance in the literature, with most research efforts being spent on attack techniques targeting traditional deep-learning components in RL systems, such as policy networks \cite{pan_characterizing_2022}, \cite{ilahi_challenges_2022}. While, in theory, some of the known attacks on deep reinforcement learning could also apply to neuro-symbolic systems, the lack of studies addressing the security issues specifically raised by the introduction of symbolic components makes their investigation a fundamental requirement for the safe deployment of such techniques in real-world scenarios.

This issue becomes particularly concerning in light of the fact that, with the recent advent of neuro-symbolic techniques, the relevance of reinforcement learning algorithms shows the potential for further increase, due to the vast possibilities opened by the combination of deep-learning techniques for low-level perception and actuation, and symbolic components for high-level action planning and knowledge representation. Furthermore, during the last decade, we witnessed the fast-paced adoption\footnote{See \url{https://www.ibm.com/watson/resources/ai-adoption} and \url{https://www.ibm.com/watson/resources/ai-adoption} for reference. URLs visited on 27/07/2023.} of artificial intelligence techniques in a plethora of both academic and industrial research fields, enterprise applications and consumer-ready services. Consequently, the number of potential risks arising from the use of AI-powered solutions has vastly increased as well, thus determining a greater need for the evaluation of their robustness.

The autonomous driving technology found in Tesla vehicles exemplifies the incredible possibilities opened by the real-world adoption of artificial intelligence and, at the same time, their huge implications from a security point of view. However, while Tesla's autopilot, alongside recently introduced generative AIs like ChatGPT and DALL-E, represent some of the most notable examples of consumer-ready AI that sparked the public discussion about their trustworthiness and safety, the scientific literature yields a plethora of other lesser-known examples that likewise call for a deeper investigation of these fundamental requirements. Drug discovery and development \cite{popova_rl_drug_2018}, industrial cooling systems optimization \cite{wong2022cooling}, brain-computer interfaces \cite{musk_integrated_2019}, and automated cloud workload scaling \cite{ibm_rl_worload_scaling} are nothing but a few of the numerous examples of applications where RL-powered technologies, in all likelihood, will begin to be adopted more and more frequently in years to come by both consumer markets and enterprise solutions.

Altogether, while it is undeniable that such applications could lead to huge benefits from a myriad of perspectives, such gains will only be possible if we are able to properly address the critical issues that currently limit the trustworthiness of AI-powered solutions, the main being the explainability of model outputs and their robustness to unforeseen circumstances, such as noisy inputs or adversarial attacks. Moreover, in response to the rising availability of AI-based products, laws and regulations regarding their use and associated responsibility are starting to emerge as well, with relevant corpora being under work in both European  and American institutions, among others. Thus, as regulations evolve and consolidate, the research on the security of artificial intelligence will also have an increasing value from a legal compliance point of view, further increasing its relevance and urgency.

Motivated by all the above considerations, with my thesis, I aim to provide the \textbf{first analysis of the security of RM-based reinforcement learning techniques}, with the hope of motivating further research in the field. Starting from a theoretical analysis of the security implications arising from the use of reward machines, I propose a novel class of attacks aimed at hindering the performance of optimally trained RM-based agents: \textbf{blinding attacks}. Then, after discussing a wide range of variations over their underlying basic approach, I conclude by presenting an analysis of the results obtained by the experimental evaluation of the proposed techniques over three partially observable benchmark domains.
\chapter{Background}

\section{Artificial Intelligence}
Artificial Intelligence (AI) is a vast field of computer science research that aims at the development of \textbf{intelligent agents}. While proposing a proper, formal definition of such a concept could lead to a breadth of interesting --- often philosophical --- discussion relating to topics such as \emph{intelligence}, \emph{agency}, and \emph{autonomy}, this is beyond the scope of my thesis. Instead, I will adopt a simpler, practical definition of AI, adapted from the one provided by \citeauthor{russel2010} in the preface of \cite{russel2010}: 
\begin{displayquote}
We define AI as the study of [artificial] agents that receive percepts from the environment and perform actions [to achieve a task].
\end{displayquote}
This definition works well in the context of my work as it highlights all the key concepts that will be fundamental in understanding subsequent topics while also being sufficiently general to showcase to the reader the potential applicability of AI to a myriad of scenarios, well beyond those that will be analyzed in the remainder of my thesis.

However, since this degree of generality inevitably induces a large amount of uncertainty as to the exact scope of the definition, the following section clarifies the three foundational concepts introduced above: \textbf{agents}, \textbf{environments}, and \textbf{tasks}.

\subsection{Tasks, agents, and environments}
Among the three, the concept of a \textbf{task} is the easiest to comprehend, as its meaning does not differ from the one suggested by common sense: it simply represents what the agent is required to do, be it a one-time objective or a recurrent activity. As such, the definition of the task is usually the first aspect that is fixed during the development of an AI system, as it strongly influences both the agent design and the definition of the environment. Another aspect that strongly relates to the task at hand is the \textbf{performance measure} that is used to quantify how good --- or bad --- the agent is. 

An \textbf{agent} is the final desired output of any AI development effort. Its design is the primary object of study of the discipline and, in most practical contexts of interest, the main technical challenge. When looking for an adequate answer to the question of what an agent is, \citeauthor{russel2010} again provide a simple, elegant definition:
\begin{displayquote}
    An agent is anything that can be viewed as perceiving its environment through sensors and acting upon that environment through actuators.
\end{displayquote}
While, at first glance, the above definition might not seem to provide a clear answer, its merit lies in its generality. For an agent to be considered as such, it only needs to be able to interact with its surrounding environment by acquiring and emitting signals that are meaningful in the context of its task of interest. Thus, this condition allows for the characterization of agents of arbitrary complexity: under this definition, a simple text-suggestion system --- such as the ones available on most commercially available smartphone keyboards --- is no less of an agent than a complex humanoid-like robot capable of object identification and fluent movement in the three-dimensional space. Indeed, just like the former might be able to perceive its --- purely software-in-nature --- environment by having access to the text typed so far and to act on it by outputting its suggestion, the latter might be able to obtain information from the physical world via a complex array of sensors and to manipulate it, for instance, via mechanical arms and legs. From a high-level point of view, none of the above examples is any less of an agent than the other, as both fully comply with the above definition.

An \textbf{environment} is the context in which an agent operates to achieve its task. Contrary to what the common sense interpretation of the word might suggest, environments are not restricted to portions of the physical world: they can also be purely virtual. For example, we could imagine a surveillance robot whose task is to patrol the perimeter of a building during nighttime. In addition to the sensors needed to sense its --- physical --- surroundings, the robot might also be designed to receive a live stream of the security cameras placed around the building, thus allowing it to sense its \emph{virtual} surroundings.

This last example allows us to highlight an essential aspect of the relationship between tasks, agents, and environments: what is considered part of the environment highly depends on the signals the agent needs to achieve its mission and the ones available to him via its sensors.

\subsection{Agent functions and programs}
Mathematically, any agent's behavior can be fully described through an \textbf{agent function}:
\begin{equation}
    f : P^* \rightarrow A
\end{equation}
where $P$ is the set of all possible percepts the agent can receive from the environment, $A$ is the set of agent actions, and $P^*$ is the set of all possible percept sequences $(p_0, p_1, ...)$, with $p_i \in P \ \forall i \in \mathbb{N}$. However, while this representation represents a useful conceptual base for the theoretical design of an agent, it does not help us in the actual implementation of the agent, which necessarily needs to be embodied by an \textbf{agent program} running on some physical computing infrastructure.

When considering an agent function, it is convenient to represent it as a \emph{table} containing a row for each possible percept sequence and a column containing the agent's associated action. However, while conceptually convenient, this representation is impossible to adopt in practice. Since, in theory, there is not an \textit{a priori} upper limit on the length of a percept sequence, such a table would require an infinite amount of memory to be stored. While this problem could be fixed by introducing a finite limit on the length of the percept sequences we are willing to consider, we are immediately faced with another one: even game environments, like chess, that can be described by a small set of rules might be characterized by state spaces with a cardinality that \emph{vastly} surpasses the number of atoms in the universe\footnote{In the specific case of chess, we are faced with a number of unique possible games in the order of $10^120$. On the other side, the number of atoms in the universe is currently estimated to be in the order of $10^80$.}. Thus, when aiming to address even more complex tasks, any hope for a simple one-fits-all tabular design for an agent program quickly vanishes.

In light of this, most of the research efforts that are put in the field of artificial intelligence are directed to the development of smart solutions for the design of agent programs, leading to a rich diversification in both general and special purpose techniques and algorithms that allow us to effectively and efficiently deal with complex environments and tasks. In recent years, the field of \textbf{neuro-symbolic AI} has emerged as a new promising paradigm relying on the use of both symbolic techniques, which are based on the use of high-level, human-understandable languages for the representation of the problem at hand, and neural-network-based ones, which instead rely on immense networks of opaque, numerical parameters. Due to this peculiar

\subsection{Characterization of task environments}
Due to the vastness of the possible environments that might arise in AI applications, it is helpful to define a few axes along which they can be characterized, as different techniques might work best or even be applicable only when faced with given kinds of environments.

The first distinction depends on whether the agent has access, via its sensors, to the \emph{complete} state of the environment at each point in time. When this is true, the environment is said to be \textbf{fully observable}. This represents the simplest case in terms of required agent design: as every percept contains all the relevant information the agent needs, it does not need to keep track of the state of the world. On the other hand, when this condition does not hold, the environment is said to be \textbf{partially observable}. In particular, this latter case is very prominent in real-world environments, where partial observability might arise due to imperfect or noisy sensors or plain unavailability --- for instance, due to resource constraints or actual physical impossibility --- of adequate sensors.

The second distinction arises between \textbf{deterministic} and \textbf{stochastic} environments: in the former case, the next state of the environment is entirely determined by the current state and the agent's actions, while in the latter, the state transitions are probabilistic. Unsurprisingly, stochastic environments present additional challenges, as they introduce \emph{uncertainty} in the agent's trajectories through the environment state space, which, in most cases, must be accounted for to achieve optimal behavior. Finally, it is interesting to note how a deterministic environment might \emph{appear} to be stochastic due to it being partially observable. If the agent has no access to the information that would be required to determine the following environment state, to its eyes, the state transitions would appear non-deterministic.

If we shift our focus from the actual environment transition dynamics to the agent's knowledge of them and the possible states, we can further distinguish between \textbf{known} and \textbf{unknown} environments. In the first case, the agent knows all the possible outcomes of his actions, while in the second, in layman's terms, he does not know what to expect. Note that this characterization is entirely orthogonal to the distinction between fully and partially observable, where the focus is on the availability of relevant information during the agent's course of action. A solitary card game is an excellent example of a partially observable yet known environment. While the game's rules are well-known by the agent, the actual state of the deck of cards is unavailable during the game. On the other side, using a computer program for the first time is a good proxy example for a deterministic environment that is also unknown: every action of the user has a well-defined, deterministic effect, but the user himself needs to figure out what he can do, how can he do it, and so on. This last example highlights one fundamental requirement most often posed by unknown environments when trying to act optimally: the ability to \emph{explore} and \emph{learn}.

Regarding the temporal dependency between current and future actions, environments can be divided into \textbf{episodic} and \textbf{sequential} ones. In episodic environments, at every time step, the agent obtains a percept from the environment and decides what to do only based on the information it contains, as current actions have no impact on future decisions. Every sense-decide-act iteration is self-contained and independent from future ones. A perfect example of such environments can be found in object recognition tasks: the agent is presented with a picture, outputs its predicted label, and goes on to the next one. Conversely, sequential environments are characterized by long-term consequences arising from present actions.

\section{Reinforcement Learning}
\label{sec:rl}
Reinforcement Learning (RL) is a sub-field of artificial intelligence that deals with finding optimal strategies for solving sequential decision-making tasks through agents that learn via \emph{trial-and-error}. Taking direct inspiration from the idea of \textbf{operant conditioning} --- widely studied in the field of \emph{behavioral psychology} --- RL agents are trained by iteratively acting in their environment under the guidance of a \textbf{reward signal}: after observing the current state of the world, the agents choose the action to be taken and --- after executing it --- receive feedback that quantifies the quality of their course of action, given the circumstances. The agents then use this feedback to tune their current \textbf{policy}, and the cycle repeats. When the agent's behavior leads to a desirable outcome, a positive reward encourages it to replicate such or similar actions, whereas, in the opposite case, a negative reward aims at driving the agent away from its current strategy. Regardless of their level of sophistication, all RL algorithms share this same basic structure, consisting of a series of \emph{observation-action-reward} loops.

This framework starkly contrasts with the \emph{learn-by-example} one adopted by Supervised Learning (SL) techniques, which require the agents to access pre-existing, labeled datasets as the knowledge source guiding the learning process. While for some tasks, like video classification or time series prediction, satisfying this requirement might be feasible, in others, it could be practically impossible for several reasons. For example, let us consider flying a helicopter in a straight line from point A to point B. The first problem we would encounter if we decided to resort to SL techniques to train such an agent is the number of possible states that need to be accounted for. Even considering a minimal configuration of sensors consisting of three accelerometers, one for each axis of the 3D space, three sensors for assessing the current velocity of the helicopter's rotors, and a simple relative location sensor, we are immediately faced with a complex \emph{continuous} state space that would need to be exhaustively represented in our training dataset to ensure the reliable operation of our helicopter agent. 

Even if we assumed the feasibility of collecting such a dataset, we face a second problem: the need for \emph{labeled} data. Indeed, every state configuration would need to be associated with the correct action --- \ie how much power to direct to each rotor --- that the agent should take in that situation. Unfortunately, even the best pilots in the world would struggle if asked to provide such information accurately, let alone have the time to review every single data point in our training set. In light of these considerations, the use of supervised learning for solving such a task appears doomed to failure\footnote{In reality, there exists a sub-field of supervised learning that allows for the solution of such problems by relying on \emph{expert demonstrations}, namely Imitation Learning. This, however --- while extremely interesting --- is out of the scope of my thesis.}. 

On the other hand, the reinforcement learning paradigm allows us to overcome said limitations by eliminating the need for a labeled dataset. To solve such a task, an RL agent would only require us to provide it with a \textbf{reward function} that accurately captures the desirability, or lack thereof, of a given scenario. Doing so eliminates the need for a precise specification of the correct actions to be taken, and we leave our agent responsible for figuring that out. Moreover, with this solution, we can now effectively rely on domain experts to guide the learning process, for example, by informing us that, to keep its trajectory stable, the agent should be rewarded for limiting the amount of horizontal acceleration it is subject to. In light of this, the reinforcement learning paradigm shows great promise in allowing us to tackle such a complex task, and it has been used to solve it\footnote{See \fullcite{helicopter_rl}.}. In general, as long as we can come up with a sufficiently expressive reward function, reinforcement learning is --- at least in theory --- a suitable solution and, in many cases, might even be the only feasible approach to produce an agent able to reach high-performance levels.

\subsubsection{Limitations of the reinforcement learning approach}

Despite the undeniable advantages of RL techniques, these do not come without limitations. Firstly, while the reward function is fundamental in eliminating the need for an explicit description of how an agent should behave, it also bears a significant responsibility in determining the quality of the learning process.
The impact of an ill-designed reward function might range from convergence to sub-optimal behavior --- in the ``best'' case --- to a complete lack of progress in learning the task. Moreover, in some pathological --- yet easily encountered --- cases, it could also cause the agent to learn gimmicky policies that gain high rewards while achieving no progress for the actual task. 

To exemplify, let us consider the toy scenario of an agent tasked with reaching a door at the end of a corridor: the obvious choice would be to reward the agent for reaching their destination. However, if the corridor were sufficiently long, requiring the agent to take a large number of steps before reaching its end, this would lead to a very \textbf{sparse} reward, a fairly common problem encountered in reinforcement learning scenarios that could significantly hinder the progress of the agent. To deal with this issue, we could encourage the agent by rewarding it every time it progresses by, for instance, 25\% of the total length of the corridor. By doing so, the agent would have access to some \emph{partial reward} that could guide it towards its actual goal. This is the basic idea behind \textbf{reward shaping}, a technique that will be further discussed in section \ref{sec:rs}. This solution, however, hides a subtle problem: if improperly designed\footnote{Fortunately, \cite{reward_shaping} provides us with a general design that allows for the elimination of some of these pathological cases, as discussed in section \ref{sec:rs}.}, the partial rewards might lead the agent in learning a ``smart'' strategy to maximize its reward, consisting in reaching the first quarter milestone, taking a step back, going back to the milestone and so on ad infinitum. While, in such a simple environment, this pathological scenario could be foreseen, more complex ones usually do not allow for a simple apriori analysis and thus require extreme care in developing and testing adequate reward functions.

Another limit in the applicability of RL techniques comes directly from one of its foundational principles: the need for the agent to interact with its environment. The reason why this might represent a limitation is twofold. On the one hand, it requires us to be able to provide the agent with a copy of its environment. In many software-based tasks, this might be trivial. However, in most robotics applications, for instance, this requirement might pose a significant obstacle: obtaining access to a copy of the physical agent at training time might impose an impractical economic cost, exacerbated by the risk of damages arising from agent errors, especially in the initial phases of the training process, when the agent might not have yet reached safe levels of proficiency. In such cases, to make training feasible, there might be the need for emph{computer-simulated environments}, which themselves introduce a significant development cost. Moreover, even if the simulation represents a good proxy for the actual environment, there is usually no guarantee for the learned policy to transfer to the physical world seamlessly: agents acting optimally in the simulated environment might be overwhelmed by the intrinsic difficulties posed by the real world, for instance, in the form of sensor noise and unpredictable environment dynamics.

On the other hand, regardless of the nature of the agent's environment, the need for iterative interaction also represents a sensible cost in computational effort. Even apparently modest environments might require millions of environmental steps to provide the agent with a solid understanding of its laws and many more to fine-tune its behavior to reach optimality. Thus, in this regard, access to high-throughput environments --- in terms of steps-per-second --- is a fundamental requirement for assuring an effective training process and further highlights the difficulties related to using simulated environments, which often need to be based on accurate physics engines.

Finally, one further limit of RL techniques --- also shared by most supervised learning approaches --- is related to the generally low \textbf{generalization potential} of the learned policies to new scenarios: even though the agent might have reached optimal performance in its training environment, subtle changes in environment parameters or structure might be sufficient to cause a catastrophic drop in its ability to solve the task. While, in theory, this could be solved by exposing the agent to a broader array of environmental conditions, the feasibility of this approach is often limited by its associated computational costs and, above all, by the impossibility of determining beforehand which changes in the environment will lead to such problems.

\subsection{Markov Decision Processes}
\label{sec:mdps}
The tasks of interest in reinforcement learning are usually described as \textbf{Markov Decision Processes} (MDPs).
\begin{definition}
\label{def:mdp}
    A \emph{Markov Decision Process} is a tuple $\fullMDP$, where:
    \begin{itemize}
        \item S is the finite set of \emph{environment states};
        \item $F \subset S$ is the subset of \emph{final -- or goal -- states};
        \item A is the finite set of available \emph{agent actions};
        \item $r : S \times A \times S \rightarrow \mathbb{R}$ is a \emph{reward function} mapping from the current state of the environment, the action taken by the agent in such state, and the resulting state of the environment to a real-valued reward;
        \item $p(s_{t+1} | s_t, a_t)$ is the --- markovian --- \emph{environment transition model}, describing the probability of reaching environment state $s_{t+1} \in S$ after executing action $a_t \in A$ while in state $s_t \in S$;
        \item $\gamma \in (0,1]$ is the \emph{discount factor} weighting the preference for imminent rewards over future ones.
    \end{itemize}
\end{definition}
The presence of the ``Markov'' connotation is not a simple matter of terminological conventions. On the contrary, it highlights a fundamental property that the environment at hand must show: the current state of the environment must provide all the information needed to determine the next state of the environment, given the action taken by the agent. In other words, the knowledge of the environment \emph{history} does not grant any benefit in terms of computing the environment transition model. Formally, $\forall s_0,...,s_t,s_{t+1}\in S,\ a_0,...a_t \in A$ this can be expressed by the following condition:
\begin{equation}
        p(s_{t+1}|s_t,a_t) = p(s'|s_0,a_0,...,s_t,a_t)
\end{equation}
This motivates the specific formulation provided for defining the environment transition model. 

By adopting the above definition for an MDP, I implicitly restricted myself to a specific class of tasks characterized by \textbf{fully-observable, stochastic environments}, as showcased, respectively, by the absence of a distinction between \emph{environment states} and \emph{agent percepts}, and by the probabilistic definition of the environment transition model. Indeed, we are assuming that:
\begin{enumerate}
    \item on each timestep $t$, the agent percept coincides with the whole environment state $s_t$. In other words, the agent has, at any given moment, complete access to the true state of the environment;
    \item the outcome of an action $a_t$ taken while in state $s_t$ is non-deterministic;
\end{enumerate}
While true in many practical RL applications, the former assumption is currently motivated by the ease of exposition. It will be relaxed during the dissertation to accurately describe the environments that were analyzed by my work. The latter condition, instead, will remain unvaried as it plays an important role in justifying the need for reinforcement learning techniques, both in my work and in many other applications of RL. Moreover, its presence determines the need for a specific definition for the \textbf{solution} of an MDP.

In fact, in the case of a fully deterministic environment, a solution would be represented by \emph{any} sequence of actions leading from the initial state $s_0$ to a goal state $s_t \in F$. Moreover, informally, an \textbf{optimal solution} would be any solution maximizing the reward cumulated during the agent's trajectory in the environment. While a fascinating topic, the solution of such problems is often better handled with techniques different from RL, usually in the form of \emph{seach-based algorithms}, that do not share the same limitations in the previous section.

Instead, in the context of MDPs, a solution, to be complete, needs to specify the action to be taken in \emph{every} state since the stochasticity of the environment dynamics could, in theory, lead the agent to any state. Thus, instead of looking for a sequence of actions, we look for solutions in the form of \textbf{policies}.
\begin{definition}
    Given an MDP $\fullMDP$, a \emph{policy} is any function:
    \begin{equation*}
        \pi : S \rightarrow A
    \end{equation*}
    mapping from any environment state $s \in S$ to the action $a \in A$ to be taken by the agent while in such state.
\end{definition}
Having defined what represents a valid solution for an MDP, there is now the need for a method for \emph{comparing} the quality of two different solutions to characterize \textbf{optimal policies}, the ultimate objective when solving an MDP. To do so, the \textbf{utility} of an environment trajectory must first be defined.
\begin{definition}
    Given an MDP $\fullMDP$ and an environment trajectory $\tau = [s_0, a_0, ..., s_{t-1}, a_{t-1}, s_t]$ with $s_i \in S, \ a_i \in A \ \forall i = 0, ..., t$ its \emph{true utility} is defined as:
    \begin{equation*}
        U(\tau) = \sum_{k=0}^{t-1} \gamma^k r(s_k, a_k, s_{k+1})
    \end{equation*}
\end{definition}
The above equation also clarifies the purpose of the discount factor introduced by the definition of MDPs: when $\gamma = 1$, future rewards are weighted equally to imminent ones. At the same time, lower values progressively reduce their influence in the computation of a trajectory's utility.

Finally, we can then assess the quality of any given policy by quantifying the \textbf{expected utility} obtained by an agent executing it. Assuming the \textbf{stationarity} of the environment transition model, the probability distribution over environment trajectories is entirely determined by the initial environmental state $s_0$. For the agent's policy $\pi(s)$, we can simplify our notation by directly talking about the expected utility of a \emph{single state under a given policy}.
\begin{definition}
\label{def:state_utility}
    Given an MDP $\fullMDP$ and a policy $\pi$, we define the \emph{expected utility} of any state $s \in S$ as:
    \begin{equation*}
        U^{\pi}(s) = \mathbb{E}\left[\ \sum_{k=0}^{t-1} \gamma^k r(s_k, a_k, s_{k+1})\ \right]
    \end{equation*}
    where $s_0 = s$, $s_{i+1} \sim p(\cdot|s_i, a_i), \ a_i = \pi(s_i) \ \forall i$. The expectation is taken with respect to the probability distribution over environment trajectories induced by $s$ and $\pi$.
\end{definition}
\begin{definition}
    Given an MDP $\fullMDP$ and an initial environment state $s$, we define \emph{optimal policy} any policy:
    \begin{equation*}
        \pi^{*}_{s} = \argmax_{\pi} U^{\pi}(s)
    \end{equation*}
\end{definition}

Since, by definition, an optimal policy maximizes the expected utility cumulated by the agent during its execution, it follows that, for any state $s \in S$, its \textbf{true utility} must be equal to $U^{\pi^*_s}(s)$. Moreover, since any two \emph{different} optimal policies must, again by definition, lead to the same set of expected state utilities, we will denote the true utility of any state $s \in S$ simply as $U(s)$.

The existence of the true utility function $U(s)$ allows us to derive a conceptually simple criterion to be followed by an agent to act optimally: choosing the action that maximizes the expected utility of the subsequent state:
\begin{equation}
\label{eq:optimal_policy}
    \pi^*(s) = \argmax_{a \in A} \sum_{s'}p(s'|s, a)U(s')
\end{equation}
This Maximum Expected Utility (MEU) formulation, coupled with the one exposed by definition \ref{def:state_utility}, highlights the direct relationship that exists between the true utility of any state and its neighbors: assuming the agent is acting optimally, we can express the true utility of any state $s \in S$ as:
\begin{equation}
\label{eq:bellman}
    U(s) = r(s) + \gamma \max_{a \in A} \sum_{s'}p(s'|s,a)U(s')
\end{equation}
where, for ease of notation, we also assumed the reward obtained by the agent to only depend on the current state of the environment. Equation \ref{eq:bellman} is commonly known as the \textbf{Bellman equation}, after Richard Bellman, the mathematician who first described it in 1957. For a state space $S$ such that $|S| = n$, there are exactly $n$ Bellman equations, each describing the utility of a state as the sum of its associated reward and the maximum --- discounted --- expected utility of any of its neighbors. Moreover, the utilities of the states are their \emph{unique solutions}. As such, we would like to solve the Bellman equations since doing so would grant us the \emph{true utility} of each environment state and, in turn, the ability to determine an optimal MEU policy via equation \ref{eq:optimal_policy}. Unfortunately, while ideal in theory, the practical application of this approach becomes rapidly infeasible as the size of the state increases, both due to the linear dependency of the number of equations to $|S|$ and the difficulties posed in their solution by their non-linear nature. 

Nonetheless, the discussion of said approach is relevant as it forms the conceptual basis from which many modern reinforcement learning algorithms are built. Two basic approaches can be followed to exploit the Bellman equations: \textbf{value iteration} and \textbf{policy iteration}.

\subsubsection{Value iteration}
Algorithm \ref{alg:value_iteration} presents the pseudocode for value iteration. As showcased by its simple, iterative structure, this approach consists of a progressive refinement in the estimates for the state utilities. It progresses from arbitrarily chosen --- zero, in our case --- values until a fixed point is reached. The core of this method lies in the \emph{Bellman update}, a direct application of the Bellman equation to the current utility estimates. With a series of local updates, the information relating to each state's utility is propagated from the right-hand side of the equation to the current estimates, progressively reducing their associated error. It can be shown that, after a sufficient number of iterations, value iteration is \emph{guaranteed} to converge to the unique set of solutions of the Bellman equations, \ie the true utility function.

One interesting property demonstrated in practice by value iteration is its ability to lead to optimal MEU policies long before $U_i$ has converged to the true utilities. Intuitively, as long as the estimates are sufficiently accurate to distinguish between \emph{good} and \emph{bad} states, there is no need for extreme precision in their values.

\begin{algorithm}
\caption{Value iteration algorithm for producing an arbitrarily accurate estimate of state utilities given an MDP $\MDP$ and a tolerance parameter $\epsilon$.}
\label{alg:value_iteration}

    \begin{algorithmic}
        \Function{Value-Iteration}{$\MDP, \epsilon$}
        \Local
            \State U, U': vectors of utility estimates, zero-initialized
            \State $\delta$: the maximum variation in utility estimates during an iteration
        \EndLocal

        \State \Comment Iterative estimation of utility values
        \Repeat 
            \State $U \gets U',\ \delta \gets 0$
            \ForAll{$s \in S$}
                \State $U'[s] \gets r(s) + \gamma\max\limits_{a} \sum\limits_{s'}p(s'|s,a)U[s']$
                \Comment Bellman update
                \If{$|U'[s] - U[s]| > \delta$}
                    \State $\delta \gets |U'[s] - U[s]|$
                \EndIf
            \EndFor
        \Until $\delta < \epsilon(1 - \gamma) / \gamma$
        \State \Return U
        \EndFunction
    \end{algorithmic}
    
\end{algorithm}

\subsubsection{Policy iteration}
The value iteration algorithm leads to optimal policies \emph{indirectly} by first computing the true state utilities and then adopting the MEU approach to compute the actual policy. On the other hand, policy iteration reverses this approach: first, the current policy is used to estimate each state's utility, then these estimates are used to improve the policy --- once again --- according to the application of the MEU framework to each state's successors. 

Algorithm \ref{alg:policy_iteration} provides the pseudocode for the policy iteration algorithm. When the current policy $\pi_t$ yields no change in the utilities, a fixed point in the Bellman update has been reached and, thus, $\pi_t$ must be optimal. Moreover, the termination of the algorithm is guaranteed since the policy space for an MDP with a finite state space is finite as well, and every iteration necessarily yields an improved policy.

The \textproc{Policy-Evaluation} subroutine has been intentionally left unspecified in the pseudocode, as there are multiple possible ways of implementing it. One possibility is to apply the Bellman equation directly. However, since the current policy fixes the action to be taken in each state, we can adopt the following \emph{simplified} version of the equation:
\begin{equation}
    U_i(s) = r(s) + \gamma \sum_{s'}p(s'|s, \pi_i(s))U_i(s')
\end{equation}
which is significantly easier to solve due to the absence of nonlinearity arising from the $\max$ operator. Using standard linear algebra methods, given a state space $S$ such that $|S| = n$, the associated system of $n$ equations can be \emph{exactly} solved with time $\bigO(n^3)$. Unfortunately, as n grows, the cubic time complexity quickly becomes prohibitive. Luckily, just like for value iteration, computing the exact utilities arising from the current policy is unnecessary. By once again applying the simplification above --- this time to the Bellman update --- we can obtain an estimated policy evaluation method that is feasible even for large state spaces\footnote{In this context, by \emph{large} state spaces we are nonetheless referring to state spaces small enough to make the application of policy iteration feasible.}:
\begin{equation}
    U_{i+1} = r(s) + \gamma \sum_{s'}p(s'|s,\pi_i{s})U_i(s)
\end{equation}
With this approach, the utility estimates can be efficiently obtained by repeatedly applying this update rule $k$ times, with $k$ being a parameter that can be tuned to find the right balance between computational cost and estimation accuracy.

\begin{algorithm}
\caption{Policy iteration algorithm for producing an optimal policy for an MDP $\MDP$.}
\label{alg:policy_iteration}

    \begin{algorithmic}
        \Function{Policy-Iteration}{$\MDP$}
        \Local
            \State U: vector of utility estimates, zero-initialized
            \State $\pi$: state-indexed policy vector, randomly initialized
        \EndLocal

        \State \Comment Iterative refinement of current policy
        \Repeat 
            \State $U \gets$ \Call{Policy-Evaluation}{$\pi, U, \MDP$}
            \State $unchanged? \gets true$
            \ForAll{$s \in S$}
                \State $improvable? \gets \max\limits_{a}\sum\limits_{s'}p(s'|s,a)U[s'] > \sum\limits_{s'}p(s'|s, \pi[s])U[s']$
                \If{$improvable?$}
                    \State $\pi[s] \gets \argmax\limits_{a}\sum\limits_{s'}p(s'|s,a)U[s']$
                    \Comment Policy improvement
                    \State $unchanged? \gets false$
                \EndIf 
            \EndFor
        \Until $unchanged?$
        \State \Return $\pi$
        \EndFunction
    \end{algorithmic}
    
\end{algorithm}

\subsubsection{Solving MDPs with reinforcement learning}
In the previous sections, I assumed the MDP of interest to be completely \emph{known} to the agent: all the presented equations and algorithms rely on the knowledge of the state space $S$, action space $A$, reward function $r(s)$, and environment transition model $p(s'|s, a)$. When this --- quite strong --- assumption holds true, there might be a need to consider reinforcement learning techniques for solving the MDP. However, many practical scenarios do not allow for such an assumption to be made and, instead, are characterized by either a lack of knowledge regarding the environment transition model, reward function, or both. Moreover, there might even be cases where the state and action spaces are partially or wholly unknown to the agent.

When the MDP is \textbf{unknown}, the agent initially has no idea of what he can do, which states it can reach, what the outcome of its actions is, or which achievements are desirable for solving the task. Thus, in such a situation, the only possibility for it to learn anything is to \textbf{explore} the environment and progressively figure out the inner workings of its surroundings while trying to maximize the rewards observed during its trials. Therefore, when faced with such a scenario, reinforcement learning techniques are not only effective but also \emph{necessary} for successfully training an agent.

\subsubsection{Partially observable MDPs}
\label{sec:pomdps}
The last assumption about the nature of an environment modeled by an MDP that we need to relax --- before describing the tasks that I analyzed during my work properly --- is that of \emph{full observability}. However, the previously provided definition of a Markov Decision Process is insufficiently expressive to do so. Therefore, this section introduces a new class of MDPs: \textbf{Partially Observable Markov Decision Processes} (POMDPs).
\begin{definition}
    A \emph{Partially Observable Markov Decision Process} is a tuple $\fullPOMDP$, where:
    \begin{itemize}
        \item $S, F, A, r, p$, and $\gamma$ are defined in the same way as for regular MDPs;
        \item $O$ is a finite set of \emph{observations} that the agent can obtain from the environment;
        \item $\omega(s|o)$ is the \emph{environment observation model}, describing the probability of the environment being in state $s \in S$ when the agent observes $o \in O$.
    \end{itemize}
\end{definition}
In the POMDP setting, the true environment state at any point in time is unavailable to the agent, which can only obtain \emph{observations} that might, depending on the environment observation model, partially reveal the information the agent needs for progressing in the task at hand. In the best case, given the latest observation $o_t \in O$ obtained by the agent, there might exist some state $\bar{s} \in S$ for which $\omega(\bar{s}|o_t) \simeq 1$, thus allowing the agent to infer the underlying environment state reliably. Unfortunately, such a condition is neither guaranteed nor common in practice. On the other hand, in the worst case, the agent could obtain an observation $\bar{o} \in O$ such that $\omega(s|\bar{o}) = \frac{1}{|S|} \ \forall s \in S$, thus effectively providing no information at all regarding the state of the environment. While, in practice, many tasks may not be characterized by an abundance of this last pathological case, the agent could have no knowledge of the environment observation model. This condition substantially increases the complexity already introduced by the environment's partial observability.

It is important to note that, even in the case of partial observability, the Markov property is still required by the above definition of the environment transition model: the knowledge of the true state of the environment is nonetheless sufficient for fully determining the probability distribution governing the state transition dynamics. However, as the agent has no access to the actual state of the world, this may \emph{appear} to him as being non-Markovian. For instance, there might be two \emph{distinct} states $s_1,s_2 \in S$, leading, after the agent's action, to the same observation $o \in O$ but different transition probabilities.

\subsubsection{Agent policies under partial observability}
In the case of partial observability, the previously provided definition for an agent policy is intrinsically inadequate, as it relies on direct knowledge of the true environmental state. To adapt it to the POMDP setting, it might be tempting to resort to the following naive extension:
\begin{equation}
\label{eq:bad_pomdp_policy}
    \pi : O \rightarrow A
\end{equation}
However, while such a definition could show \emph{some} level of effectiveness in practice, from a theoretical standpoint, it is inherently inadequate due to POMDPs being non-Markovian concerning the environment observation space. Intuitively, this formulation is equivalent to requiring the agent to act based \emph{only} on what he currently sees. To consolidate this intuition, I will resort to the example of a blackjack-playing agent, as such card games are a perfect example of POMDPs. The current order of the cards in the deck, \ie the state of the environment, is fixed and fully determines the order in which they will be extracted, \ie the environment state transitions. On the other side, the drawn cards represent the observations available to the agent. Resorting to the formulation proposed by equation \ref{eq:bad_pomdp_policy} would be equivalent to requiring the agent always to do the same thing when faced with the same cards, even though previous extractions might have removed cards from the deck, thus changing the overall probability distribution of future extractions\footnote{For the sake of simplicity, I am knowingly ignoring the fact that, in real casino blackjack, the deck of cards is changed between games to avoid the possibility for card counting strategies.}. 

To solve this problem, we need to consider policies in the form of:
\begin{equation}
\label{eq:better_pomdp_policy}
    \pi : O^* \rightarrow A
\end{equation}
which considers the whole history of observations up to the present moment. By doing so, the agent can effectively exploit all the information it can gather \emph{from the environment} to decide its course of action. Unfortunately, this formulation can again not lead to optimal behavior in the general case. While keeping track of past observations is undoubtedly an improvement, there is a source of influence on the environment state transitions that are being unaccounted for: the agent's actions. Thus, this observation leads to the following formulation:
\begin{equation}
\label{eq:best_pomdp_policy}
    \pi : (O \times A)^* \rightarrow A
\end{equation}
which effectively allows the agent to exploit \emph{all} the information it has to try and act optimally under an MEU interpretation. This is possible as, by keeping track of the history of observations and actions, the agent can make its best possible guess for the true underlying state of the environment. Formally, this last concept of "agent's guess" is usually presented as a \textbf{belief state}:
\begin{definition}
    Given a POMDP $\fullPOMDP$, a \emph{belief state} is a probability distribution over the set of true environment states $S$:
    \[
        b : S \rightarrow [0,1] \qquad \text{s.t. } \sum_{s \in S}b(s) = 1
    \]
    that represents the probability of each state being the current environment state.
\end{definition}
On the first timestep, the agent's belief state solely depends on the environment observation model and the first observation received from the environment:
\[
    b_0(s) = \omega(s|o) \qquad \forall s \in S
\]
Then, at any timestep $t \geq 1$, the new belief state also depends on the environment transition model and the agent's action:
\[
    b_{t+1}(s') = \alpha \omega(s|o)\sum_{s \in S}p(s'|s,a)b(s) \qquad \forall s,s' \in S,\ a \in A
\]
It can be shown that, in the context of POMDPs, the optimal action at each point in time depends only on the agent's current belief state. In light of this, we can provide one last formulation for a POMDP policy:
\begin{equation}
\label{eq:belief_state_pomdp_policy}
    \pi : \mathcal{B}_S \rightarrow A
\end{equation}
where $\mathcal{B}_S$ represents the set of all possible belief states over a set $S$ of environment states. However, since partially observable reinforcement learning tasks are usually characterized by unknown environments, this last formulation is rarely explicitly used in practice and is mainly adopted for the theoretical analysis of POMDPs. In fact, explicitly computing belief states is impossible without knowledge regarding $p(s'|s,a)$ and $\omega(s|o)$. Therefore, the policy formulation most commonly adopted during the design of reinforcement learning algorithms is the one presented in equation \ref{eq:best_pomdp_policy}.

To conclude this section, it is important to note that the study of POMDPs is an important research topic in the field of reinforcement learning, as many real-world tasks might fall under this category for a variety of reasons, such as the unavailability of sufficiently exhaustive sensor arrays, perceptual noise, or intrinsic uncertainty of available information. In light of this, there exists an active research community investigating the algorithms and techniques required to deal with the unique challenges posed by such environments effectively. Among them, one approach that stood out in recent years in the neurosymbolic RL literature is \textbf{reward machine-based reinforcement learning}. Due to its central role in my work, the entirety of chapter \ref{chap:rms} is dedicated to thoroughly describing such a framework.

\subsection{Classes of RL techniques}
Reinforcement learning algorithms are usually classified based on three main features:
\begin{enumerate}
    \item the object of focus of the learning process;
    \item the reliance on an explicit model for predicting the evolution of the environment;
    \item the source of the experiences the agent uses to derive the optimal policy.
\end{enumerate}

Starting from their object of focus, RL techniques can either be \textbf{value-based} of \textbf{policy-based}, depending on whether they follow an overall approach that is comparable with the value or policy iteration algorithms presented in section \ref{sec:mdps}: in the former case, the algorithms aim at learning utility estimates, which are then used to derive the optimal policy. On the other hand, policy-based algorithms directly tweak their current policy based on its observed quality: the learning process continues as long as the performance of the policy improves, and then it stops. As this approach can be interpreted as a search in the space of policies, another connotation commonly used to refer to such techniques is, unsurprisingly, that of \emph{policy-search} methods. Usually, the policy space that is explored is induced by a specific parametrization in the policy representation.

Then, depending on whether the agent relies on the environment transition model to learn its policy or each state's utility, RL algorithms can be classified as either \textbf{model-based} or \textbf{model-free}. In the case of model-based techniques, the model for $p(s'|s,a)$ can either be known a priori, for instance, in game-playing settings, or progressively learned by the agent. In light of this, both value iteration and policy iteration are model-based techniques. On the other side, model-free techniques do not rely on any kind of prediction of the environment's evolution and rely solely on the experience obtained via direct exploration of the environment. This, in practice, provides model-free methods with a few advantages over model-based ones: on the one hand, they apply to a broader range of problems; on the other hand, they are usually computationally cheaper in terms of time and space complexity. At the same time, model-free methods come with some disadvantages when compared with model-based ones: they are usually characterized by a lower \emph{sample efficiency}, meaning they need a larger number of trials to progress in the learning process, and, depending on the exact techniques at hand, they might present the risk of relying on biased samples of agent experience.

Finally, a third distinction can be made between \emph{on-policy} and \emph{off-policy} techniques. To understand the difference, a distinction must be introduced between the \textbf{target policy} and the \textbf{behavior policy}: the latter is the policy the agents use for deciding which actions need to be taken during learning, while the former represents the current agent's estimate of the optimal policy it aims to learn. When the target and behavior policy are distinct from each other, the algorithm is \textbf{off-policy}; on the other side, when the agent acts to improve its current policy, the two policies are, in fact, the same, and the technique is said to be \textbf{on-policy.} Intuitively, an off-policy agent observes samples of behavior and uses them to determine how one should ideally act. In contrast, an on-policy agent acts and tries to understand how it could improve its strategy.

\subsection{Tabular Q-learning}
One commonly used and widely studied reinforcement learning algorithm is \textbf{Q-learning}. The name of this technique derives from its use of a \textbf{Q-function}, which gives the value of taking an action in a given environmental state:
\begin{definition}
    Let $S, A$ be the sets of environment states and agents actions of an arbitrary MDP. A \emph{Q-function} is any function:
    \[
        Q : S \times A \rightarrow \mathbb{R}
    \]
    such that:
    \[
        U(s) = \max_{a}Q(s,a) \quad  \forall s \in S,\ a \in A
    \]
    where $U(s)$ is the true utility of state $s$, as for equation \ref{eq:bellman}.
\end{definition}
Using a Q-function is not a simple change in the utility's adopted representation. Indeed, it shows a fundamental property: an agent learning a Q-function does not need any model $p(s'|s,a)$ for either learning or selecting the best action. In light of this, Q-learning is a \emph{model-free} method. Moreover, it is a \emph{value-based} method following the same conceptual structure of value-iteration: starting from initially arbitrary values, the agent keeps an estimate for the Q-value of each state-action pair, then chooses an action based on the MEU principle:
\[
    \pi(s) = \argmax_{a}Q(s,a)
\]
Then, after taking action $\pi(s)$, the agent uses the obtained reward to update its estimates of Q-values. The cycle is repeated until convergence to the true Q-values and, consequently, to the optimal policy. Indeed, Q-learning is guaranteed to converge to the optimal Q-values in the limit of visiting each state-action pair infinitely often \cite{watkins1992}. 

While the Bellman equation for utilities can be readily adapted to the use of a Q-function to provide both a condition that must hold for optimal Q-values and a possible update rule:
\[
  Q(s, a) = r(s) + \gamma \sum_{s'}p(s'|s,a)\max_{a'}Q(s',a')
\]
The Q-learning algorithm does not directly use this equation to learn the correct Q-values, as it does require knowledge of the environment transition model. Instead, the update rule used by the method is given by:
\begin{equation}
    Q(s,a) \gets Q(s,a) + \alpha (r(s) + \gamma \max_{a'}Q(s',a') - Q(s,a)
\end{equation}
where $\alpha \in [0,1]$ is a hyper-parameter controlling the magnitude of the updates, usually referred to as the \emph{learning rate}. This update rule is applied every time an action $a$ is executed in state $s$, leading to state $s'$. 

Moreover, notice that Q-learning is an \emph{off-policy} method, as the Q-value updates do not depend on the policy used to choose the action taken. Thus, in principle, a Q-learning agent can use any behavioral policy to generate the experiences used to update the Q-value estimates and, thus, the current best-known policy. Indeed, in light of the condition for the convergence of Q-learning, most implementations resort to \textbf{exploratory policies} that allow the agent to explore the state-action space thoroughly, as an insufficient number of experiences relating to even one critical, state-action pair might be sufficient for Q-learning not being able to converge to the optimal policy. At the same time, however, resorting to poor policies might lead the agent not to visit relevant sequences of states and actions, thus hindering the time required for convergence. Therefore, an agent must adequately balance the need for \textbf{exploration} with the benefits associated with the \textbf{exploitation} of its current knowledge regarding the best possible course of action.

One standard solution to this problem is using an \textbf{$\epsilon$-greedy exploration} strategy. At each time step, the agent chooses a random action with probability $\epsilon \in [0,1]$ or the action suggested by its current Q-values otherwise. In many practical cases, doing so allows for an adequate balance between exploiting the current best-known policy and exploring the environment, given the proper tuning of the $\epsilon$ parameter. In conclusion, as a concise summary of all the exposed concepts, algorithm \ref{alg:q_learning} provides the pseudo-code for the core of a Q-learning agent training process under an $\epsilon$-greedy exploration strategy.

\begin{algorithm}
\caption{Pseudocode for one step of a Q-learning agent's learning process, using an $\epsilon$-greedy exploration strategy, given the current state $s'$ and reward $r'$}
\label{alg:q_learning}

    \begin{algorithmic}
        \Function{Q-Learning-Agent}{$s',r'$}
        \Static
            \State $Q[s,a]$: table of Q-value estimates, zero-initialized;
            \State $s, a, r$: the previous state, action, and reward, null-initialized
        \EndStatic

        \If{\Call{Terminal?}{$\MDP,s$}}
            \State $Q[s, null] \gets r'$
        \EndIf
        \State
        \If{$s \neq null$}  \Comment Update Q-value estimates
            \State
            \State $Q[s,a] \gets Q[s,a] + \alpha (r + \gamma \max\limits_{a'}Q[s',a'] - Q[s,a])$
            \State $explore?\ \gets\ $ \Call{Random}{$0, 1$}
            \If{$explore? < \epsilon$} \Comment $\epsilon$-greedy exploration
                \State $action \gets$ \Call{Random-Action}{{}}
            \Else
                \State $action \gets \argmax\limits_{a'}Q[s',a']$
            \EndIf
            \State \Comment Update previous state, action, and reward
            \State $s,a,r \gets s', action, r'$ 
        \EndIf

        \Return $action$

        \EndFunction
    \end{algorithmic}
    
\end{algorithm}

\subsection{Deep Q-learning}
Despite its theoretical convergence guarantee and the practical advantages of its off-policy, model-free nature, Q-learning has one major drawback that strongly limits its applicability in practical scenarios. As the state-action space grows, it rapidly becomes impractical due to its lack of generalization potential. Since the Q-values are estimated separately for each state-action sequence, training an agent to learn the optimal policy for every possible scenario would require an enormous amount of computational resources, both in terms of time, to visit each sequence multiple times, and space, to store the Q-value estimates of each state-action pair, making the training process completely unfeasible for sufficiently large state-action spaces.

However, since the Q-learning approach remains theoretically valid regardless of the Q-function's representation, the solution adopted in practice consists of substituting the tabular design in favor of some parametrized representation $Q_(s,a;\theta)$. In the case of simple Q-functions, this might be possible with some linear function approximators; however, when facing complex environments, a more suitable solution is to use non-linear deep neural network models due to their unique properties as generalized function approximators. When used to approximate a Q-function, such models are usually referred to as \textbf{deep Q-networks} (DQN), and the resulting algorithm is called \textbf{deep Q-learning}.

As for any neural network model, a DQN can be trained by iteratively adjusting its parameters in such a way that minimizes the value of a \emph{loss function} quantifying the distance between the current network's outputs and the correct values --- a.k.a. the \emph{targets} --- of the function being approximated. This can be done thanks to the \textbf{backpropagation} algorithm, which uses \textbf{gradient descent} to compute the parameter updates that lead to the minimization of the loss function. In the specific case of Q-learning, the DQN can be trained by minimizing:
\begin{equation}
\label{eq:dqn_loss}
    L_i(\theta_i) = \E_{s,a \sim \rho(\cdot)} \left[ (y_i - Q(s,a;\theta_i))^2\right]
\end{equation}
for iteration $i$, where $y_i$ are the target Q-values and $\rho(s,a)$ is the distribution governing the probability of encountering the state-action pair $(s,a)$. As this distribution depends on the agent's behavior while collecting its experiences, it is usually referred to as the \emph{behavior distribution}. However, if the targets $y_i$, \ie the real Q-values, were known, there would be no point in going through the process of training a DQN: instead, they too need to be estimated, and this is done by resorting to the MEU approach on the \emph{previous} Q-value estimates:
\begin{equation}
\label{eq:dqn_targets}
    y_i = \E_{s' \sim p(\cdot)}\left[r + \gamma\max_{a'}Q(s', a'; \theta_{i-1})\right]
\end{equation}

When the parameter updates are carried out for every agent step, and the full expectations are replaced with single samples from the behavior and environment model distributions --- as motivated by the \emph{stochastic} variant of the gradient descent algorithm --- the learning process is analogous to the one from standard tabular Q-learning, leading to a model-free, off-policy technique.

Unfortunately, naively adapting Q-learning to the use of DQNs leads, in most cases, to abysmal performance or even the Q-network diverging\footnote{\ie causing a steady increase in the value of the loss function as training progresses.}, as the experience samples collected by the agent violate the \iie \footnote{\ie independent and identically distributed} assumption that neural network training algorithms rely on. Consecutive samples of experience are usually highly correlated as environments mostly evolve sequentially.

An effective solution for this problem was introduced by \cite{mnih_playing_2013}, in the form of an \textbf{experience replay} mechanism: instead of using each experience sample $e_t = (s_t, a_t, r_t, s_{t+1}$ immediately, the agent stores them in a set $\mathcal{D}$, the \textbf{replay memory}. Then, after collecting a given number of samples, the Q-learning update is applied using a batch of experiences randomly drawn from the replay memory. This approach presents two critical advantages over standard Q-learning. Firstly, it allows for improved data efficiency, thanks to each experience sample being used in many updates. Then, thanks to the random sampling of experiences from the replay memory, the \iie assumption is recovered, thus significantly reducing the risk of oscillations or, in the worst case, the divergence of the training process. In light of this, experience replay proved to be a fundamental requirement for the practical use of DQNs, leading to an enormous increase in their applicability to many previously infeasible tasks. To summarise, algorithm \ref{alg:dqn_experience_replay} presents the pseudocode for deep Q-learning with experience replay.

\begin{algorithm}
\caption{Pseudocode for the Deep Q-learning with Experience Replay algorithm for a given MDP $\MDP$, running over $M$ episodes, using replay memory size $N$, batch size $n$ and update frequency $k$.}
\label{alg:dqn_experience_replay}

    \begin{algorithmic}
        \Function{DQN-ER}{$\MDP, M, N, n, k$}
        \Local
            \State $\mathcal{D}$: FIFO queue of $N$ agent experiences, initially empty.
            \State  $Q(s,a;\theta)$: Q-network with weight vector $\theta$, randomly initialized.
        \EndLocal

        \For{$episode = 1, ..., M$}
            \State $t \gets 1$
            \State $s_1 \gets$ \Call{Initial-Environment-State}{$\MDP$}
            \Repeat
                \State $a_t \gets \max\limits_{a}Q(s_t,a;\theta)$
                \State $s_{t+1}, r_t \gets$ \Call{Execute-Action}{$\MDP, a_t$}
                \State \Call{Push}{$\mathcal{D}, (s_t, a_t, r_t, s_{t+1})$}
                \State
                
                \If{$t\ mod\ k = 0$} \Comment Update weights $\theta$
                    \State $batch \gets $ \Call{Sample-Experiences}{$\mathcal{D}, n$}
                    \For{$i = 1, ..., n$}
                        \State $s_j, a_j, r_j, s_{j+1} \gets batch[i]$ 
                        \State
                        \State
                        $
                            y_i =
                            \begin{cases}
                                r_j \quad &\text{if \Call{Is-Terminal?}{$s_{j+1}$}} \\
                                r_j + \gamma \max_{a'}Q(s_{j+1}, a';\theta) \quad &\text{otherwise}
                            \end{cases}
                        $
                    \EndFor
                    \State
                    \State \Comment Loss function as in equation \ref{eq:dqn_loss}
                    \State $\theta \gets $ \Call{Gradient-Descent}{$y_1, ..., y_n$} 
                    \State $t \gets t+1$
                \EndIf
                
                \State
            \Until \Call{Is-Terminal?}{$\MDP, s_{t+1}$}
        \EndFor

        \Return $Q$

        \EndFunction
    \end{algorithmic}
    
\end{algorithm}

\subsection{Reward Shaping}
\label{sec:rs}
As previously anticipated, one common issue encountered in reinforcement learning is the \textbf{sparseness} of the reward signal. In many environments, the agent might be required to take many steps before receiving a reward $r \neq 0$. This presents a problem as it leaves the agent without feedback on its actions for long and, consequently, without information to update its estimates for the encountered states' utility. In the worst case, this phenomenon might severely hinder the agent's ability to find the optimal policy; for instance, in the unfortunate case the latter involves a significantly larger number of zero-reward steps --- before finally reaching a highly rewarding terminal state --- than a sub-optimal policy, which leads to a medium-rewarding terminal state. In the case of Q-learning, its convergence guarantee might seem like a solution to this problem. While true on a theoretical level, the convergence requirement of an infinite number of visits to \emph{every} state-action pair is impossible to satisfy perfectly in practice, leaving failure as a possibility unless adequate measures are taken to ensure the environment is explored thoroughly. Nonetheless, such precautions could lead to a sensible increase in the time required to converge. Moreover, when using algorithms that do not benefit from any kind of convergence guarantee, the risk of consequences arising from the sparseness of rewards is even higher.

To solve this problem, one solution is to \emph{lead} the agent into following good paths in the environment by providing a series of intermediate rewards\footnote{Intuitively, this is the same idea motivating the practice of luring someone to a location of choice by leaving a trail of money on the ground.}. This is the main idea behind \textbf{reward shaping}, which formally translates into a substitution of the original reward function $r(s, a, s')$ with a new one:
\[
    r'(s,a,s') = r(s,a,s') + f(s, a, s')
\]
for a given function $f : S \times A \times S \rightarrow \mathbb{R}$, which is commonly defined as a \textbf{shaping function}.
However, as anticipated by the corridor example from section \ref{sec:rl}, for the shaping reward function to help, intermediate rewards must be designed in a way that avoids the introduction in the environment of new policies that, while numerically optimal, lead to behavior not relevant for the task's completion.

Fortunately, \cite{reward_shaping} provides a simple condition that the $f(\cdot)$ function must satisfy to guarantee the preservation of the original set of optimal policies by the shaping reward function, based on the concept of \textbf{potential-based} shaping functions:
\begin{definition}
    Given an MDP $\fullMDP$ and a function $f : S \times A \times S \rightarrow \mathbb{R}$, we say that $f$ is a \emph{potential-based shaping function} if there exists a function $\Phi: S \rightarrow \mathbb{R}$ such that:
    \[
        f(s, a, s') = \gamma \Phi(s') - \Phi(s) \quad \forall s \in S - {s_0}, a \in A, s' \in S
    \]
\end{definition}

The use of such a class of shaping functions is sufficient in guaranteeing the preservation of every optimal policy in the original MDP.

\chapter{Reward machine-based RL}
\label{chap:rms}

Reward machines (RMs) are a simple yet effective, automata-based formalism recently proposed in the neurosymbolic reinforcement learning literature to facilitate an agent's learning process by granting it access to a structured representation of the reward signal. In particular, this approach is very effective in dealing with tasks characterized by a strong hierarchical structure. By decomposing the reward function in the form of a finite-state automaton, agents can keep explicit track of their current progress in the task, thus allowing them to:
\begin{itemize}
    \item increase their \emph{overall proficiency} by learning a set of sub-policies, each expressing the adequate behavior for progressing in a specific portion of the task;
    \item improve the \emph{sample efficiency} of the learning process by allowing for the generation of synthetic experiences.
\end{itemize}

Initially proposed in the context of fully observable MDPs by \cite{icarte_rms_2018}, the framework has been extended to deal with partially observable environments in \cite{icarte_lrm_2019}. In such problems, one key issue is the need for effective \textbf{memory mechanisms} that allow the agents to maintain a model of the unobservable portion of the environmental state. In this regard, reward machine-based techniques demonstrated the ability to converge to optimal policies in benchmark environments where standard state-of-the-art RL algorithms equipped with LSTM-based neural network memory architectures could make no progress. Thus, the approach proved to be a viable choice for dealing with memory-related issues in partially observable environments.

\section{Intuition, definitions, and properties}
\label{sec:rm_intuition}
The fundamental intuition behind the use of reward machines is to introduce a higher layer of abstraction in the representation of the environment: every time the agent observes the current state of the world, it supplies it to a \textbf{labeling function} which, in turn, has the role of detecting any \emph{high-level events} that might have verified. Depending on the task at hand, detecting an event of interest might trigger a transition in the reward machine state: this, intuitively, corresponds to a progression in the overall task. Moreover, each state of a reward machine can be associated with a different reward function, thus allowing for the effective modeling of complex reward dynamics, possibly consisting of conditional choices or loops defined over the given set of high-level events.

\begin{figure}
\label{fig:simple_cw}
    \centering
    \includegraphics{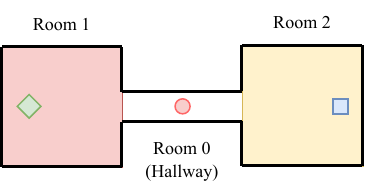}
    \caption{Simple Cookie World environment. The agent is represented by the red circle, the cookie by the green diamond, and the button by the blue square.}
\end{figure}

To clarify the above intuition, let us consider an example of a simple, fully observable environment: \emph{Simple Cookie World}\footnote{The ``simple'' connotation will be clear to the reader in Section \ref{sec:domains}.}, depicted in figure \ref{fig:simple_cw}. The agent starts in the hallway and aims to eat the cookie in room 1, which gives a reward of $\rwP{+1}$. Moreover, due to its incredible hunger, the agent receives a reward of $\rwN{-0.1}$ for each step before eating the cookie, except when pressing the button. However, the cookie appears only after the agent has pressed the button in room 2. Full observability is granted by allowing the agent to see the content of both rooms while in either of them. The high-level events for this simple task are $\ao{B}$, indicating the agent has pressed the button, and $\ao{C}$, indicating the agent has eaten its desired treat. In addition, the dummy event $\ao{--}$ represents the absence of any other event.

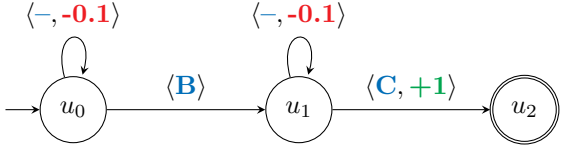
\begin{figure}
\label{fig:scw_rm}
\centering

    \begin{tikzpicture}[initial text={}, >=stealth, auto, node distance=3cm, on grid, initial where=left]

        \node[state,initial]            (0) at (0,0) {$u_0$};
        \node[state]                    (1) at (3,0) {$u_1$};
        \node[state,accepting]          (2) at (6,0) {$u_2$};

        \path[->] (0) edge  node  [above]  {$\rmEdge{\ao{B}}$} (1);
        \path[->] (1) edge  node  [above]  {$\rmEdge{\ao{C},\rwP{+1}}$} (2);
        
        \path[->] (0) edge [loop above] node [above] {$\rmEdge{\ao{--},\rwN{-0.1}}$} (0);
        \path[->] (1) edge [loop above] node [above] {$\rmEdge{\ao{--},\rwN{-0.1}}$} (1);
        
    \end{tikzpicture}

\caption{Reward machine for the Simple Cookie World.}
\end{figure}

The 3-state reward machine in figure \ref{fig:scw_rm} accurately describes the structure of the task and, dually, its reward signal. The reward machine's initial state $u_0$ represents the beginning of the task when the agent has not yet pressed the button. Then, after pressing the button, the state transitions to $u_1$, indicating the cookie's presence in room 1. Finally, after the cookie is eaten, the RM transitions to its terminal state $u_2$, marking the successful completion of the task.

\subsection{Full observability}
Starting from the case of fully observable environments, let us formally define a reward machine:
\begin{definition}
\label{def:rm}
    Let $\propsSet$ be a set of propositional symbols, and $S, A$ be the sets of environment states and agent actions for an arbitrary MDP. \\
    A \emph{Reward Machine} is a tuple $\fullRM$, where:
    \begin{itemize}
        \item $U$ is a finite set of \emph{states};
        \item $u_0 \in U$ is the \emph{initial state};
        \item $T$ is a finite set of \emph{terminal states} such that $U \cap T = \emptyset$;
        \item $\delta_u : U \times 2^\propsSet \rightarrow U \cup T$ is the \emph{state-transition function};
        \item $\delta_r : U \rightarrow [S \times A \times S \rightarrow \mathbb{R}]$ is the \emph{state-reward function}.
    \end{itemize}
\end{definition}
In light of the intuition presented in the previous section, the definition is easily interpreted: $\propsSet$ represents the set of every possible high-level event that is relevant to the task at hand, $\delta_u$ is the function governing the state transitions of the reward machine, based on its current state and the latest subset of events detected in the environment. Finally, $\delta_r$ is the function associating each reward machine state to a different reward function. Let us now properly define a labeling function:
\begin{definition}
    Let $\propsSet$ be a set of propositional symbols and $S, A$ be the sets of environment states and agent actions for an arbitrary MDP. \\
    A \emph{labeling function} is any function in the form:
    \begin{equation*}
        L : S \times A \times S \rightarrow 2^\propsSet
    \end{equation*}
\end{definition}
Given the state of the environment $s \in S$, the action taken by the agent $a \in A$, and subsequent state $s' \in S$, $L(s, a, s') \subseteq 2^\propsSet$ is the set of propositional symbols, \ie events, that currently hold in the environment.

All the formal tools needed for presenting a fundamental observation that motivates the techniques which will be examined later in this chapter are now available:
\begin{observation}
\label{obs:mdp_rm}
    Given an arbitrary MDP $\fullMDP$, a labeling function $L$, and a reward machine $\fullRM$, a new MDP $\MDPRM = (S', A, F, p', r', \gamma)$ can be defined, where:
    \begin{itemize}
        \item the new state space is defined as the cross product between environment and reward machine states:
        \[
            S' = S \times U
        \]
        \item the new reward function $r': S' \times A \times S' \rightarrow \mathbb{R}$ produces its output based on the reward function associated with the current reward machine state:
        \[
        r'(\langle s,u \rangle, a, \langle s',u' \rangle) =
        \begin{cases}
            \delta_r(u)(s, a, s') & \text{if } u \notin T \\
            0 & \text{otherwise}
        \end{cases}
        \]
        \item the new environment transition model $p'$ is defined by:
        \[
        p'(\langle s',u' \rangle|\langle s,u\rangle, a) = 
        \begin{cases}
            p(s'|s,a) & \text{if } u \in T \land u' = u \\
            p(s'|s,a) & \text{if } u \in U \land u' = \delta_u(u, L(s, a, s')) \\
            \ \ \quad  0 & \text{otherwise}
        \end{cases}
        \]
        While apparently complex, this definition requires the reward machine state dynamics to be respected for a transition from $\langle s, u \rangle \in S'$ to $\langle s', u' \rangle \in S'$ to be possible in the new cross-product state space. In other words, it associates a null probability to every transition leading from a non-terminal RM state to another one unless the output of the labeling function properly triggers such transition. This aside, the underlying environment transition model is left unmodified.
    \end{itemize}
    
\end{observation}

The above observation serves a dual purpose: on the one hand, it provides a proper formalization for the integration of reward machine-based reward signals into arbitrary MDPs; on the other hand, it motivates the idea of augmenting the environmental state space with the high-level information provided by reward machine states, which -- as we will see -- is the main advantage offered by the adoption of the RM framework, especially when dealing with partial observability.

\subsubsection{Dealing with non-markovian reward signals}
In terms of their expressive potential, reward machines are significantly more powerful than the regular reward functions introduced by definition \ref{def:mdp}:
\[
    r: S \times A \times S \rightarrow \mathbb{R}
\]
In fact, by providing the latter definition I implicitly extended the assumption of a \textbf{Markovian} environment transition model to the reward signal as well. However, when faced with \textbf{non-Markovian} reward signals, such a definition is intrinsically inadequate in representing, for instance, rewards that change over time, or depending on the history of agent actions and/or environment transitions.
On the contrary, the reward machine-based formulation is able to represent a subclass of signals in the form:
\[
    R : (S \times A \times S)^* \rightarrow \mathbb{R}
\]
mapping environment trajectories to their associated reward --- intuitively --- as long as such trajectories can be correctly summarized by the reward machine states, thus leading, at any given time, to the usage of the correct reward function $\delta_r(u)$. In such cases, by passing from the original MDP to the RM-based MDP characterized by the cross-product state space, the Markov property of the reward function can be recovered, thus highlighting another advantage arising from the state space augmentation.

\subsubsection{Simple reward machines}
The definition of the state-reward function $\delta_r$ presented in definition \ref{def:rm} might be unnecessarily complex for many practical tasks of interest. In such cases, we can restrict ourselves to a smaller class of reward machines:
\begin{definition}
\label{def:simple_rm}
    Given a set $\propsSet$ of propositional symbols, a \emph{Simple Reward Machine} is a tuple $\fullRM$, where:
    \begin{itemize}
        \item $U, u_0, T, \delta_u$ are defined as for a regular RM;
        \item the state-reward function
        \[
            \delta_r : U \times 2^\propsSet \rightarrow \mathbb{R}
        \]
        also depends on $2^\propsSet$ and directly returns a real number.
    \end{itemize}
\end{definition}
Due to a lack of need for the increased expressiveness of general reward machines, from now on, when referring to reward machines, I will implicitly assume this more straightforward formulation. 

As a final note, it is easy to show that simple reward machines are, in fact, a subclass of regular reward machines:
\begin{observation}
Let L be an arbitrary labeling function. Any simple reward machine $\fullRM$ is equivalent to a regular reward machine $\RM' = (U, u_0, T, \delta_u, \delta'_r)$, with:
\[
    \delta'_r(u)(s, a, s') = \delta_r(u, L(s, a, s'))\quad \forall u \in U,\ s,s' \in S,\ a \in A
\]
\end{observation}

\subsection{Partial observability}
\label{sec:rms_po}
As previously anticipated, using a reward machine can be a very effective tool for dealing with partially observable environments. Recalling the definitions presented in section \ref{sec:pomdps}, POMDPs require policies that depend on the history of agent observations and actions. From a practical point of view, this requires the agents to be provided with some type \textbf{memory mechanism} allowing them to keep track of such history effectively. However, implementing effective memory mechanisms is a complex technical challenge, especially when the task requires deep-learning-based RL algorithms. 

As argued by \citeauthor{morad_popgym_2023} in \cite{morad_popgym_2023}, when de-facto standard implementations of state-of-the-art RL algorithms are analyzed, it is easy to observe the lack of availability of a wide array of different memory architectures. Most implementations resort to \textbf{frame stacking}, \ie the simple concatenation of subsequent observations, while others resort to the use of \textbf{recurrent neural network} architectures such as Long Short-Term Memory (LSTM) or Gated Recurrent Units (GRU). However, the former approach is only suitable for tasks that pose extremely short-term memory requirements; on the other side, the latter, while able to handle much longer temporal dependencies, imposes a significant computational cost to the training procedure. While they do not represent a one-fits-all solution for the memory problem in POMDPs, reward machines can represent a powerful tool for increasing an agent's ability to keep track of crucial information, thus increasing its overall memory capacity even without any other memory mechanism.

Since the formalization of a --- simple --- reward machine presented by definition \ref{def:simple_rm} does not depend on any specific component of an MDP, it can be adopted without modification in the context of POMDPs. The same does not apply to the definition of a labeling function, as it requires environmental states. Nonetheless, adjusting such a definition to the case of partial observability is trivial:
\begin{definition}
    Let $\propsSet$ be a set of propositional symbols and $O, A$ be the sets of agent observations and actions for an arbitrary POMDP. \\
    A \emph{labeling function} is any function in the form:
    \begin{equation*}
        L : O_\emptyset \times A_\emptyset \times O \rightarrow 2^\propsSet
    \end{equation*}
    where, for any set $X$, $X_\emptyset = X \cup \{\emptyset\}$ by definition.
\end{definition}
Aside from the substitution of environment states with agent observations, the only minor difference lies in the first two terms of the function's domain, which are now allowed to be empty. Conventionally, $L(\emptyset, \emptyset, o)$ represents the initial set of high-level events that can be detected upon entering the environment before the agent takes any action.

Thus, by having access to a proper labeling function, an agent can exploit a reward machine in a partially observable setting in the same way it would if faced with a fully observable environment: given its current reward machine state $u_t$, the agent, for each observation $o_t$ it obtains, determines the action $a_t$ to be taken according to its policy and executes it, receiving a new observation $o_{t+1}$. Then, the agent receives an output $\sigma_t = L(o_t, a_t, o_{t+1})$ from its labeling function, which can finally be used to determine the next state of the reward machine $u_{t+1} = \delta_u(u_t, \sigma_t)$ and reward $r_t = \delta_r(u_t, \sigma_t)$. After updating its policy via its method of choice, the agent starts a new iteration of this procedure.

\subsubsection{Reward machines as agent memory}
While, in a fully observable environment, the usefulness of reward machines mainly derives from the use of specially designed learning algorithms that exploit their structure, in partially observable settings, they intrinsically allow the agent to keep track of information that might no longer be available from its current observations. To clarify, let us refer to the example environment from section \ref{sec:rm_intuition}. When its reward machine is in state $u_1$, the agent knows that the cookie is in room 1. However, due to the assumed full observability, the agent does not need the RM to access this information; it can simply see it by itself. However, if we remove the agent's ability to see everything and instead allow him only to see the content of the room it is currently in, thus making the environment partially observable, suddenly the $u_1$ reward machine state enable the agent to know about the presence of the cookie \emph{even though it may not be able to see it}. Therefore, the reward machine effectively becomes a form of \textbf{memory} for the agent, thus providing him with a fundamental tool to deal with the complexity introduced by partial observability. In other words, a reward machine can provide an agent with \emph{additional} information, otherwise unavailable, that it can use to guide its learning and decision-making processes.

However, the reward machine is not the source of such information: it could more appropriately be described as its \emph{storage medium}\footnote{For the sake of precision, it is the current reward machine state that stores such information. Thus, the set of reward machine states could equivalently be seen as a set of propositions about the state of the environment. Only the proposition associated with the current state can be deemed true at any given time.}. Instead, the \emph{actual source} of such information is the labeling function, which, via its outputs, allows the reward machine to transition between different states, thus allowing the agent to update its current knowledge regarding the high-level state of the environment. 

This observation leads to two fundamental insights: On the one hand, a proper design of the labeling function allows the agent to be provided with information that is not yet available. For instance, in the partially observable Simple Cookie World, the agent's RM transitions to $u_1$ upon pressing the button\footnote{For example, when $L(o, a, o') = \ao{B}$.}, thus, before the agent has had any opportunity to observe the presence of the cookie in the other room. Furthermore, a labeling function could even provide the agent with information that could \emph{never} be available to him via its observations. Under the latter scenario, the labeling function could be seen as a potential method of giving \textbf{advice} to the agent.
On the other hand, the role of the labeling function makes its correct design --- and operation --- highly critical to the agent's training and decision-making correctness. This observation is the fundamental idea motivating the attack techniques I propose with my thesis, thoroughly presented in chapter \ref{chap:attacks}.

\subsubsection{Perfect reward machines}
In light of the above considerations, ideally, in a partially observable setting, a reward machine should allow keeping track of relevant past information in such a way that renders the environment Markovian with respect to $O \times U$. A reward machine showing this property is called a \textbf{perfect reward machine} for the environment at hand and can be formally defined as follows:
\begin{definition}
    A reward machine $\fullRM$ is said to be \emph{perfect} for a given POMDP $\POMDP$ with respect to a labeling function $L$ if, and only if, for each trajectory $(o_0, a_0, ..., o_t, a_t)$ generated by any policy over $\POMDP$:
    \[
        Pr(o_{t+1}, r_t | o_0, a_0, ..., o_t, a_t) = Pr(o_{t+1}, r_t | o_t, u_t, a_t)
    \]
    where $Pr(A|B)$ denotes the conditional probability for $A$ given $B$.
\end{definition}
Indeed, the above definition requires the Markov property to hold for the probability of observing any given observation-reward pair, given the current observation, reward machine state, and agent action. Also, it is to be noted that whether a reward machine is perfect or not for a given environment depends on the choice of the labeling function, thus once again highlighting its fundamental role in the proper operation of RM-based RL agents.

Finally, this definition leads to two important theorems, proved by \cite{icarte_lrm_2019}, that effectively enable the use of perfect reward machines:
\begin{theorem}
    Given any arbitrary POMDP $\POMDP$ with a \emph{finite}, \emph{reachable} belief space, there will always exist at least one perfect reward machine $\RM$ for $\POMDP$ with respect to \emph{some} labeling function L.
\end{theorem}
Despite the nice existence guarantee provided by this theorem, its usefulness is mainly theoretical, as it gives no indication of such a reward machine's size or structure, nor does it hint at the labeling function that must be used. Nonetheless, it ultimately provides the theoretical basis that allows one to know that finding a perfect reward machine is, at least, always possible as long as the POMDP of interest is characterized by a specific type of belief space.

On the contrary, the following theorem provides a much more important guarantee, as it ensures that the use of a perfect reward machine is sufficient for finding optimal policies for any given POMDP:
\begin{theorem}
    Let $\RM$ be a perfect reward machine for a given POMDP $\POMDP$ with respect to a labeling function $L$. Then, any optimal policy for the $\RM$-augmented\footnote{Once again, I am referring to the equivalent POMDP obtained by adapting observation \ref{obs:mdp_rm} to the case of partial observability, \ie the POMDP characterized by the cross-product space $O \times U$.} POMDP is also optimal for $\POMDP$.
\end{theorem}

\section{Exploiting RMs for learning}
As anticipated in the introduction of this chapter, the main purpose of a reward machine is to aid the agent's learning process: doing so requires the use of specially designed algorithms. 

This section presents two such techniques: \emph{Q-learning with Reward Machines} (QRM) --- \cite{icarte_rms_2018} --- and \emph{Automated Reward Shaping} (ARS) --- \cite{icarte_rms_2022}. The former is a variation of Q-learning that relies on the state-space augmentation presented in observation \ref{obs:mdp_rm} to allow for the generation of synthetic experiences that the agent can use to speed up its convergence, while the latter is a novel technique that exploits the agent's reward machine to automatically compute a potential-based shaping function that can then be used to introduce reward shaping in the agent's training.

\subsection{Q-Learning with Reward Machines}
Being a variation of standard Q-learning, the \textbf{Q-learning with Reward Machines} algorithm follows the same conceptual design: it uses the agent experiences to update its Q-value estimates iteratively until convergence to the true Q-function. Moreover, it also uses the Q-values to determine the current agent's behavioral policy under the MEU framework. In light of this, just like standard Q-learning, QRM is a value-based, model-free, off-policy algorithm. Furthermore, it also allows for the seamless use of DQNs for the representation of the Q-function, thus leading to the DQRM variation of the algorithm. However, by operating over the --- equivalent --- RM-augmented MDP obtained from the original MDP via the transformations presented by observation \ref{obs:mdp_rm}, the QRM algorithm introduces two key modifications aimed at allowing the agent to fully exploit the high-level information and task structure exposed by its reward machine.

The first of these expedients is the use of a different Q-function $\Tilde{q}_u(s,a)$ for each reward machine state. Thus, the global Q-function becomes:
\[
    Q(\ab{s,u}, a) = \Tilde{q}_u(s, a) \quad \forall u \in U
\]
This effectively allows the agent to learn a different policy for each reward machine state, potentially aiding the learning process when the task of interest is characterized by a series of sub-phases, each requiring the agent to adopt a substantially different strategy. Moreover, the potential benefits arising from this are even higher in partially observable environments in light of the memory interpretation of RMs presented in section \ref{sec:rms_po}. On the other hand, the main downside of this approach lies in an increase in the space required to store the Q-function, which now effectively needs to store $n = |U|$ values for each state-action pair or, for DQRM, $n$ distinct neural networks.

\subsubsection{Counterfactual reasoning with reward machines}
The second, most important, tweak over standard Q-learning consists of the use of counterfactual reasoning as a tool for simulating the outcome the agent would have obtained were it acting while in a different reward machine state, thus allowing for the generation of synthetic experience that can be used to speed up the learning process. This approach --- aptly termed \emph{counterfactual experiences for reward machines} (CRM) --- is based on the following observation:
\begin{observation}
    Suppose the agent took action $a$ while in cross-product state $\ab{s,u}$, consequently reaching $\ab{s',u'}$ and obtaining a reward $r$. Then, had the initial RM state been $\Bar{u}$, the agent would have reached state $\ab{s', \delta_u(\Bar{u}, L(s, a, s'))}$ while receiving a reward $\Bar{r} = \delta_r(\Bar{u}, L(s,a,s'))$.
\end{observation}
This observation follows directly from the definition of a reward machine and is made possible by the fact that the use of a reward machine does not alter the environment dynamics: regardless of the current RM state, taking action $a$ in environment state $s$ will always lead to a new state $s'$ that depends solely on the environment transition model $p(s'|s,a)$. Therefore, since the reward machine dynamics are known and superimposed over the environmental ones, they can be easily simulated by the agent via the above method, effectively allowing for the generation of synthetic experience in the RM-augmented state space. Therefore, thanks to CRM, after each step, the agent, in addition to $(s, u, a, r, s', u')$, can use the following set of experiences to update its Q-value estimates:
\[
    \{(s, \Bar{u}, a, \delta_r(\Bar{u}, L(s, a, s')), s', \delta_u(\Bar{u}, L(s, a, s'))\ |\ \forall \Bar{u} \in U\}
\]
effectively multiplying the data acquisition rate by a factor of $n = |U|$.
As a final summary of the presented concepts, algorithm \ref{alg:qrm} provides the pseudocode for the QRM algorithm.

\begin{algorithm}
\caption{Pseudocode for the Q-learning with Reward Machines algorithm for a given MDP $\MDP$, reward machine $\RM$ and labeling function $L$, running over $k$ episodes with learning rate $\alpha$. Note that $x \overset{\alpha}{\gets} y$ is used as shorthand notation for $x \gets x + \alpha(y - x)$}
\label{alg:qrm}

    \begin{algorithmic}
        \Function{DQN-ER}{$\MDP, \RM, L, k, \alpha$}
        \Local
            \State $\Tilde{q}_u[s,a]: \forall u \in U$, table of Q-values estimates, zero initialized.
        \EndLocal

        \For{$episode = 1,...,k$}
            \State $u \gets u_0$
            \State $s \gets $ \Call{Initial-Environment-State}{$\MDP$}
            
            \Repeat
                \State $a \gets \max\limits_{a'}\Tilde{q}_{u}[s,a']$
                \State $s' \gets $ \Call{Execute-Action}{$\MDP, a$}
                \State $l \gets L(s, a, s')$ \Comment Detect high-level events
                \State $r, u' \gets \delta_r(u, l),\ \delta_u(u, l)$
                
                \State \Comment Counterfactual reasoning
                \State $experience \gets \{\ab{s, \Bar{u}, \Bar{a}, \delta_r(\Bar{u}, l), s', \delta_u(\Bar{u}, l}\ |\ \forall \Bar{u} \in U\}$
                \For{$\ab{s, \Bar{u}, a, s', \Bar{u}'} \in experience$}
                    \If{\Call{Is-Terminal?}{$\MDP, s'$} or \Call{Terminal-State?}{$\RM, u'$}}
                        \State $\Tilde{q}_{\Bar{u}}[s,a] \overset{\alpha}{\gets} r$
                    \Else
                        \State $\Tilde{q}_{\Bar{u}}[s,a] \overset{\alpha}{\gets} r + \gamma\max\limits_{a'}\Tilde{q}_{\Bar{u}'}[s,a']$
                    \EndIf
                    
                \EndFor
                \State
                \State $s, u \gets, s', u'$

            \Until \Call{Is-Terminal?}{$\MDP, s'$} or \Call{Terminal-State?}{$\RM, u'$}
            
        \EndFor

        \Return $\{\Tilde{q}_u\ |\ \forall u \in U\}$ 

        \EndFunction
    \end{algorithmic}
    
\end{algorithm}

\subsection{Automated Reward Shaping}
Reward machines allow for more than the generation of synthetic agent experience. Indeed, by exploiting their structure, it is possible to automatically determine a potential-based shaping function that can be used to introduce reward shaping into the agent's training process and, thus, all its associated advantages.
The main intuition behind this technique --- named \textbf{Automated Reward Shaping} --- is the same motivating reward shaping itself: rewarding an agent for making progress towards the completion of its task. In general, this requires one to arbitrarily decide which $(s,a,s')$ tuples are associated with meaningful progress that should be rewarded. However, the task structure exposed by an agent's reward machine already contains such information, which can, therefore, be extracted automatically by considering the RM itself as an MDP and using the value iteration algorithm to compute the utility $V(u)$ of each RM state. Then, the obtained utilities can be used to define the potential function $\Phi(s,u) = V(u)$\footnote{In the original publication, the authors actually define $\Phi(s,u) = -V(u)$. By experimenting with both approaches, I found the positive-signed alternative to also work in practice, while leading to easier-to-interpret shaping rewards. In light of this, I decided to prefer it to the original formulation.} for each environment state $s$ and RM state $u$. The following observation and algorithm \ref{alg:ars} formalize the described approach.

\begin{observation}
    Given any simple reward machine $\fullRM$, one can construct an MDP $\fullMDP$, where:
    \begin{itemize}
        \item the MDP state space is defined by the reward machine state space:
        \[
            S = U \quad F = T
        \]
        \item the agent actions are defined by every possible sub-set of high-level events:
        \[
            A = 2^\propsSet
        \]
        \item the reward function is induced by the reward machine:
        \[
            r(u,\sigma,u') = \delta_r(u, \sigma) \quad \forall u,u'\in U,\ \sigma \in 2^\propsSet
        \]
        \item the environment transition model derives from the reward machine state dynamics:
        \[
            p(u'|u,\sigma) = 
            \begin{cases}
                1 \quad & \text{if } u \in T \text{ and } u' = u \\
                1 \quad & \text{if } u \in U \text{ and } u' = \delta_u(u,\sigma) \\
                0 \quad & \text{otherwise}
            \end{cases}
        \]
        \item the discount factor $\gamma$ can be chosen arbitrarily in $(0,1]$ 
        
    \end{itemize}
\end{observation}

\begin{algorithm}
\caption{Pseudocode for the Automated Reward Shaping algorithm for a given reward machine $\fullRM$ using discount factor $\gamma$}
\label{alg:ars}

    \begin{algorithmic}
        \Function{ARS}{$\RM, \gamma$}
        \Local
            \State $V[u]$: array of utility estimates $\forall u \in U \cup T$, zero initialized.
        \EndLocal

        \State $error \gets 1$
        \While{$e > 0$}
            \State $error \gets 0$
            \For{$u \in U$}
                \State $v' \gets \max\limits_{\sigma \in 2^\propsSet}\{\delta_r(u, \sigma) + \gamma V[\delta_u(u,\sigma)]\}$
                \State $error \gets$ \Call{Max}{$e, |V[u] - v'|$}
                \State $V[u] \gets v'$
            \EndFor
        \EndWhile

        \Return V
        \EndFunction
    \end{algorithmic}
    
\end{algorithm}

\chapter{Attacking reward machine-based agents}
\label{chap:attacks}
Given the many benefits originating from the use of reward machines in reinforcement learning, it is reasonable to believe that, in future years, their adoption will steadily increase, allowing for the solution of many real-world tasks that are now deemed too hard to deal with. However, as for any real-world deployment of recently developed technologies, a solid understanding of their associated security implications is imperative for avoiding the potentially huge risks arising from their use, possibly originating from design flaws, environmental factors, or, in the worst case, adversarial agents knowingly aiming at undermining the system functioning.
Moreover, a deeper understanding of the conditions and techniques that could lead to undesirable outcomes arising from the use of reward machines might also have an intrinsic value, providing meaningful insights that could, in turn, lead to progress being made toward the improved effectiveness of the original approach.

While, from a theoretical point of view, the RL research community rapidly started investigating many variations of the basic RM-based approach in numerous different settings, its security implications remain unexplored by currently available literature. For this reason, in this chapter, I present what is, to the best of my knowledge, the first proposed class of attacks on reward machine-based reinforcement learning.

\section{Threat model}
Before any type of attack on RM-based agents can be designed, it is fundamental to obtain a solid understanding of every component involved in their operation, its relationships with other entities, its impact on the system as a whole, and the practical requirements it poses to real-world deployments of RM-based agents. Following standard practice in security research, I thus lead the first part of my investigation according to the methodology of \textbf{threat modeling}, a systematic approach aimed at highlighting the aforementioned information relating to the system under analysis, together with any assumption relating to the attacker's objective, knowledge and capabilities. After being collected, all such information can then be used to create a \emph{threat model} accurately outlining all the requirements that an attacker must be able to satisfy to carry out an attack successfully.

\subsection{Objective, knowledge and capabilities}
Before analyzing the different elements of vulnerability characterizing an RM-based agent, in this section, I  clarify the preliminary assumptions motivating the choices that I will describe later, thus giving the reader a clearer picture of the design process that ultimately led to the proposed class of attacks. By adopting such assumptions, I aimed to define an attack methodology characterized by:
\begin{enumerate}
    \item applicability to a wide range of scenarios;
    \item weak requirements for the attacker;
    \item ease of execution and implementation;
    \item high impact on the agent's behavior;
\end{enumerate}
thus potentially posing as much of a risk for practical deployments of reward machines as possible.

\subsubsection{Attacker's objective}
In light of this, the first assumption relates to the objective pursued by the attacker. Given the reinforcement learning setting, there are many possibilities as to what an attacker could aim to obtain from its attack. In the most general case, the adversarial actor might simply want to degrade the agent's performance, leading to a situation commonly referred to as \emph{denial-of-service} (DoS). From a cybersecurity perspective, this type of attack is both very common and impactful, since it \emph{usually} does not require highly specialized knowledge or techniques to be carried out, and it prevents the victim's organization from acquiring the value, arising from the trained agent's operativity, it needs to compensate for the high amount of resources needed to train and deploy the agent in the first place. As an example, let us consider the domain of financial portfolio management\footnote{A simple Google Scholar query for "deep reinforcement learning portfolio management" leads to a plethora of research work, thus making the chosen example reasonable for our purposes.}, a highly safety-critical field where any agent's malfunction could lead to enormous economic loss for its parent organization. Assuming an agent is trained to decide the amounts of resources to be allocated for each investment in its portfolio, an adversarial actor could carry out a DoS attack to hinder the agent's ability to make good decisions, thus leading to bad investments and, consequently high monetary losses. 

Alternatively, the attacker might aim at inducing the agent to act following a \emph{specific trajectory} of its choice, in order to collect any benefit he could gain from such a course of action. This is equivalent, in terms of the above example, to the adversarial actor attempting to drive the agent into making a specific set of investments that might somehow benefit him. This type of attacks --- named \textbf{evasion attacks}, as they allow the attacker to "evade" the agent's decision-making logic --- are usually more sophisticated than DoS, as they typically require some form of knowledge regarding the attacked model or, in lack thereof, some technique allowing for \textbf{model extraction}. They generally allow the attacker to pursue more subtle objectives and are thus harder to detect. When the subtlety requirement is weaker than the attacker's need for consistent agent behavior, the former might aim to induce the agent into learning a specific policy, leading to what is commonly referred to as a \textbf{poisoning attack}, for instance by causing it to \emph{systematically} making a specific set of investments in favor of attacker-controller enterprises, regardless of their quality as an asset.

Among all the different possibilities, for which the above examples provide just a notable selection, I decided to assume the \textbf{DoS objective} for the attacker, as it represents the scenario that is applicable to the widest set of practical scenarios. While there might be a limited number of tasks where an attacker could be reasonably interested in inducing specific behavior from an agent, any time someone resorts to reinforcement learning for solving some problem --- reasoning under the "everyone is hostile" assumption that characterizes most cybersecurity research --- there could be someone else, such as an unfair business rival, aiming at causing the malfunctioning of the system. Moreover, evaluating a DoS scenario carries the additional benefit of allowing one to assess the robustness of the agent to both adversarial settings and unforeseen environmental noise, a common occurrence in the real world. Finally, the last motivating factor behind my choice lies in its associated lack of need for specialized knowledge of the victim model, thus aiding in achieving a simpler attack design.

\subsubsection{Attacker's knowledge}
From an attacker's knowledge point of view, there are three types of assumptions that are usually considered when carrying out cybersecurity research. The first one is the \textbf{white-box} assumption, which grants the attacker \emph{complete knowledge} of the internal structure of the system under analysis. In the context of reinforcement learning, this translates to the attacker having full knowledge of the algorithm used to train the agent, its policy, and any other relevant parameter, depending on the context at hand. For instance, in the case of a DQN agent, the white box assumption would grant the attacker unlimited knowledge regarding the internal structure of the agent's deep Q-networks, including the value for each network weight. It is immediately apparent how this type of assumption is really hard to satisfy in practice. Nonetheless, its adoption is useful in allowing an analyst to assess the potential impact arising from a worst-case scenario point of view: by granting an attacker with unlimited knowledge one can determine an upper bound for the damages arising from any realistic attack.

On the other end, the opposite possibility consists of the \textbf{black-box} assumption, which grants the attacker no knowledge whatsoever regarding the inner workings of the target system. Thus, the only information he can resort to consists of what is publicly available, for instance in the form of the system's well-known inputs and outputs. As an example, assuming a black-box scenario in reinforcement learning might translate to allowing the attacker only to see the agent acting in its environment, with no type of access to its internal state. From an attack's impact perspective, this type of assumption usually leads to a lower impact, since the adversary has a substantially lower amount of information to help him achieve its goals. However, this reduced impact is counterbalanced by the now-higher attack feasibility, which makes black-box attack far more common in practice than their white-box counterpart. At the same time, assuming a complete lack of information from the attacker's perspective might represent an unrealistically restrictive hypothesis, as rarely would one try and attack a system without any type of prior gathering of relevant information. 

Moreover, many attacks are carried out in an incremental fashion, starting from the acquisition of an initial foothold on the target system and progressively leading to further exploitation thanks to the gradual increase in available information and capabilities. In light of this, the third possibility consists of the \textbf{gray-box} assumption, which can be seen as the middle point between the previous ones: like the white-box one, it allows some degree of knowledge regarding the inner workings of the system, but, like in the black-box case, no knowledge is assumed regarding the rest of the system. Being a combination of the previous approaches, the gray-box assumption likely represents the most realistic proxy for most real-world cybersecurity attacks, usually requiring some kind of prior knowledge and, at the same time, allowing for a considerable impact on the victim system. 

During my investigation on the security of RM-based agents, I tried to adhere to the black-box assumption, with the aim of designing an attack that requires the lowest possible amount of prior information, thus increasing its overall practical feasibility. Even though, as will be discussed in the following sections, clearly classifying the attacks that I developed as either black-box or gray-box might be, to some degree, debatable, I nonetheless discarded, during the design phase, any choice that would intrinsically require a substantial relaxation of the black-box hypothesis.

\subsubsection{Attacker's capabilities}
Defining what might represent a relevant \emph{capability} in the context of a cybersecurity attack clearly depends on both the theoretical and practical context of interest. In the case of reinforcement learning, a plethora of them could be easily identified: manipulating the agent's environment to alter its dynamics, modifying the agent percepts before it can obtain them, and altering the reward signal are all reasonable examples of things that an attacker might need to do to achieve its goals. Therefore, to reduce the breadth of possibilities that could be considered, a first fundamental distinction must be made between \textbf{training-time} and \textbf{deploy-time} capabilities, where the firsts include any action the attacker could carry out during the agent's training process and, conversely, the seconds consist of any type of intervention targeting the agent acting to achieve the task it was previously trained for. Generally speaking, obtaining training-time access to a reinforcement learning agent is very difficult. Given its critical role, the agent's training process is usually conducted in isolated development environments protected by layers of firewalls, intrusion prevention systems, and other kinds of defenses.\footnote{In practice, the degree of truth of this statement \emph{highly} depends on the organization at hand and its overall security posture. Nonetheless, given any entity with a sufficient amount of resources and expertise, it is reasonable to assume this to be at least partially true.} On the contrary, once an agent is ready to be deployed, its operation is more likely to be supported by near-public systems with just a limited amount of indirection introduced between them and the end-user. Therefore, once again for feasibility's sake, I decided to assume the hypothetical adversary only to have deploy-time access to its victim.

A second, general class of potential capability assumptions arises depending on whether the attacker is able only to observe some kind of agent-related data --- thus having \textbf{read} capabilities --- or also to introduce new data ---  \ie \textbf{injection} capabilities --- or to modify existing data --- \ie \textbf{tampering} capabilities. While it may appear impossible for an attack to be feasible when only having read capabilities, some attacks, such as model extraction, might indeed just rely on such an assumption. Nonetheless, restricting an attacker to this scenario drastically reduces the amount of possible attacks, with the remaining ones usually being characterized by a low impact. This is especially true when the read-only assumption is adopted in conjunction with the black-box and deploy-time ones. Therefore, during the design of my attacks, I allowed the attacker to have some type of write access when needed.

\subsection{Elements of vulnerability}
By recalling the defining elements of a reward machine $\fullRM$, the potential high-level targets for an attack can be readily identified in:
\begin{itemize}
    \item its \emph{structure}, represented by its set of states $U \cup T$ and the connections between them, implicitly defined by $\delta_u$;
    \item its \emph{state dynamics}, represented by the interactions between the state-transition function $\delta_u$ and the agent's labeling function $L$;
    \item its \emph{reward dynamics}, represented by the state-reward function $\delta_r$ and, again, its interactions with the agent's labeling function $L$.
\end{itemize}
Given their fundamental importance in determining the agent's behavior both during its training, as is true for all three of them, or during its post-training operativity, as is the case for just the former two, all three of the identified alternatives potentially represent a valid target for an attack aimed at impairing the agent's performance. 
However, thanks to the assumptions presented in the previous section, two of them can immediately be ruled out, thus showcasing the value of the threat modeling methodology in allowing one to outline its possible routes of attack clearly. Indeed, since the agent's reward machine structure is fixed before the training process is even started, looking for some way to alter it would violate the deploy-time assumption. Moreover, even assuming some degree of knowledge relating to the reward machine states or their connections would translate into granting the attacker access to a large portion of the agent's internal structure, thus conflicting with my stated preference towards the black-box setting. Similar considerations apply to the choice of the RM's reward dynamics as a potential target: on the one hand, the reward signal is only relevant for the agent during its training\footnote{For the sake of simplicity, I am knowingly ignoring online-learning settings, where an agent could still be able to learn during the operativity \emph{following} its deployment.} and thus any attempt to exploit it is ruled out once again by the deploy-time assumption. On the other hand, granting the agent any knowledge relating to its reward signal would represent no less of a violation of the black-box assumption' than granting him read access to its reward machine's structure.

Therefore, the only remaining possibility lies in the choice of the RM's state dynamics as the attack's high-level target. Then, after determining the pursued plan, the attack's scope can be further restricted by considering the single components that play a role in determining how an RM transitions between its states. As previously observed, there are two of such components: the RM's \emph{state-transition function} $\delta_u$, and the \emph{labeling function} $L$, which cooperate in determining which state the agent's reward machine should transition to, based on its current state and the high-level events detected in the environment. Thus, this suggests a possible attack plan: An adversarial actor could aim at tampering with such dynamics in a way that leads the agent into a false belief regarding the high-level state of the environment, represented by its reward machine state according to the memory-based interpretation presented in section \ref{sec:rms_po}. 

From a theoretical standpoint, this looks highly promising as an agent might need to rely heavily on its reward machine state to derive optimal policies, especially in partially observable environments. To exemplify, recall the Simple Cookie World example from chapter \ref{chap:rms}, in its partially observable variation. In such a scenario, the optimal policy for the agent is very simple and maps perfectly onto its reward machine: at the beginning of the episode, \ie when its RM is in state $u_0$, and there is no cookie yet to be eaten, the agent needs to go right until it finds the button, then press it. Conversely, when the agent knows the cookie is available, \ie when its RM is in state $u_1$, the only thing the agent needs to do is go left until it reaches the cookie. Therefore, it is easy to see how an error in the agent's reward machine could lead to its complete inability to achieve the task: if, for instance, the agent's reward machine were to transition to state $u_1$ mistakingly, with the cookie not yet available, the agent would be stuck \emph{ad infinitum} in going left, trying to reach a cookie that does not exist. While a toy example in nature, the showcased scenario is not to be excluded a priori as a possibility in any more complex environment, where the increase in complexity is very likely to hide subtle dependencies between the agent's policy and its reward machine state, thus potentially giving rise to analogous situations. For these reasons, I chose to pursue this exact attack plan to try and hinder an RM-based agent's performance.

Thus, in light of the above considerations, there remains only one element that needs to be defined before the actual attack strategy can be outlined: the low-level target of the attack. Between the RM's state-transition function and the agent's labeling function, I chose the latter.
The labeling function, with respect to the stated assumptions and objectives for the design of the attack, represents the perfect target for two reasons. On the one hand, its fundamental role in triggering the agent's RM state transitions grants an attacker a wide range of possibilities in terms of the potential consequences arising from its careful manipulation. On the other hand, by being a conceptually separated entity from the agent's reward machine, the labeling function might, in practice, allow for a wider range of solutions to be adopted for its implementation. Indeed, it is not unreasonable to hypothesize an agent having the need for two separate programs running on its physical architecture: one computing its policy and, therefore, its reward machine dynamics, and one with the sole role of computing the labeling function outputs. For instance, this could be a valid practical solution in the case of tasks characterized by the agent receiving observations in the form of images. In such cases, detecting high-level events, such as the presence of a given object, might require the use of complex computer vision models, likely in the form of convolutional neural networks. Thus, in such a scenario, it would be reasonable to decouple the actual agent's program from the event-detection one to obtain an array of potential advantages, including easier development, lower maintenance, efficiency, and modularity. In some cases, it might even be desirable to resort to network-based solutions requiring the agent to forward its $(s,a,s')$ tuples to a labeling server which, in turn, reports the detected events back to the agent via some type of communication channel. All of the above scenarios introduce many openings that an attack could exploit for tampering with the labeling function, thus highlighting another reason for its choice.

\subsection{Potential attack methodologies}
\label{sec:attack_methodologies}
\begin{figure}
    \centering
    \includegraphics[width=\textwidth]{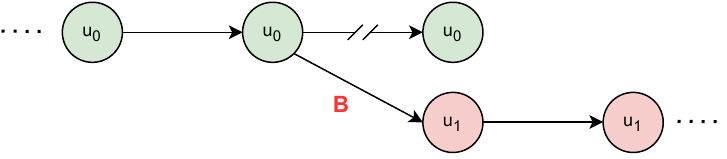}
    \caption{Example of altered RM state dynamics arising from an attacker \emph{injecting} an arbitrary high-level event --- denoted by the red label --- to the victim's reward machine in the Simple Cookie World environment. Green nodes represent the true reward machine states, while red nodes denote the attacker-induced ones.}
    \label{fig:lf_mirage}
\end{figure}

As a quick summary of all the decisions that were taken up to the definition of the target, the proposed attacks:
\begin{itemize}
    \item aims at hindering the pre-trained agent's performance;
    \item by manipulating its labeling function to alter the reward machine's state dynamics;
    \item assuming access to the least amount of information possible;
    \item attempting to minimize the number of actions required from the attacker.
\end{itemize}
The methodology that I initially considered consists of the attacker injecting arbitrary sets of high-level events \emph{in place} of the correct labeling function outputs. By doing so, he could, in theory, induce the agent's RM into transitioning to any of the states reachable by its current one. For instance, in the Simple Cookie World example, the attacker could lead the agent to believe that the cookie is available, \ie induce its RM to transition from $u_0$ to $u_1$ --- by \emph{falsely} informing him that the button has been pressed, as depicted in figure \ref{fig:lf_mirage}. This type of attack is, in theory, very powerful, as it allows one to control the evolution of the agent's reward machine arbitrarily and, consequently, its policy. However, being able to achieve this in practice is far from trivial for a number of reasons. 

Firstly, it requires the attacker to know \emph{what} and \emph{when} he must inject to obtain the desired effect: providing the agent with a set of high-level events that do not trigger any transition from its current state has no effect other than creating noise that could aid the victim in detecting the attempted attack. Thus, the attacker needs to either be \emph{very} lucky and indifferent to the possibility of being detected or, more realistically must be armed with in-depth knowledge regarding the victim's reward machine. Secondly, even if we assume the attacker to be able to trigger any arbitrary transition at will, a second problem is quick to emerge: How can he determine which transition would lead to the greatest decrease in the victim's performance? While simple scenarios might allow for some speculation in this regard, any sufficiently complex environment would lead to proportionally sophisticated agent behavior that would likely render such analysis practically impossible. Even in case the attacker had control over a \emph{complete} copy of the agent, allowing him to simulate every possible attack route, he is faced with the task of choosing the best path --- in terms of worst agent performance --- among all the possible ones that exist in the victim's RM. On the one hand, the number of such paths might even be infinite in the presence of any loop in the reward machine's structure. On the other hand, even assuming their number is finite and small enough to make the task tractable in theory, the huge computational costs arising from such simulations would cause them to become almost infeasible in practice, even for highly motivated adversaries.

In light of the above considerations, this class of attacks --- for which I propose the name of \emph{hallucination attacks}\footnote{This choice was made in line with the common practice in cybersecurity research of finding picturesque names for novel attack techniques.} --- is as impactful in theory as it is hard to carry out effectively in practice. Therefore, motivated by my original objectives, I decided to adopt a simpler attack methodology that could be reasonably effective while also eliminating most of the described problems. Interestingly, the strategy that I ultimately devised originates from an apparently minor change in the previous approach: instead of trying to hinder the victim's performance by leading him into following bad trajectories in its reward machine's state space, my attack tries to achieve the same goal by disturbing the agent's path along the legitimate ones. In other words, instead of injecting an arbitrary set of events, the attacker exclusively \emph{deprives} the victim of some of its intended labeling function outputs, thus leading to the name of \textbf{blinding attacks}. This choice is heuristically motivated by the assumption that simply depriving the agent of information it was trained to rely on may be enough to cause him to reduce its proficiency drastically. Interestingly, the results that I obtained by experimentally evaluating my final attack design give this assumption some credit, as will be discussed in chapter \ref{chap:results}.

It is important to note that blinding attacks represent a sub-class of the more general set of hallucination attacks. Indeed, in formal terms, given a set of high-level propositions $\propsSet$, any hallucination attack can be represented by a \emph{tampering function}:
\[
    \tau_{\mathcal{A}} : \mathbb{N} \times 2^\propsSet \rightarrow 2^\propsSet
\]
mapping the agent's labeling function output at every timestep to an arbitrary set of high-level events. Similarly, any blinding attack can be represented by a function:
\begin{align*}
    &\tau_{\mathcal{B}} : \mathbb{N} \times 2^\propsSet \rightarrow 2^\propsSet \quad  \text{s.t.}\\
    &\tau_{\mathcal{B}}(t, \sigma) \in 2^\sigma \quad \forall t \in \mathbb{N},\ \forall \sigma \in 2^\propsSet
\end{align*}
where at each timestep, the labeling function output is replaced with a --- possibly empty --- subset of the events originally composing it.
Therefore, the relationship between the set $T_{\mathcal{A}}$ of all possible hallucination attacks and the set $T_{\mathcal{B}}$ of all possible blinding attacks is easily expressed:
\[
    T_{\mathcal{B}} = \{\tau\ |\ \tau \in T_{\mathcal{A}}\ \land\ \tau(t, \sigma) \in 2^\sigma\ \forall t \in \mathbb{N},\ \forall \sigma \in 2^\propsSet \} \subset T_{\mathcal{A}}
\]

\begin{figure}
    \centering
    \includegraphics[width=\textwidth]{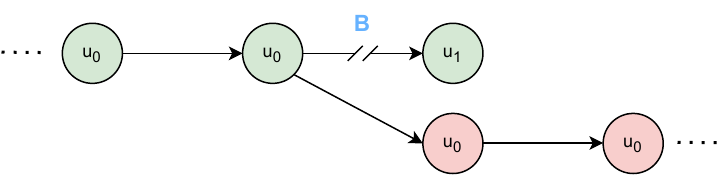}
    \caption{Example of altered RM state dynamics arising from an attacker \emph{dropping} an arbitrary, legitimate labeling function output --- denoted by the blue label --- in the Simple Cookie World environment. Green nodes represent the true reward machine states, while red nodes denote the attacker-induced ones.}
    \label{fig:lf_blind}
\end{figure}

From an attack's impact point of view, blinding attacks can still achieve the same \emph{qualitative} results as hallucination attacks, leading to a desynchronization between the victim's reward machine state and the actual state of the environment. For instance, once again resorting to the Simple Cookie World example, the attacker could induce the victim into not realizing that the cookie is available --- \ie prevent its RM from transitioning from $u_0$ to $u_1$ --- by depriving him of the labeling function output communicating that the button has been pressed, as depicted in figure \ref{fig:lf_blind}. Thus, the main difference between the two approaches lies in the fact that blinding attacks allow for a reduced number of inducible RM state trajectories: instead of allowing the attacker to potentially lead the victim into following any valid trajectory, blinding attacks can only induce the victim's RM state to follow any trajectory that can be obtained by preventing an arbitrary number of transitions from triggering. Nonetheless, this reduction in possibilities might not be enough to render the computation of the worst-performing trajectory feasible, due to the same complexities that were discussed in the case of hallucination attacks. However, by dropping the search for the optimal attack strategy altogether, the limited amount of available attack paths suddenly becomes highly valuable, as it allows the attacker only to consider a restricted number of alternatives, which can then be compared according to a simple set of heuristic criteria to determine the most promising one. Ultimately, this is the approach that characterizes the final design of the attacks that I will discuss in the following sections.

\section{Blinding attacks on reward machines}
While the first half of this chapter presented an accurate outline of the main concepts, objectives, and assumptions motivating the high-level design of the attacks that I developed, this section finally discusses all the details that still need to be defined to obtain a practical, effective implementation of \textbf{blinding attacks}. The concepts hereby discussed are presented in \emph{chronological} order, starting from the first, basic techniques that I developed and progressively introducing the various refinements that led to the, more sophisticated, final approaches.

\subsection{Random noise attacks}
I developed the first type of blinding attack to be as naive as possible. In this way, I tried to gain a preliminary understanding of the approach's overall validity, before transitioning to the design of more complex techniques that could later turn out to be badly motivated in the first place. The underlying idea is straightforward: if, by resorting to a very simple attack plan, the adversary is able, at least, to induce a moderate decrease in the victim's performance, then this can be evidence supporting the possibility of more sophisticated strategies being more effective and, thus, worth investigating. Any failure in this regard, while not hard evidence of the intrinsic ineffectiveness of the approach, could motivate a deeper investigation of its foundational ideas.

Therefore, the first type of blinding attack I propose is represented by \textbf{random-blinding attacks}, where an attacker operates by randomly depriving its victim of the actual labeling function outputs with a given probability. The attacker may drop any labeling function output entirely or, alternatively, drop just a subset of the original events, again depending on chance. Thus, it is evident how this attack plan represents the most naive one possible, in that, in fact, it does not represent a plan at all: the attacker simply acts randomly and hopes for the best. Algorithm \ref{alg:rand-blind} formalizes the described approach and allows for two observations to be made regarding the requisites the attacker must satisfy for the attack to be feasible. In terms of knowledge, the attack is entirely \emph{black-box}, as it requires no information relating to the agent's inner workings or labeling function structure\footnote{Recalling the note that was previously made regarding the debatability of blinding attacks black-box classification, this arises depending on whether one considers the labeling function to be part of the agent or not. Therefore, the proposed characterization might vary depending on the implementation details of the agent under attack. In the worst case, the attack could be considered a gray-box attack with very mild information requirements.}. On the other side, in terms of capabilities, the attacker only needs to be able to interpose himself between the labeling function and the victim, thus making it an instance of a \emph{man-in-the-middle} attack.

\begin{algorithm}
\caption{Pseudocode for random-blinding attacks to a given RM-based agent $\mathcal{A}$ associated with a set of high-level events $\propsSet$ and labeling function $L$, using an attack probability of $\rho$.}
\label{alg:rand-blind}

    \begin{algorithmic}
        \Function{Random-Blinding}{$\mathcal{A}, \propsSet, L, \rho$}

        \State
        \Repeat
            \State $\sigma \gets$ \Call{Intercept-Output}{$L$}
            
            \State
            \State$blind? \gets $ \Call{Random}{$0, 1$}
            \If{$blind? < \rho$}
                \State $\Tilde{\sigma} \gets$ \Call{Random-Choice}{$2^\sigma - \{\sigma\}$}
                \State $\sigma \gets \Tilde{\sigma}$
            \EndIf

            \State
            \State \Call{Forward-Events}{$\mathcal{A}, \sigma$}

        \Until \Call{Is-Done}{$\mathcal{A}$}
        
        \EndFunction
    \end{algorithmic}
    
\end{algorithm}

\subsection{Choice of blinding strategy}
\label{sec:blinding_strats}
Motivated by the promising results --- thoroughly discussed in chapter \ref{chap:results} --- obtained during the experimental evaluation of random-blinding attacks, I refined the approach by devising a more sophisticated method to compute effective attack plans. This requires determining two things: \emph{which events} the victim must be deprived of, and \emph{when} to carry out such removals. I will refer to the former choice as the \textbf{blinding strategy} and the latter as the \textbf{timing strategy}. This section presents how an attacker might identify potential blinding strategies, leaving the discussion of timing strategies as the topic for the following section.

The method I designed for determining the \emph{most promising} blinding strategies available to the attacker is based on two main objectives:
\begin{enumerate}
    \item maintaining the black-box quality of random-blinding attacks;
    \item being reasonably efficient for practical instances of the attack.
\end{enumerate}
Thus, to facilitate their achievement, I resorted to a \emph{passive} approach requiring the attacker to first \emph{observe}, without intervening, a given number of labeling function outputs. By doing so, he can collect a set of valid targets for a potential blinding strategy. Eventually, given a sufficient number of observations and assuming the set $\propsSet$ of high-level events is finite, the agent is guaranteed to converge to the set of all possible labeling function outputs. Therefore, if enough time to observe its victim is available, the attacker can be sure not to overlook any potential strategy, with the consequent risk of missing out on some effective way of carrying out the attack. On the other hand, when faced with small windows of opportunity, the attacker can nonetheless gather as many alternatives as possible and use them to carry out an attack to the best of its possibilities.

After collecting a given number of potential blinding strategies, the attacker must then decide which one is the most promising for inducing the greatest decrease in the victim's performance and, thus, the one that will be used when carrying out the attack. As argued in section \ref{sec:attack_methodologies}, devising an algorithm to determine the provably optimal strategy in this regard is both extremely complex and demanding in terms of computational resources, thus making this option conflict with my design goals. Therefore, I discarded this possibility and opted for a simpler approach aimed at providing the attack with reasonably good performance in practice. Specifically, the method I propose consists of sorting the identified strategies according to three \emph{heuristic criteria} and then choosing the resulting top-ranked strategy, leading to the overall approach outlined in algorithm \ref{alg:strategy_choice}.

In order of priority, these criteria are:
\begin{enumerate}
    \item \textbf{reliability heuristic}, indicating a preference for the events that were seen at least once in a larger number of episodes;
    \item \textbf{early-appearance heuristic}, indicating a preference for events observed early during the agent's trajectory;
    \item \textbf{rare-event heuristic}, indicating a preference towards the events that were observed a lower number of separate times.
\end{enumerate}
Each of the above heuristics is aimed at summarizing one desirable property that good blinding strategies can be reasonably expected to possess. In light of this, the \emph{reliability} heuristic is designed to prioritize the strategies that are less likely to leave the attacker with no labeling function output to remove. For instance, consider an agent acting in an environment characterized by two possible high-level events: a common event $\ao{A}$, characterized by a $90\%$ chance of being generated during any given episode, and a rare event $\ao{B}$, only seen in $10\%$ of the episodes. From the perspective of the attacker, choosing $\ao{B}$ as its target is an evidently bad choice, as it would leave the attacker with nothing to do in 9 out of 10 attempted attacks, thus making them a complete failure. However, on the other hand, it could also be argued, under an information-theoretic interpretation,
that rarer events hold a greater amount of information for the victim than commonly seen ones. Indeed,
the \emph{rare-events} heuristic is designed to also account for this interpretation by leaning towards the events that are characterized by a lower absolute frequency, computed among all the observed victim's episodes. Finally, the \emph{early-appearance} heuristic is designed under the assumption that the earlier the attacker is able to cause a desynchronization between the agent's RM and the true state of the environment, the more damage is likely to arise from such discrepancy. Thus, this last criterion introduces an inclination toward the events that, in each observed episode, appeared during earlier time steps.

There is a possibly unlimited number of arguments that could be made in favor of each of the presented heuristics and their design, and perhaps even more could be constructed against all three of them. The solution to this apparently non-settleable debate lies in a simple concept: due to their heuristic nature, all the proposed criteria have the potential to both be good and bad, heavily depending on the exact context at hand. Therefore, the technique I propose is to be interpreted as a practical declination of the more general approach consisting of heuristically choosing the most promising blinding strategy. For instance, in certain environments, the rare-event heuristic might work best if used as the primary sorting criterion, while in others it may be completely useless. Similarly, there is no strict requirement for the use of a multi-step sorting strategy: one could, for instance, use a linear combination of $n$ criteria:
\[
    H(\sigma) = \sum_{i=1}^n \alpha_i h(\sigma) \quad \forall \sigma \in 2^\propsSet
\]
weighted according to a set of arbitrarily chosen weights $(\alpha_1, ..., \alpha_n)$.

\begin{algorithm}
\caption{Pseudocode for determining the most promising blinding target for attacking an RM-based agent $\mathcal{A}$ associated with a labeling function $L$, after observing $k$ of its outputs.}
\label{alg:strategy_choice}

    \begin{algorithmic}
        \Function{Choose-Blinding-Strategy}{$\mathcal{A}, L, k$}
        \Local
            \State $h_1[\sigma]$: values for the reliability heuristic, zero-initialized;
            \State $h_2[\sigma]$: values for the early-appearance heuristic, zero-initialized;
            \State $h_3[\sigma]$: values for the rare-event heuristic, zero-initialized;
        \EndLocal

        \State $targets \gets \emptyset$
        \State $observations \gets 0$
        \While{$observations < k$}

            \State
            \For{$\sigma \in targets$}
                \State $seen[\sigma] \gets false$
            \EndFor

            \State $t \gets 0$ \Comment Start of a new episode
            \Repeat
                
                \State $\sigma \gets$ \Call{Intercept-Output}{L}
                \State $observations \gets observations + 1$
                \State $t \gets t + 1$

                \State \Comment Update targets and heuristics
                \State $targets \gets targets\,\cup \{\sigma\}$
                \State $h_3[\sigma] \gets h_3[\sigma] + 1$
                \If{$\lnot\, seen[\sigma]$}
                    \State $seen[\sigma] \gets true$
                    \State $h_1[\sigma] \gets h_1[\sigma] + 1$
                    \State $h_2[\sigma] \gets$ \Call{Min}{$h_2[\sigma], t$}
                \EndIf

                \State \Call{Forward-Events}{$\mathcal{A}, \sigma$}

            \Until \Call{Is-Done}{$\mathcal{A}$}

        \EndWhile

        \State \Comment Sort targets based on the heuristics
        \State \Call{Stable-Ascending-Sort}{$targets, h_3$}
        \State \Call{Stable-Ascending-Sort}{$targets, h_3$}
        \State \Call{Stable-Descending-Sort}{$targets, h_1$}

        \Return \Call{First}{$targets$}

        \EndFunction
    \end{algorithmic}
    
\end{algorithm}

\subsection{Timing strategies}
\label{sec:timing_strats}
Once the blinding strategy has been defined, the last decision the attacker must make relates to the \textbf{timing strategy} to be used for carrying out the attack. This is needed because, in general, there is nothing preventing any set of high-level events from being produced more than once during a given episode. Therefore, the attacker might, for instance, decide that the best available course of action consists of depriving its victim of the first, third, and fifth instances of the target labeling function output. Like in the case of the blinding strategy choice, determining the provably optimal solution for this problem is extremely complex, requiring a huge computational effort and, importantly, a deep violation of the black-box assumption. In light of this, I discarded the prospect of determining the optimal ones and, instead, experimented with three simple heuristic timing strategies. 

The first one is the \textbf{all-instances} timing strategy, where the attacker deprives the victim of \emph{every instance} of the blinding target, effectively preventing it from receiving such events for the whole duration of the attack. This has two main consequences: On the one hand, the impact of the attack may be very high, due to the huge amount of information hidden from the victim. On the other hand, this type of strategy might allow for easier detection from the victim's side, since it requires a large number of interventions from the attacker.

The second option I considered consists of the \textbf{first-stream} timing strategy. In this case, the attacker deprives the victim of the \emph{first contiguous stream} of occurrences of the target. In other words, the target labeling function output is hidden from the victim on its first occurrence; then, if the following LF output is different from the target, the attack ends. Otherwise, the attack continues until the labeling function output switches from the target to any other set of events. Instead of obscuring just on its first occurrence, I preferred this solution as it allows for a more robust attack that avoids being immediately rendered useless by the target showing up twice in a row, as might be possible in a number of different scenarios. Indeed, by being denied a whole contiguous stream of occurrences, the victim might be more likely to move to a different portion of the environment state space where the target events can't be generated. However, it must be noted that this assumption is purely speculative in nature, and might not actually represent the outcome arising from the use of this type of timing strategy.

Finally, the third strategy I designed is the \textbf{triggered-stream} timing strategy, which consists of a variation over the first-stream one. Indeed, it is based on the same stream-oriented approach, in that it too deprives the victim of a contiguous sequence of identical labeling function outputs. However, instead of activating upon the first such stream, every time one is encountered, there is a given probability for the attack to begin. Then, after the stream of events that triggered this strategy
is exhausted, the attack terminates. Therefore, this timing strategy could be seen as a way of relaxing its parent strategy's stone-hard preference for the first stream: By tuning the trigger chance, one can, in the case of a high trigger chance, introduce a \emph{bias} towards the elimination of early streams or towards later ones for low trigger probabilities\footnote{This can be shown by observing that the probability of the strategy triggering after k occurrences follows a geometric distribution: $P(X = k) = p(1-p)^{k-1}$, where $p$ is the probability of any occurrence triggering the attack. Therefore, the expected attack start time is $1/p$, which grows as $p$ decreases and vice-versa.}. Thus, this timing strategy allows for a wider set of attack paths to be considered. However, the downside of this approach lies in an increased chance of the attack not obscuring any labeling function outputs in the unfortunate case the attack was never to trigger.

\subsection{Event-blinding attacks}
After discussing how an attacker can identify the most promising blinding strategy, and a number of timing strategies available to him, I can finally connect all the presented concepts and describe the general framework of \textbf{event-blinding attacks}. Indeed, doing so is especially easy since the reader is already familiar with all the ideas that make up such attacks. 

The preliminary step of any event-blinding attack consists of the choice of the blinding and timing strategies to be followed. After these are established, the actual attack can begin: for every labeling function output the attacker intercepts, he consults his blinding strategy to determine if the observed set of events corresponds to one of the targets of the attack. If this is the case, the attacker then consults his timing strategy to decide if the victim should be deprived of this particular instance of the target. If this is, once again, determined to be the case, the attacker tampers with the labeling function output and forwards it to the victim, starting a new iteration until either the episode ends or the timing strategy determines the attack to be over. 

Algorithm \ref{alg:evt_blind} summarizes the outlined method, highlighting its modular design: the framework can be seamlessly adapted to a number of different contexts by providing a different implementation for any of the functions in the pseudocode. For instance, in section \ref{sec:timing_strats} I described three different approaches for implementing the \textproc{Should-Blind?} function, while in section \ref{sec:blinding_strats} I outlined a method for selecting a promising blinding strategy. In that context, for the sake of simplicity, I implicitly assumed each attack to be characterized by a single target the attacker wants to deprive the victim of. However, in general, it needs not to be the case: a blinding strategy might consist of multiple targets, time-dependent targets, or any variation one could come up with, all leading to different implementations for the \textproc{Is-Target?} function. Similarly, there is no hard requirement for a given labeling output to be fully obscured: the attacker might want to simply remove one of the events composing it, for instance, to reduce the chance of being detected. By producing an appropriate implementation of the \textproc{Remove-Events} function, one could adopt any such mechanism.

In light of this, it is immediately evident how event-blinding attacks allow for a wide array of variations to be defined over this basic framework. Indeed, I experimented with two such variations, namely \textbf{compound} event-blinding attacks and \textbf{atomic} event-blinding attacks. The former type consists of the attacks that I implicitly described until now, where the attacker treats each labeling output \emph{as a whole} as a potential target, and obscures it \emph{completely} when needed. On the other side, \emph{atomic} event-blinding attacks consider each atomic proposition $p \in \propsSet$ as a potential target and thus, tamper with the labeling function outputs in a more subtle manner, by removing
only the desired targets and forwarding the remaining events unaltered. Despite this minor difference, all the techniques that I described up until this point can seamlessly be adapted to the context of atomic event-blinding attacks.

\begin{algorithm}
\caption{Pseudocode for \emph{event-blinding attacks} to a RM-based agent $\mathcal{A}$ and its associated labeling function $L$.}
\label{alg:evt_blind}

    \begin{algorithmic}
        \Function{Event-Blinding}{$\mathcal{A}, \propsSet, L, k$}

            \State 
            \State \Comment Determine strategies
            \State $\mathcal{S}_B \gets$ \Call{Choose-Blinding-Strategy}{$\mathcal{A}, L$}
            \State $\mathcal{S}_T \gets$ \Call{Choose-Timing-Strategy}{$\mathcal{A}, L$}

            \State \Comment Actual attack
            \Repeat
                \State $\sigma \gets$ \Call{Intercept-Output}{L}

                \State
                \If{\Call{Is-Target?}{$\mathcal{S}_B, \sigma$}}
                    \State $blind?$, $done? \gets$ \Call{Should-Blind?}{$\mathcal{S}_T, \sigma$}
                    \If {$blind?$}
                        \State $\Tilde{\sigma} \gets$ \Call{Remove-Events}{$\mathcal{S}_B, \sigma$}
                        \State $\sigma \gets \Tilde{\sigma}$
                    \EndIf
                \EndIf
                \State

                \State \Call{Forward-Events}{$\mathcal{A}, \sigma$}

            \Until $done?\ \lor$ \Call{Is-Done?}{$\mathcal{A}$}

        \EndFunction
    \end{algorithmic}
    
\end{algorithm}

\subsection{Edge-blinding attacks}
Up until this point, all the attacks that I described are characterized by the black-box assumption, which is highly convenient for assuring their practical feasibility under a wide range of conditions. However, from a theoretical point of view, relaxing such an assumption might allow one to discover novel attack techniques that, although less applicable in practice, are noticeably more effective due to their access to a higher amount of information. The last class of attacks that I developed, \textbf{edge-blinding attacks} --- for which algorithm \ref{alg:edg_blind} provides a pseudocode implementation --- originates from this exact observation. Their name comes from the key difference that separates them from event-based attacks: the attacker, instead of trying to deprive his victim of its labeling function outputs, now explicitly targets his victim's RM state transitions, \ie its \emph{edges}, with the aim of preventing them from triggering. In the case of event-based attacks, this still happens, but it is a \emph{consequence} of the attacker's action, not his actual objective. In other words, during an event-based attack, the adversarial actor chooses one or more target events and removes them from any labeling function outputs they appear in. On the other side, in the case of an edge-blinding attack, the attacker chooses one or more of the victim's reward machine state transitions and removes any event leading to any of them triggering from the labeling function outputs. 

Despite this difference, all the timing strategies proposed in section \ref{sec:timing_strats} remain perfectly applicable in the case of edge-blinding attacks. Moreover, even the method that I described for determining the most promising blinding strategy also works in the case of edge-based attacks, with just a few minor differences: To be able to target a \emph{specific} transition, the attacker, at each timestep, needs access to an additional piece of information: the victim's current reward machine state. From a theoretical point of view, due to the victim's RM being a piece of information regarding its inner structure, edge-blinding attacks violate the black-box assumption and should therefore be classified as gray-box attacks. Nonetheless, in practice, the heightened standard that they pose in terms of the attacker's knowledge might not represent an excessively demanding requirement to satisfy. In any case, the main motivation behind the design of edge-blinding attacks lies in the fact that, in theory, they can achieve the same effects as event-based ones --- \ie the desynchronization between the victim's RM state and the true state of the environment --- while requiring a lower number of interventions from the attacker: Since event-based attacks do not take into consideration whether or not depriving an agent of a given set of events would actually prevent its RM from undergoing a state transition, they risk generating a lot of noise before achieving their objective. On the other side, in the case of edge-based attacks, the adversary only tampers with the labeling function outputs when he knows that doing so would cause the victim's RM to desynchronize, thus introducing a lower amount of noise that could be used by the victim to detect the attacker.

\begin{algorithm}
\caption{Pseudocode for \emph{edge-blinding attacks} to a RM-based agent $\mathcal{A}$ and its associated labeling function $L$.}
\label{alg:edg_blind}

    \begin{algorithmic}
        \Function{Edge-Blinding}{$\mathcal{A}, \propsSet, L, k$}

            \State 
            \State \Comment Determine strategies
            \State $\mathcal{S}_B \gets$ \Call{Choose-Blinding-Strategy}{$\mathcal{A}, L$}
            \State $\mathcal{S}_T \gets$ \Call{Choose-Timing-Strategy}{$\mathcal{A}, L$}

            \State \Comment Actual attack
            \Repeat
                \State $u \gets $ \Call{Observe-RM-State}{$\mathcal{A}$}
                \State $\sigma \gets$ \Call{Intercept-Output}{L}

                \State
                \If{\Call{Is-Target?}{$\mathcal{S}_B, (u,\sigma)$}}
                    \State $blind?$, $done? \gets$ \Call{Should-Blind?}{$\mathcal{S}_T, (u,\sigma)$}
                    \If {$blind?$}
                        \State $\Tilde{\sigma} \gets$ \Call{Remove-Events}{$\mathcal{S}_B, \sigma$}
                        \State $\sigma \gets \Tilde{\sigma}$
                    \EndIf
                \EndIf
                \State

                \State \Call{Forward-Events}{$\mathcal{A}, \sigma$}

            \Until $done?\ \lor$ \Call{Is-Done?}{$\mathcal{A}$}

        \EndFunction
    \end{algorithmic}
    
\end{algorithm}

Finally, as in the case of event-based attacks, I also developed and experimented with a variation on the basic edge-based approach: \textbf{state-blinding attacks}. Despite the name, state-based attacks are regular edge-based attacks that target more than one transition. Specifically, they target all the transitions entering into a given victim's RM state. However, targeting every transition leading to a specific state could, equivalently, be seen as targeting the state itself, hence the name of the technique.
\chapter{Evaluation of the proposed attacks}
\label{chap:results}

\renewcommand{\arraystretch}{1.5}

Due to the substantial number of techniques and algorithms that I designed during my work, carrying out an extensive experimental evaluation of their effectiveness was fundamental for determining the degree of risk that they could pose to practical deployments of RM-based reinforcement learning. In light of this, this chapter describes all the steps that I took to carry out such an evaluation and the results that I obtained. Specifically, the first half of this chapter provides a description of the experimental setting I prepared for conducting the evaluation, starting from the choice of the benchmark environments, followed by the algorithms used to train the victim agents and the results of the training sessions, and concluding with the definition of the testing sessions that were carried out along with the rationale behind each of them.
Then, the second half of the chapter provides the reader with a critical analysis of all the numerical results that were obtained for each testing session, aimed at highlighting the instances where the proposed attacks were successful in achieving their goals, those where they weren't and the reasons behind each of these situations. 

\section{Experimental setting}
The definition of an adequate setting is of paramount importance in allowing the experimental evaluation of the proposed techniques to lead to results that are both representative and insightful. In this regard, every decision I took was motivated by a careful analysis of the available alternatives. However, since any given design for an experimental evaluation cannot reasonably be expected to be perfectly representative of every possible situation one might be faced with in practice, I sometimes had to make a compromise between conflicting objectives. Thus, in this section, I present the rationale behind all the decisions that led me to the setting under which I ultimately conducted the evaluation of every attack technique I propose.

\subsection{Benchmark domains}
\label{sec:domains}
The first, and possibly most important, aspect I needed to define relates to the choice of the benchmark environments. Indeed, a poor choice in this regard could lead to a number of different problems, a prominent one being the inability of RM-based training algorithms to converge to optimal policies. This scenario would make any investigation on the effectiveness of my attacks intrinsically flawed, as the lack of proficiency on the agents' side would likely play a substantial role in leading to any potentially high impact of the attacks not being due to their effectiveness. Therefore, the benchmark environments should not pose too hard of a challenge for the agents to be able to learn how to deal with them adequately. Moreover, even assuming the agents could converge if highly complex environments were to be picked, such a choice would considerably increase the computational effort required for both training the agents and testing the attacks, thus introducing a practical limitation for the amount of experimentation that I could carry out and, consequently, a reduction in its overall quality. On the other side, a choice of excessively simple environments would be problematic as well, due to the risk of the results being inadequate in representing a valid proxy for real-world deployments of RM-based reinforcement learning. In light of these considerations, I
decided to resort to \emph{partially observable} environments instead of fully observable ones. This choice allowed me to exploit the full learning potential of RM-based techniques, while also introducing one of the most prominent sources of complexity that characterize real-world tasks.

Another factor that I took into consideration in the choice of the environments is their popularity in currently available RL literature, as relying on widely used benchmarks allows one to resort to similar research work as a meter for comparison. Unfortunately, it appears that, in its current state, partially observable reinforcement learning research suffers from a lack of standard benchmarks. Indeed, most publications rely on ad-hoc environments or, in the best case, on the ones proposed by the original authors of the techniques under scrutiny. Nonetheless, I was able to identify one notable effort towards the standardization of partially observable benchmarks in \cite{morad_popgym_2023}. In their work, \citeauthor{morad_popgym_2023} propose a diverse set of environments designed to be a testbed for assessing the memory capacity of reinforcement learning models in partially observable environments. In this regard, despite its currently limited adoption, this set of benchmarks appeared like a solid choice to carry out my experimentation. Unfortunately, a preliminary investigation of the effectiveness of RM-based techniques in such environments quickly revealed their intrinsic inadequacy for the type of tasks composing the library. Specifically, those tasks require the agent to remember information that is inconvenient to represent with the language used by reward machines: propositional logic\footnote{This problem is also present, in a limited form, in one of the benchmarks that I ultimately chose: Symbol World. During its presentation, I will further elaborate on this issue.}. Therefore, I had to resort to a different set of benchmark environments, and I ultimately settled with using the three environments proposed by \citeauthor{icarte_lrm_2019} in \cite{icarte_lrm_2019}: Cookie World, Keys World, and Symbol World.

\begin{figure}
\label{fig:domains_map}
    \centering
    \includegraphics{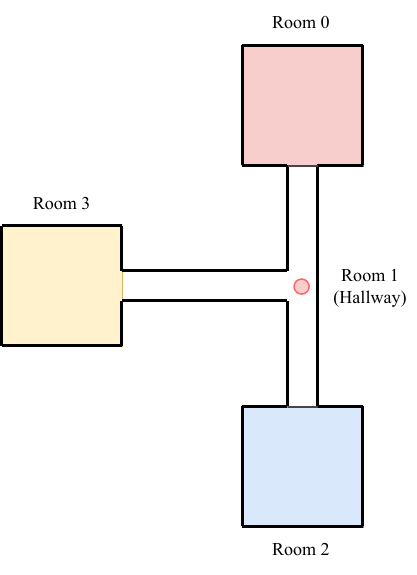}
    \caption{Map of the environment for the three benchmark domains. The red dot represents the agent in its initial position.}
\end{figure}

These are all grid-world environments based on the same map, depicted in figure \ref{fig:domains_map}. Each environment is characterized by a different task that the agent can solve by moving between the three rooms and the hallway connecting them. Despite their lack of widespread adoption, the choice of these environments presents a number of desirable properties. First, as shown by \citeauthor{icarte_lrm_2019}, their associated tasks could not be solved by state-of-the-art RL algorithms\footnote{Namely, Double DQN (DDQN), Asynchronous Advantage Actor-Critic (A3C), Actor-Critic with Experience Replay (ACER) and Proximal Policy Optimization (PPO).} equipped with LSTM-based memory architectures in a reasonable amount of training steps. This showcases how these environments represent a considerable challenge for RL techniques. Secondly, thanks to the availability of the results from \citeauthor{icarte_lrm_2019}, they allow for the results obtained by training the agents to be easily validated to confirm convergence to optimal policies. Finally, their conceptually simple-for-humans tasks allow for an easy interpretation of the reward machines that will be used, in terms of both structure and dynamics. 

On the other hand, a couple of issues can also be identified: On the one hand, they are likely too simple environments to represent a good proxy for real-world applications of RM-based reinforcement learning. This could limit the generalization potential for the observations arising from the analysis of the obtained results. On the other hand, they are also characterized by a set of tasks that do not widely differ from each other and, thus, might not be fully representative of all the possible situations that an attacker might be faced with.

Before delving into the tasks and perfect reward machines associated with each environment, a few notes must be made regarding their commonalities. In terms of high-level events, the set of propositions $\propsSet$ for each environment consists of the union of two subsets: the first containing task-specific events, and the second being:
\[
    \propsSet_r = \{ \ao{0}, \ao{1}, \ao{2}, \ao{3} \}
\]
These events, generated on \emph{every} timestep, indicate the number of the room the agent is currently in, according to the numbering scheme from figure \ref{fig:domains_map}. Therefore, $\ao{0}$ indicates that the agent is currently in the red room, $\ao{1}$ corresponds to the hallway, $\ao{2}$ to the blue room, and $\ao{3}$ to the yellow one. The partial observability of each environment arises from the agent being provided, on each percept, only with the content of the room it is currently in. Finally, each environment is characterized by a $10\%$ probability of any action, \ie agent movement, leading to a random outcome. This is a common strategy used when conducting RL research in grid-world-like environments that allows for the introduction of a degree of stochasticity in the environment, thus increasing the level of challenge, it poses to the agent and decreasing the possibility for it to just memorize the path to its objective.

\subsubsection{Cookie World}
\begin{figure}
\label{fig:cw_rm}
\centering

    \begin{tikzpicture}[initial text={}, >=stealth, auto, node distance=3cm, on grid, initial where=left]

        \node[state,initial] (0) at (0,0) {$u_0$};
        \node[state]         (1) at (0,2) {$u_1$};
        \node[state]         (2) at (4,3) {$u_2$};
        \node[state]         (3) at (-4,3) {$u_3$};
        \node[state,accepting]         (4) at (0, 6) {$u_4$};

        \path[->] (0) edge  node  [right]  {$\rmEdge{\ao{3B}}$} (1);
        
        \path[->] (1) edge  [bend right] node  [below right]  {$\rmEdge{\ao{0c}\I\ao{2}}$} (2);
        \path[->] (2) edge  [bend right] node [above left]   {$\rmEdge{\ao{3B}}$} (1);
        
        \path[->] (1) edge  [bend left] node  [below left]  {$\rmEdge{\ao{2c}\I\ao{0}}$} (3);
        \path[->] (3) edge  [bend left] node  [above right]    {$\rmEdge{\ao{3B}}$} (1);

        \path[->] (3) edge  [bend left] node [above left]  {$\rmEdge{\ao{2C},\rwP{+1}}$} (4);
        \path[->] (2) edge  [bend right]  node [above right] {$\rmEdge{\ao{0C},\rwP{+1}}$} (4);
        
    \end{tikzpicture}

\caption{Perfect reward machine for the Cookie World benchmark domain.}
\end{figure}
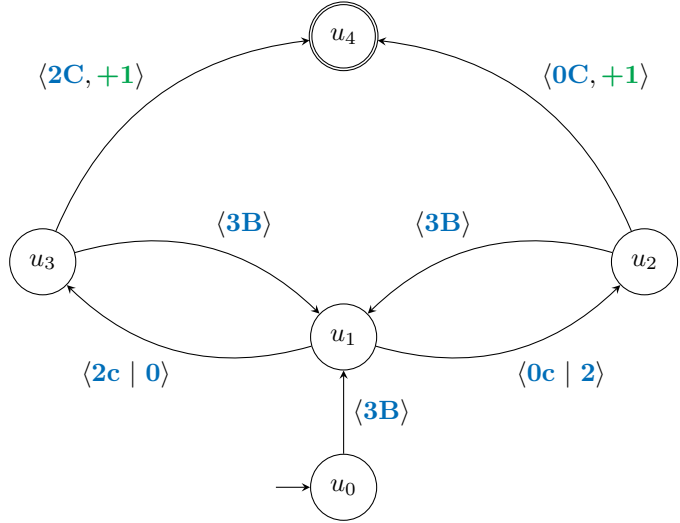

Unsurprisingly, the Cookie World environment is completely analogous to the Simple Cookie World example from Chapter \ref{chap:rms}, with the only difference being the map where the agent acts, which now has three rooms instead of two. This aside, the task is the same: starting from the hallway, the agent must reach the button inside room 3 and press it. Once the button has been pressed, a cookie randomly appears in either room 2 or room 0. If the agent presses the button again, the previous cookie disappears, and a new cookie is created, again, randomly, in either room. The agent succeeds in its task after eating the cookie. For doing so, the agent receives a reward of $\rwP{+1}$. Figure \ref{fig:cw_rm} presents the \emph{perfect} reward machine that was supplied to each agent, which is based on the following set $\propsSet_{cw}$ of high-level events:
\[
    \propsSet_{cw} = \propsSet_r \ \cup \ \{ \ao{b}, \ao{c}, \ao{B}, \ao{C} \}
\]
where $\ao{b}$ and $\ao{c}$ inform the agent that he is currently in the same room as, respectively, the button or the cookie. Similarly, $\ao{B}$ and $\ao{C}$ represent, respectively, that the agent has pressed the button or eaten the cookie. The reward machine from figure \ref{fig:cw_rm} can now easily be interpreted:
\begin{itemize}
    \item $u_0$: the agent must press the button;
    \item $u_1$: the cookie is present, but the agent must discover its location;
    \item $u_2$: the agent knows the cookie is in room 0, either because it saw it directly -- \ie $\ao{0c}$ --- or because it did not see it in room 2 --- \ie $\ao{2}$;
    \item $u_3$: the agent knows the cookie is in room 2, with considerations analogous to those presented for $u_2$;
    \item $u_4$: the agent has eaten the cookie and, thus, the episode is over.
\end{itemize}

\subsubsection{Keys World}
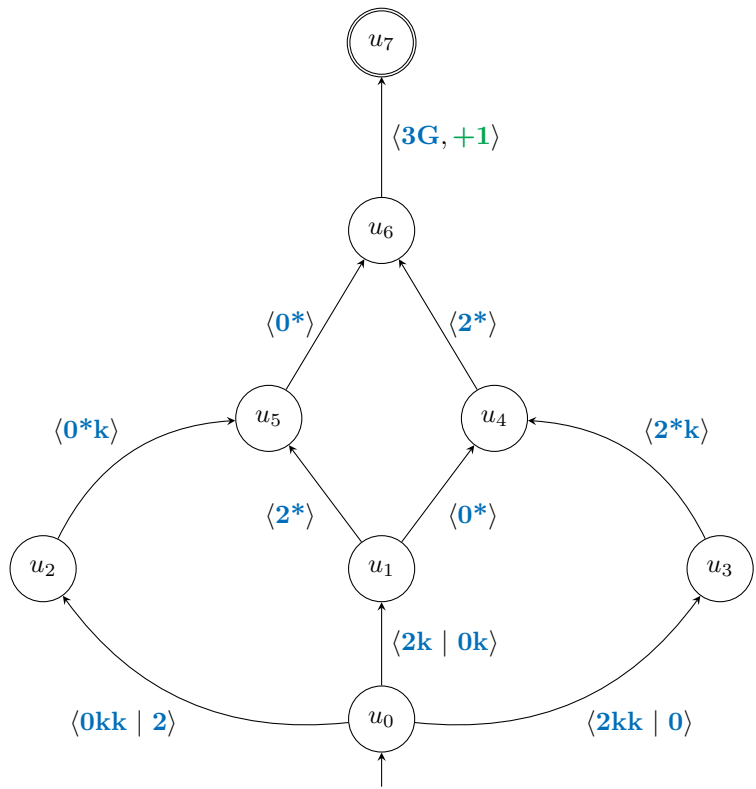
\begin{figure}
\label{fig:kw_rm}
\centering

    \begin{tikzpicture}[initial text={}, >=stealth, auto, node distance=3cm, on grid, initial where=below]

        \node[state,initial]            (0) at (0,0) {$u_0$};
        \node[state]                    (1) at (0,2) {$u_1$};
        \node[state]                    (2) at (-4.5,2) {$u_2$};
        \node[state]                    (3) at (4.5,2) {$u_3$};
        \node[state]                    (4) at (1.5, 4) {$u_4$};
        \node[state]                    (5) at (-1.5, 4) {$u_5$};
        \node[state]                    (6) at (0, 6.5) {$u_6$};
        \node[state,accepting]          (7) at (0, 9) {$u_7$};

        \path[->] (0)   edge                   node    [right]      {$\rmEdge{\ao{2k}\I\ao{0k}}$}   (1);
        \path[->] (0)   edge    [bend left]    node    [below left] {$\rmEdge{\ao{0kk}\I\ao{2}}$}   (2);
        \path[->] (0)   edge    [bend right]   node    [below right]{$\rmEdge{\ao{2kk}\I\ao{0}}$}   (3);
        \path[->] (1)   edge                   node    [below right]{$\rmEdge{\ao{0*}}$}            (4);
        \path[->] (1)   edge                   node    [below left] {$\rmEdge{\ao{2*}}$}            (5);
        \path[->] (2)   edge    [bend left]    node    [above left] {$\rmEdge{\ao{0*k}}$}           (5);
        \path[->] (3)   edge    [bend right]   node    [above right]{$\rmEdge{\ao{2*k}}$}           (4);
        \path[->] (4)   edge                   node    [right]      {$\rmEdge{\ao{2*}}$}            (6);
        \path[->] (5)   edge                   node    [left]       {$\rmEdge{\ao{0*}}$}            (6);
        \path[->] (6)   edge                   node    [right]      {$\rmEdge{\ao{3G},\rwP{+1}}$}   (7);
        
    \end{tikzpicture}

\caption{Perfect reward machine for the Keys World benchmark domain.}
\end{figure}

In the Keys World environment, the agent is tasked with reaching the \emph{goal} in room 3. The goal, however, is initially inaccessible for the agent, due to two closed doors being in the way. Each of them must be opened with a key. At the beginning of the episode, each key is randomly placed in either room 0 or 2, with no limit on the number of keys in the same room. Both keys can open either door, thus making them completely identical from the agent's point of view. At any given moment, the agent can only carry one key at a time. After opening both doors and reaching the goal, the agent completes its task, receives a reward of $\rwP{+1}$, and the episode is over. Figure \ref{fig:kw_rm} presents the \emph{perfect} reward machine that was supplied to each agent, based on the following set $\propsSet_{kw}$ of high-level events:
\[
    \propsSet_{kw} = \propsSet_r \ \cup \ \{ \ao{*}, \ao{g}, \ao{d}, \ao{k}, \ao{G} \}
\]
where $\ao{*}$ indicates that the agent is carrying a key, $\ao{g}$, $\ao{d}$, and $\ao{k}$ signal that the agent is in the same room as, respectively, the goal, a closed door, or a key. If needed, both $\ao{d}$ and $\ao{k}$ can be included twice in the labeling function output. Finally, $\ao{G}$ indicates that the agent has reached the goal and, thus, the episode is over. The Keys World task is significantly harder than the Cookie World one, as suggested by its more complex perfect reward machine, which is depicted in figure \ref{fig:kw_rm}. Nonetheless, its interpretation is straightforward:
\begin{itemize}
    \item $u_0$: the agent must figure out the location of the keys;
    \item $u_1$: the agent knows that each room contains one key, as he only found one upon entering either room 0 --- \ie $\ao{0k}$ --- or room 2 --- \ie $\ao{2k}$;
    \item $u_2$: the agent knows that both keys are in room 0, either because it saw them directly there --- \ie $\ao{0kk}$ --- or did not see any key in room 2 --- \ie $\ao{2}$;
    \item $u_3$: the agent knows that both keys are in room 2, under considerations analogous to the ones for $u_2$;
    \item $u_4$: the agent knows the remaining key is in room 2. This can happen in two situations: if arriving from $u_3$, both keys were in room 2, and the agent already picked up one of them --- \ie $\ao{2*k}$. Otherwise, if arriving from $u_1$, the keys were in both rooms, and the agent has already picked up the one in room 0 --- \ie $\ao{0*}$;
    \item $u_5$: the agent knows the remaining key is in room 0, under considerations analogous to those presented for $u_5$;
    \item $u_6$: the agent knows there are no keys left;
    \item $u_7$: the agent has reached the goal in room 3 --- \ie $\ao{3G}$ --- and, thus, the episode is over.
\end{itemize}

\subsubsection{Symbol World}
\begin{figure}
\label{fig:sw_rm1}
\centering

    \begin{tikzpicture}[initial text={}, >=stealth, auto, node distance=3cm, on grid, initial where=left]
    
        \node[state,initial]            (0) at (0,0) {$u_0$};
        \node[state]                    (1) at (4.5,0) {$u_1$};
        \node[state,accepting]         (10) at (10,0){$u_{10}$};

        \path[->] (0)   edge                    node    [above]     {$\rmEdge{\ao{3ax}}$}   (1);

        \path[->] (1)   edge     [bend left, looseness=1.5]    node    [above]     {$\rmEdge{\ao{0abC}\I\ao{2abC},\rwN{-1}}$}   (10);
        \path[->] (1)   edge                    node    [above]     {$\rmEdge{\ao{0Abc}\I\ao{2Abc},\rwP{+1}}$}   (10);
        \path[->] (1)   edge     [bend right, looseness=1.5]   node    [below]     {$\rmEdge{\ao{0aBc}\I\ao{2aBc},\rwN{-1}}$}   (10);
        
    \end{tikzpicture}

\caption{Partial view of the perfect reward machine for the Symbol world benchmark domain, restricted to the edges associated with RM state $u_5$, which represents the task of reaching the \textbf{a} symbol in either the north or south room.}
\end{figure}
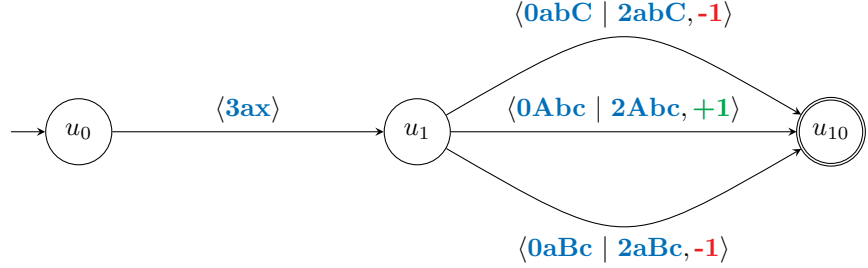

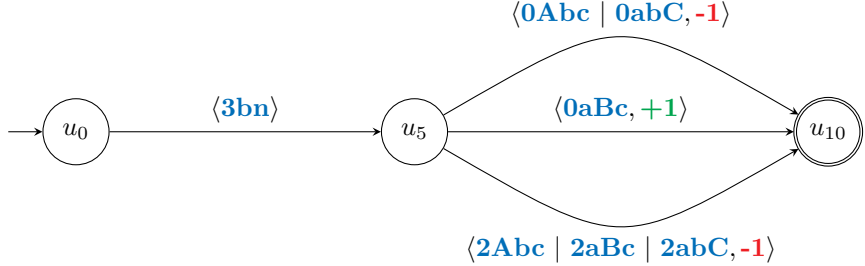
\begin{figure}
\label{fig:sw_rm2}
\centering

    \begin{tikzpicture}[initial text={}, >=stealth, auto, node distance=3cm, on grid, initial where=left]
    
        \node[state,initial]            (0) at (0,0) {$u_0$};
        \node[state]                    (5) at (4.5,0) {$u_5$};
        \node[state,accepting]         (10) at (10,0){$u_{10}$};

        \path[->] (0)   edge                    node    [above]     {$\rmEdge{\ao{3bn}}$}   (5);

        \path[->] (5)   edge     [bend left, looseness=1.5]    node    [above]     {$\rmEdge{\ao{0Abc}\I\ao{0abC},\rwN{-1}}$}   (10);
        \path[->] (5)   edge                    node    [above]     {$\rmEdge{\ao{0aBc},\rwP{+1}}$}   (10);
        \path[->] (5)   edge     [bend right, looseness=1.5]   node    [below]     {$\rmEdge{\ao{2Abc}\I\ao{2aBc}\I\ao{2abC},\rwN{-1}}$}   (10);
        
    \end{tikzpicture}

\caption{Partial view of the perfect reward machine for the Symbol world benchmark domain, restricted to the edges associated with RM state $u_5$, which represents the task of reaching the \textbf{b} symbol in the north room.}
\end{figure}

In the Symbol World environment, both room 0 and room 2 contain an identical set of three symbols, \textbf{a}, \textbf{b}, and \textbf{c}, while room 3 contains, on each episode, a randomly chosen set of instructions indicating which symbol the agent must collect to complete the task successfully. These indications consist of two parts: the symbol that must be reached, and the room where it must be collected. In this regard, there are three options. The instructions might indicate that the symbol must be collected in room 0, room 2, or either of them. Upon observing the instructions, the agent must reach one symbol in compliance with the received indications. However, contrary to the previous environments, the agent in Symbol World can also \emph{fail}, by collecting any symbol that does not comply with the indications, which gives a reward of $\rwN{-1}$. Conversely, if the agent reaches the correct symbol, it obtains a reward of $\rwP{+1}$. The set of high-level events for this environment is given by:
\[
    \propsSet_{sw} = \propsSet_r \ \cup \ \{ \ao{a}, \ao{b}, \ao{c}, \ao{n}, \ao{s}, \ao{x}, \ao{A}, \ao{B}, \ao{C} \}
\]
where $\ao{a}$, $\ao{b}$, and $\ao{c}$ indicate that the agent is in the same room as each of the corresponding symbols. This applies to both the symbols to be collected and the ones composing the agent's indications. Events $\ao{n}$, $\ao{s}$, and $\ao{x}$ are associated with the portion of the indications that relate to the agent having to collect its target symbol in, respectively, the north room --- \ie room 0 --- the south room --- \ie room 2 --- or either of them. Finally, $\ao{A}$, $\ao{B}$, and $\ao{C}$ are generated upon the agent collecting the associated symbols. The perfect reward machine for the Symbol World environment is composed of 11 states: $u_0$ is the initial state, where the agent knows it must reach its instructions. On the other side, state $u_{10}$ is reached by the task upon collecting a symbol, indicating the termination of the episode, either successful or unsuccessful. Then, the states $u_1, ..., u_9$ are each associated with one of the nine possible indications the agent can receive regarding its objective. For instance, figure \ref{fig:sw_rm1} contains the portions of the reward machine relating to the task of reaching the \textbf{a} symbol in either room. Upon observing the associated indications --- \ie \ao{3ax} --- the reward machine transitions from its initial state to $u_1$. Then, the agent can either: 
\begin{itemize}
    \item fail by collecting the \textbf{c} symbol in either room --- \ie $\ao{0abC}$ or $\ao{2abC}$;
    \item fail by collecting the \textbf{b} symbol in either room --- \ie $\ao{0aBc}$ or $\ao{2aBc}$;
    \item succeed by collecting the \textbf{a} symbol in either room --- \ie $\ao{0Abc}$ or $\ao{2Abc}$;
\end{itemize}
The same interpretation applies to any other state $u_2, ... u_9$. As a second example, figure \ref{fig:sw_rm2} shows the portion of the reward machine relating to the task of collecting the \textbf{b} symbol in the north room --- \ie $\ao{3bn}$. In this case, the agent succeeds only upon actually collecting the indicated symbol --- \ie $\ao{0aBc}$. In every other case\footnote{That is, in terms of high-level events: $\ao{0Abc}$, $\ao{0abC}$, $\ao{2Abc}$, $\ao{2aBc}$, and $\ao{2abC}$,}, the agent fails. 

As the reader might have noticed, this representation of the information the agent needs to solve its task properly is highly inefficient. In this regard, the Symbol World environment allows for one of the key limitations of reward machines to be highlighted. To successfully achieve its task, the agent only needs to remember two things: the requested symbol and the location where it must be collected. If the agent's memory were to be as simple as two numerical values, it could hold such information very efficiently, even if the set of symbols or possible locations were greatly larger. However, RM-based agents must store their information in the reward machine states. Therefore, to account for all the possible combinations of symbols and locations and, at the same time, maintain the \emph{perfection} property that is needed in partially observable environments, the reward machine for the Symbol World domain must contain one state \emph{for each possible combination} of symbols and location. While this does not pose a huge challenge in the Symbol World, due to there being only nine possible such combinations, if the set of either symbols or locations were to grow, this would lead to an exponential increase in the number of states required to deal with the task properly, quickly leading to its infeasibility.

\subsection{Agent training}
After the selection of the benchmark environments, the second aspect I had to consider was the choice of the learning algorithms that I would use for training the agents. In this regard, I decided to resort to Deep-QRM for two reasons: On the one hand, it represents the basic approach for the use of reward machines, thus representing a solid choice for the investigation of novel attack techniques. On the other hand, the use of the tabular version of QRM would have been impractical, due to the size and complexity of the benchmark environments. I also decided to experiment with the use of Automated Reward Shaping, to investigate if its benefits on the learning process could lead to any difference in terms of the agents' robustness to the proposed attacks or, dually, on the attack's effectiveness. In hindsight, this choice also proved fundamental in allowing for the proper convergence of the trained agents in the Keys World environment, as will be discussed later.
Finally, in terms of the exploratory policy to be used by the agents during training, I resorted to the $\epsilon$-greedy approach.

Altogether, I conducted six training sessions in total, two for each environment. Specifically, for each of them, I trained a total of twenty agents: ten with vanilla DQRM only, and ten with both DQRM and ARS. For the remainder of this chapter, I will refer to the former as ``regular agents'' and the latter as ``RS agents''.

\subsubsection{Training parameters}
In terms of the parameters relating to each agent's Q-network, I used the same values of \citeauthor{icarte_lrm_2019}, to allow for the obtained results to be comparable.
In particular, these are:
\begin{itemize}
    \item \textbf{number of hidden layers}: $5$;
    \item \textbf{hidden layer size}: $64$;
    \item \textbf{learning rate}: $5 \times 10^{-5}$;
    \item \textbf{batch size}: $32$;
    \item \textbf{experience buffer size}: $10^5$;
\end{itemize}

In terms of generic RL parameters, I resorted to a discount factor $\gamma = 0.9$, a reasonable and commonly chosen starting value in many RL research works, and a value of $\epsilon = 0.1$ for the $\epsilon$-greedy exploration policy. Moreover, I set a hard time limit of $500$ steps on the length of any episode: if, in such a window of time, the agent could not achieve its task, it would fail with zero reward.
Finally, in terms of total training steps, I chose different values for each environment, to account for their different levels of complexity. Specifically, I trained each agent for $3 \times 10^5$ steps in Cookie World, for $1 \times 10^6$ steps in Keys World, and for $2 \times 10^5$ steps in Symbol World. Indeed, while the Cookie and Symbol domains are quite comparable in terms of complexity, the Keys domain proved itself substantially harder to solve, as will be apparent when discussing the learning curves obtained during the training sessions.

\subsubsection{Training results}
\begin{figure}
    \centering
    \includegraphics[width=0.8\textwidth]{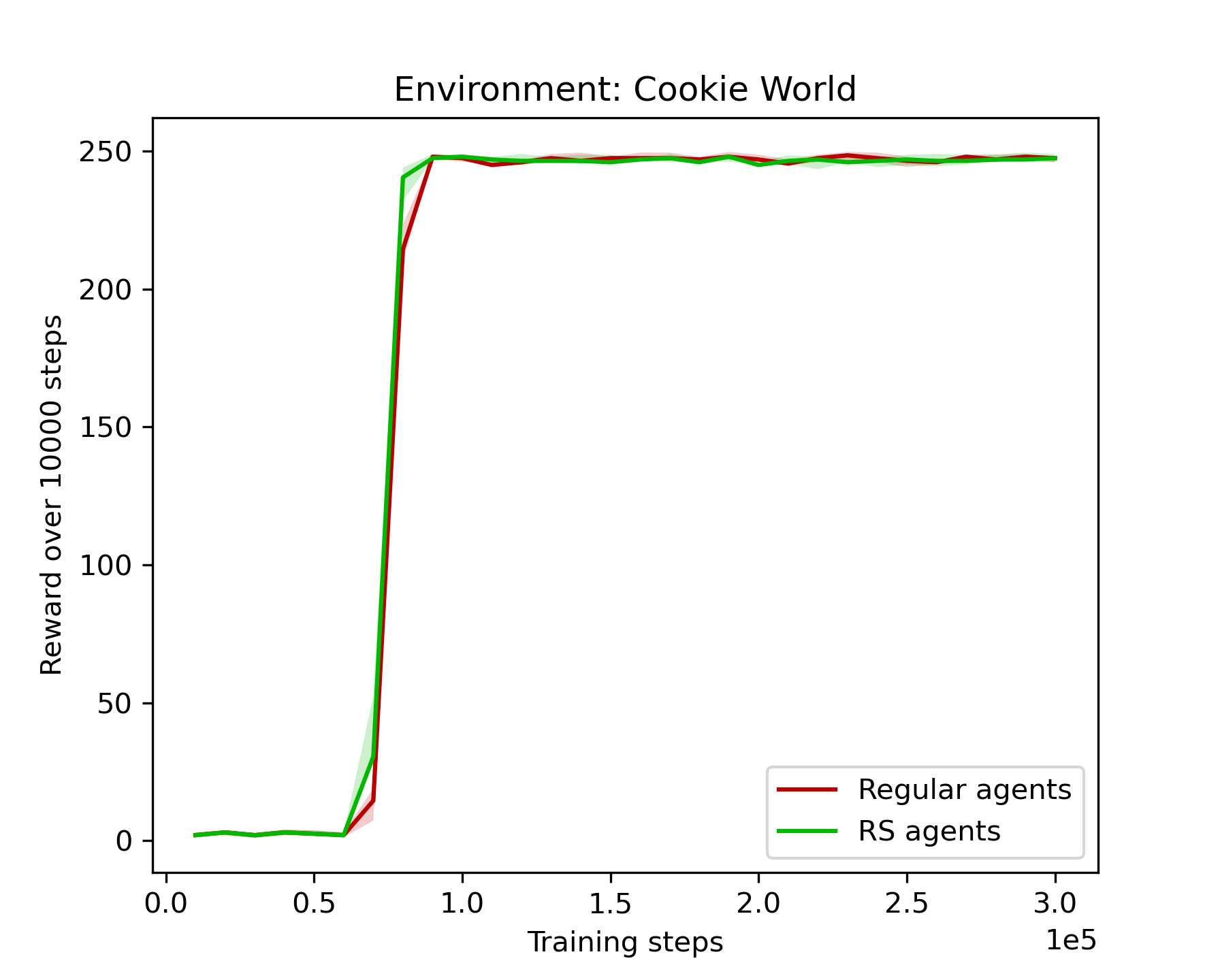}
    \caption{Results obtained during the agent training sessions for the Cookie World environment.}
    \label{fig:cw_training}
\end{figure}

\begin{figure}
    \centering
    \includegraphics[width=0.8\textwidth]{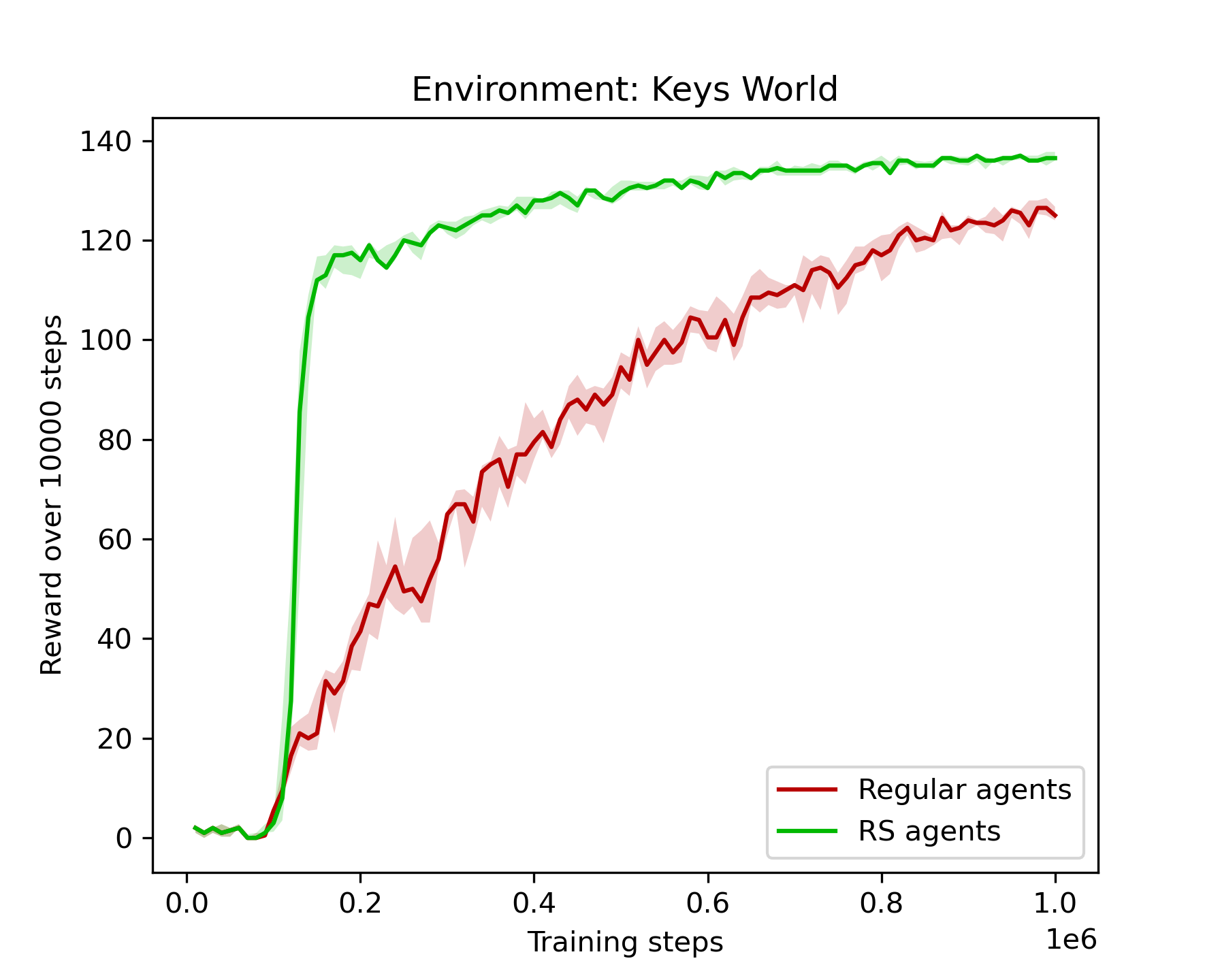}
    \caption{Results obtained during the agent training sessions for the Keys World environment.}
    \label{fig:kw_training}
\end{figure}

\begin{figure}
    \centering
    \includegraphics[width=0.8\textwidth]{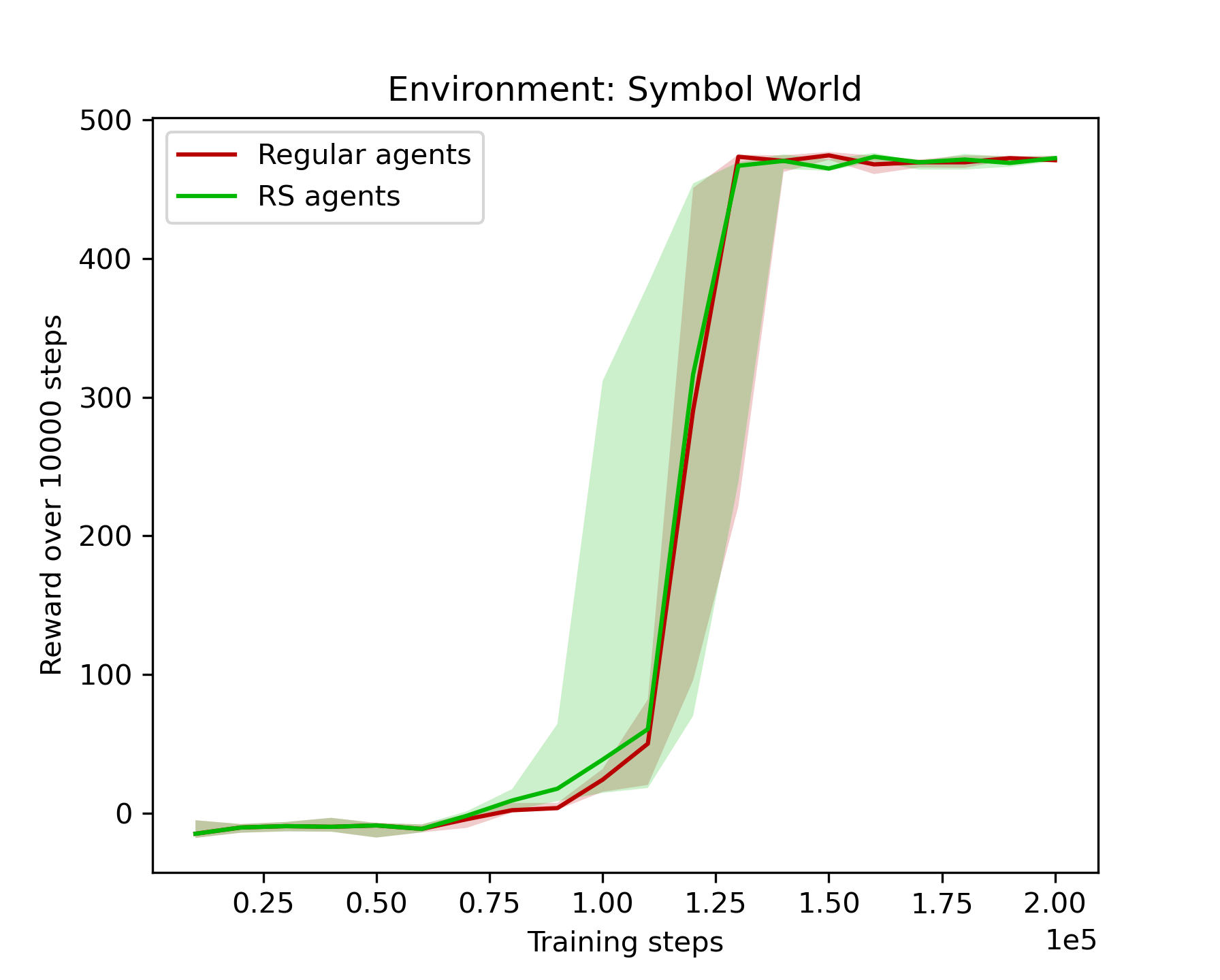}
    \caption{Results obtained during the agent training sessions for the Symbol World environment.}
    \label{fig:sw_training}
\end{figure}

Figures \ref{fig:cw_training}, \ref{fig:kw_training}, and \ref{fig:sw_training} present the learning curves obtained from, respectively, the Cookie World, Keys World, and Symbol World agents. Specifically, the y-axis represents the reward accumulated by the agents over ten thousand training steps. This choice is again derived directly from the work of \citeauthor{icarte_lrm_2019} and is one of the many possible metrics that could be used for assessing the level of proficiency achieved by the agent during its training. The solid lines in the plots represent the median value among the ten different agents, while the shaded area represents the range between the 25th and 75th percentiles. Together, these values allow for the assessment of the overall performance of the agents, as well as its variability among individuals.

In the Cookie World domain, all the agents converged to the optimal policies in under $1 \times 10^5$ steps. The use of ARS led only to a very slight increase in the convergence speed, probably due to the environment itself not representing a difficult challenge for the DQRM agents. In both cases, there was little to no variability in terms of performance among the agents. Similarly, the Symbol World agents needed slightly more steps to converge, circa $1.30 \times 10^5$. In terms of median performance, ARS again lead to no significant improvements. However, in this case, ARS led to an increase of the 75th percentile of the data, indicating that some RS agents were able to progress faster in the initial phase of the training process. That aside, convergence to the optimal policy was achieved roughly in the same number of iterations for all the agents, again indicating that the environment is probably not sufficiently complex for the agents to need ARS. 

On the other side, the Keys World domain was significantly harder for the agents to learn. After the $1 \times 10^6$ training steps, all the regular agents, \emph{except one}, still couldn't converge to optimal policies. However, when trained with ARS, all the agents reached optimal performance after around $8 \times 10^5$ steps. Moreover, the RS agents demonstrated a substantial spike in performance in the early steps, reaching the same median performance achieved at the end of the training process by regular agents, after less than $2 \times 10^5$ iterations. After this point, they spent the remaining $8 \times 10^5$ steps for refining their policy. In other words, the RS agents spent $20\%$ of their time to find good starting policies, and $80\%$ of the time to refine them to optimality, while the regular agents spent all their time slowly refining mediocre policies. This showcases the huge benefits made possible by the use of reward shaping. Moreover, the RS agents showed almost no variation in performance among each individual, while the regular ones showed some degree of variability in this regard, although limited.

\subsection{Testing sessions}
After being trained, the agents were used to carry out a number of experiments, each aimed at collecting meaningful data that could be used to analyze the effectiveness of the proposed attacks. Specifically, I carried out four different testing sessions. In each session, all the experiments were carried out by observing the agent's behavior over \textbf{one thousand episodes}, with a time limit of \textbf{five hundred steps} for each of them.

First, I conducted a \textbf{baseline} session, aimed at assessing the performance of the agents acting in the absence of any type of attack. This was needed to confirm that the agents were, indeed, able to achieve their tasks consistently. In this regard, the baseline session was fundamental in assuring that any agent failure observed in the following experiments could be ascribed only to the attacks, not the unreliability of the agents.

Then, I proceeded by carrying out a session to evaluate the impact arising from \textbf{random noise attacks}, in both their \emph{hallucination} and \emph{blinding} variations. Despite hallucination attacks not being further investigated by my work, I chose to include their random form in this set of experiments as a way of gaining a preliminary understanding of their difference, in terms of effectiveness, from blinding attacks. One interesting property of random noise attacks lies in the fact that they could be seen as not being attacks at all, due to their intrinsic lack of an attacker's strategy. Instead, they could be interpreted as the effect of some type of environmental noise affecting the labeling function outputs. In light of this, the results from this session could be equivalently interpreted in two ways: adhering to the attack interpretation, they dictate the minimum level of effectiveness that should be required from any more sophisticated attack. On the other side, under the environmental noise interpretation, they allow for the assessment of the agents' \emph{robustness} to noisy environments. In practice, this session was carried out by subjecting each trained agent to different amounts of noise, specifically $1\%,\ 5\%,\ 10\%,\ 20\%,\ 30\%,\ 40\%$, and $50\%$.

Finally, the third and fourth testing sessions were aimed, respectively, at the evaluation of \textbf{event-blinding attacks} and \textbf{edge-blinding attacks}. Specifically, the former were tested in both their \emph{atomic} and \emph{compound} variations, while the latter were tested in both their basic and \emph{state-based} variations. In both sessions, all the attacks were conducted according to the most promising blinding strategy identified by the algorithm presented in section \ref{sec:blinding_strats}.  Furthermore, all the attacks were tested under five timing strategies: once for the \emph{all-instances} and \emph{first-stream} ones, and three times for the \emph{triggered-stream}, with different values for the trigger probability, namely $30\%, 40\%$, and $50\%$.

\subsection{Evaluation metrics}
To properly investigate the effectiveness of my attacks, I resorted to a number of different metrics, starting with those allowing for the quantification of agent performance. 
In this regard, for each individual agent $\mathcal{A}$, I measured, over all episodes:
\begin{itemize}
    \item the total number of successes it achieved:
    \[
        s(\mathcal{A}) \in [0, 1000]
    \]
    \item the total number of failures it suffered: 
    \[
        f(\mathcal{A}) = 1000 - s(\mathcal{A})\in [0, 1000]
    \]
    \item the \emph{average} number of steps taken before successfully completing the task: 
    \[
        t_s(\mathcal{A}) \in [0, 500]
    \]
\end{itemize}
These metrics provided me with the basic information I needed to assess the performance of each agent.
Then, exclusively for the Symbol World agents, I also measured:
\begin{itemize}
    \item the \emph{average} number of steps taken before failing the task:
    \[
        t_f(\mathcal{A}) \in [0, 500]
    \]

    \item the \emph{average} reward obtained in case of task failure: 
    \[
        r_f(\mathcal{A}) \in [-1, 0]
    \]
\end{itemize}
These metrics were not considered for the Cookie World and Keys World agents, as, in those environments, the only way an agent can fail is by exceeding the time limit, which leads to zero reward. Therefore, the only possible values for these metrics would be $t_f(\mathcal{A}) = 500$ and $r_f(\mathcal{A}) = 0$.

For each domain and type of agent --- \ie regular or RS --- the above quantities were then averaged over their associated set $\Lambda$ of agents, leading to the metrics that I ultimately used to assess both the quality of the agents trained via DQRM and the effectiveness of my attacks, namely:
\begin{itemize}
    \item the \textbf{Average Success Rate}, which estimates the probability of an agent successfully completing its task, \ie its \emph{reliability}:
    \[
        \textbf{ASR}(\Lambda) = \frac{1}{|\Lambda|}\sum_{\mathcal{A} \in \Lambda} \frac{s(\mathcal{A})}{1000} \in [0, 1];
    \]
    \item the \textbf{Average Failure Rate}, which estimates the probability of an agent failing in its task, \ie its \emph{unreliability}:
    \begin{align*}
        \textbf{AFR}(\Lambda) &= 1 - \textbf{ASR}(\Lambda) \\
                     &= \frac{1}{|\Lambda|}\sum_{\mathcal{A} \in \Lambda} \frac{f(\mathcal{A})}{1000} \in [0, 1];
    \end{align*}
    \item the \textbf{Average Time-to-Success}, which estimates the \emph{expected efficiency} of an agent:
    \[
        \textbf{ATS}(\Lambda) = \frac{1}{|\Lambda|}\sum_{\mathcal{A} \in \Lambda} t_s(\mathcal{A}) \in [0, 500];
    \]
    \item the \textbf{Average Time-to-Failure}, which --- \emph{only for Symbol World agents} --- gives an indication of how fast an agent is in failing its task:
    \[
        \textbf{ATF}(\Lambda) = \frac{1}{|\Lambda|}\sum_{\mathcal{A} \in \Lambda} t_f(\mathcal{A}) \in [0, 500];
    \]
    \item the \textbf{Average Reward when Failing}, which --- \emph{only for Symbol World agents} --- is useful for determining the ratio between the number of failures due to time exhaustion and those due to other causes:
    \[
        \textbf{ARF}(\Lambda) = \frac{1}{|\Lambda|}\sum_{\mathcal{A} \in \Lambda} r_f(\mathcal{A}) \in [-1, 0];
    \]
\end{itemize}

In addition to the above metrics, I resorted to two metrics that relate specifically to the proposed attacks. To compute the first, I measured, for every agent $\mathcal{A}$ subject to a given attack technique $T$, the average ratio $\tau(\mathcal{A}, T) \in [0, 1]$ between the number of labeling function outputs that were tampered with and the length of the episode, \ie the total number of labeling outputs, computer over all the episodes. Then, I averaged these quantities over each set $\Lambda$ of agents of a given type and domain, to obtain the \textbf{Average Tampering Rate} of the attack:
\[
    \textbf{ATR}(T) = \frac{1}{|\Lambda|}\sum_{\mathcal{A} \in \Lambda} \tau(\mathcal{A}, T) \in [0, 1];
\]
which gives an estimate for the level of \emph{detectability} of the attack.
On the other hand, I designed the second attack-related metric to provide a score that could summarize the attack's overall impact, accounting for both its effectiveness and detectability. Specifically, for any attack technique $T$ carried out over a set $\Lambda$ of agents, I computed an \textbf{Impact Score} as follows:
\[
    \textbf{IS}(T, \Lambda; \alpha) = \frac{\alpha \sqrt{\textbf{AFR}(\Lambda)}}{\textbf{ATR}(T) + \alpha} \in [0,1]
\]
where the $\alpha$ parameter is introduced to normalize the output of the metric in the [0,1] range. 
This metric is adequate for my analysis as it summarizes two assumptions I made relating to the quality of an attack: On the one hand, its effectiveness is directly proportional to the failure rate it induces in its victims, and inversely proportional to its likelihood of being detected. Ideally, the highest impact possible arises from an attack that leads to a $100\%$ probability of agent failure and a $0\%$ probability of being detected. Indeed $\textbf{IS}(T, \Lambda; \alpha)$ is maximum when $\textbf{AFR}(\Lambda) = 1$ and $\textbf{ATR}(T) = 0$. 
On the other hand, the benefit derived by an attacker for an increase in the failure rate it induces in its victim is lower for easily detectable attacks and higher for stealthier ones. In other words, a highly successful attack that can be easily detected might be as good as a stealthier, less reliable one. This last observation is the rationale that motivated me in the choice of the value for the $\alpha$ parameter, which I set to $0.5$ for every attack technique and set of agents. Specifically, I chose this value by \emph{arbitrarily} assuming that a $20\%$ tampering rate corresponds to a $50\%$ chance of the attack being detected. Therefore, an attack characterized by a $50\%$ chance of it leading to the victim failing and a $0\%$ tampering rate is as good as an attack characterized by a $100\%$ victim failure rate and a $20\%$ tampering rate, as they are both as likely to be successful. 
\begin{figure}
    \centering
    \includegraphics[width=0.8\textwidth]{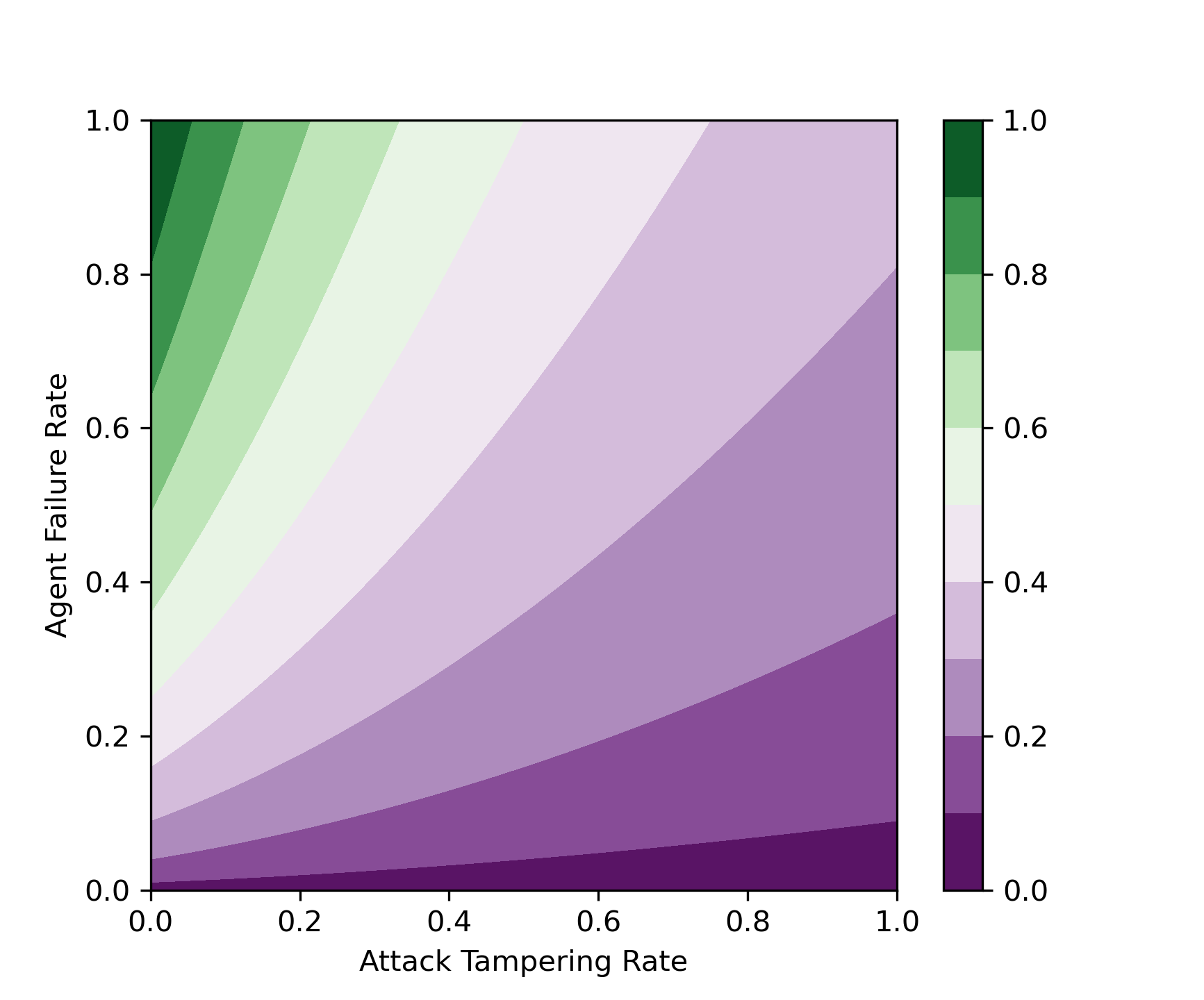}
    \caption{Contour plot for the Impact Score metric. The solid lines represent equal-value curves for $\textbf{IS}(T, \Lambda; 0.5)$}
    \label{fig:score_contours}
\end{figure}
Figure \ref{fig:score_contours} presents a contour plot for the impact score metric, which allows for all the above considerations to be easily confirmed by visual inspection.

\section{Experimental results}

\subsection{Baseline results}
Table \ref{tab:baselines} presents the results obtained, for each benchmark domain, by both the regular and RS agents. Specifically, this session's results allowed me to confirm that all the tested agents were able to achieve their tasks consistently in the minimum number of steps possible, also accounting for the $10\%$ probability of any action leading to a random outcome that characterized each domain. This was a strong requirement I wanted satisfied, as I was interested in evaluating the impact of my attacks on optimally trained agents. 

However, one important note must be made regarding the set of regular Keys World agents I used for the rest of the experimentation. As showcased by the learning curves from figure \ref{fig:kw_training}, the regular Keys World agents were not able to converge to the optimal policy. Moreover, when I first conducted the baseline testing session, their performance proved to be even worse than observed during training. After investigating the underlying reason, I found that most of the agents learned a policy that heavily relied on the stochasticity of the environment and the random-action aspect of the $\epsilon$-greedy strategy used during training. Specifically, the majority of them learned that the ``optimal'' action for one or more cells in the hallway was to simply try and move against a wall, which obviously led to them not moving at all. However, during the training session, this issue was not highlighted by the learning curves, because the agents were eventually able to evade the problematic locations either by means of the $10\%$ chance of random action outcome or $10\%$ chance of random action selection via $\epsilon$-greedy. However, when tested, the agents could only rely on their learned policy, without any form of non-determinism in their action selection process, thus achieving a lower level of performance. Specifically, this issue was so prominent that, when first tested, the baseline results for the regular Keys World agents were characterized by a $30\%$ average success rate and 100 steps circa of average time-to-success. In light of the objective of my evaluations, these agents were inadequate for assessing the quality of my attacks, and I decided to discard them. Fortunately, probably due to some lucky seed value, one regular Keys World agent did not suffer from this problem and, on the contrary, managed to learn a solid, optimal policy. Therefore, all the following results for the regular Keys World agents are, in reality, based on the performance of this one notable agent. This must be considered when interpreting the obtained results, as it strongly influences their statistical significance.

\begin{table}[h]
\centering
\resizebox{\textwidth}{!}{%
\begin{tabular}{@{}rrrr@{}}
\toprule
\textbf{Domain} & \textbf{Reward Shaping?} & \textbf{Avg. Success Rate} & \textbf{Avg. Time-to-Success} \\ \midrule
Cookie       & No       & 100\%                      & 35.8                                                     \\ \hd
Cookie       & Yes       & 100\%                      & 35.8                                                     \\ \HD
Keys         & No      & 100\%                      & 65.0                                                     \\ \hd
Keys         & Yes      & 100\%                      & 65.0                                                    \\ \HD
Symbol       & No       & 100\%                      & 18.6                                                    \\ \hd
Symbol       & Yes       & 100\%                      & 18.6                                                     \\ \bottomrule
\end{tabular}%
}
\caption{Baseline results obtained by both the regular and RS agents in the three benchmark domains that were used for all subsequent experiments.}
\label{tab:baselines}
\end{table}

\subsection{Random noise attacks}

\subsubsection{Hallucination noise}
Tables \ref{tab:cw_randomlf}, \ref{tab:kw_randomlf}, and \ref{tab:sw_randomlf} present the results obtained by subjecting each domain's regular agents to increasing amounts of random hallucination noise. 
\begin{table}[h]
\centering
\resizebox{\textwidth}{!}{%
\begin{tabular}{@{}rrrr@{}}
\toprule
\textbf{Noise level} & \textbf{Avg. Failure Rate} & \textbf{Impact Score}  & \textbf{Avg. Time-to-Success} \\ \midrule
1\%         & 1.72\%           & 0.1286     & 36.0                         \\ \hd
5\%         & 7.98\%           & 0.2568     & 36.5                         \\ \hd
10\%        & 15.76\%           & 0.3308    & 37.2                         \\ \hd
20\%        & 28.14\%           & 0.3789    & 39.3                         \\ \hd
30\%        & 38.65\%           & 0.3886    & 41.3                         \\ \hd
40\%        & 46.87\%           & 0.3803    & 42.9                         \\ \hd
50\%        & 54.82\%           & 0.3702    & 45.0                         \\ \bottomrule
\end{tabular}%
}

\caption{Results obtained by the Cookie World agents when acting under varying amounts of random hallucination noise.}
\label{tab:cw_randomlf}
\end{table}
The Cookie World agents showed an increase in their average failure rate that grew slightly faster than the amount of noise they were subjected to, peaking at a nearly $55\%$ chance of failing when faced with a $50\%$ chance of any labeling function output being noisy. However, according to the impact score metric, the attack achieved its best performance for a $30\%$ level of noise, due to the diminishing returns arising from increased amounts of noise. Moreover, the performance of the agents was slightly affected by the attack even when they managed to achieve their goal, with a maximum increase in their time-to-success of about $25\%$ for the highest amount of tested noise.

\begin{table}[h]
\centering
\resizebox{\textwidth}{!}{%
\begin{tabular}{@{}rrrr@{}}
\toprule
\textbf{Noise level} & \textbf{Avg. Failure Rate} & \textbf{Impact Score}  & \textbf{Avg. Time-to-Success} \\ \midrule
1\%         & 1.40\%           & 0.1160   & 65.5                              \\ \hd
5\%         & 8.70\%           & 0.2681   & 67.8                              \\ \hd
10\%        & 18.50\%          & 0.3584   & 70.4                             \\ \hd
20\%        & 31.80\%          & 0.4028   & 79.3                             \\ \hd
30\%        & 48.40\%          & 0.4348   & 87.1                             \\ \hd
40\%        & 64.30\%          & 0.4455   & 99.2                             \\ \hd
50\%        & 80.90\%          & 0.4497   & 112.6                            \\ \bottomrule
\end{tabular}%
}
\caption{Results obtained by the Keys World agent when acting under varying amounts of random hallucination noise.}
\label{tab:kw_randomlf}
\end{table}
For the Keys World agent, the attack was significantly more impactful in every aspect, with the amount induced failure rate growing noticeably faster than the amount of hallucination noise. The best-performing attack was achieved when the noise level was at its maximum tested value of $50\%$, leading to a substantial $80\%$ chance of the agent failing its task and a $71\%$ increase in the time required for solving the task.

\begin{table}[h]
\centering
\resizebox{\textwidth}{!}{%
\begin{tabular}{@{}rrrrrr@{}}
\toprule
\textbf{Noise level} & \textbf{Avg. Failure Rate} & \textbf{Impact Score} & \textbf{Avg. Time-to-Success}  & \textbf{Avg. Time-to-Failure} & \textbf{Avg Failure Reward} \\ \midrule
1\%         & 0.09\%           & 0.0294        & 18.6                              & 18.8                  & -1                 \\ \hd
5\%         & 0.56\%           & 0.0680       & 18.9                              & 18.8                  & -1                 \\ \hd
10\%        & 0.70\%           & 0.0697       & 19.1                              & 19.1                  & -1                 \\ \hd
20\%        & 2.03\%           & 0.1018       & 19.8                              & 19.6                  & -1                 \\ \hd
30\%        & 3.42\%           & 0.1156       & 20.5                              & 20.6                  & -1                 \\ \hd
40\%        & 4.65\%           & 0.1198       & 21.4                              & 21.5                  & -1                 \\ \hd
50\%        & 7.13\%           & 0.1335       & 22.3                              & 22.6                  & -1                 \\ \bottomrule
\end{tabular}%
}
\caption{Results obtained by the Symbol World agents when acting under varying amounts of random hallucination noise.}
\label{tab:sw_randomlf}
\end{table}
Surprisingly, the Symbol World agents proved to be the more robust to hallucination noise, with the attack not being able to reach as low as a $10\%$ chance of inducing an agent failure even when subject to a $50\%$ level of noise and requiring only as much as 4 extra steps to succeed in their task. However, interestingly, the average failure reward achieved by the agents for \emph{every} amount of noise showed a value of exactly $-1$. This indicates that all of the induced agent failures were not due to time exhaustion but to the agent reaching the wrong symbol. This suggests an interpretation of the low failure rates induced by the attack: If the random labels were introduced when the agents had already observed the indications relating to its objective, their impact was basically null. However, if the attacker managed to inject the \emph{right} observation at exactly the right time, he could lead the agent into misunderstanding its objective. From that point on, the agent would then go, and try to achieve the wrong task, unbothered by all subsequent noise due to its reward machine having already transitioned to the wrong state. This interpretation is also supported by the equal values obtained for the time-to-success and time-to-failure metrics.

Proceeding to the RS agents, tables \ref{tab:cwrs_randomlf}, \ref{tab:kwrs_randomlf}, and \ref{tab:swrs_randomlf} present the results they obtained, in each domain, when faced with random hallucination noise.
\begin{table}[h]
\centering
\resizebox{\textwidth}{!}{%
\begin{tabular}{@{}rrrr@{}}
\toprule
\textbf{Noise level} & \textbf{Avg. Failure Rate} & \textbf{Impact Score}  & \textbf{Avg. Time-to-Success} \\ \midrule
1\%         & 1.49\%           & 0.1197     & 35.9                              \\ \hd
5\%         & 7.81\%           & 0.2541     & 35.9                              \\ \hd
10\%        & 14.97\%          & 0.3224      & 36.0                             \\ \hd
20\%        & 26.90\%          & 0.3705      & 36.3                             \\ \hd
30\%        & 36.66\%          & 0.3784      & 36.6                             \\ \hd
40\%        & 43.85\%          & 0.3679      & 36.9                             \\ \hd
50\%        & 50.53\%          & 0.3554      & 37.2                             \\ \bottomrule
\end{tabular}%
}
\caption{Results obtained by the Cookie World RS agents when acting under varying amounts of random hallucination noise.}
\label{tab:cwrs_randomlf}
\end{table}
In the Cookie World environment, the RS agents proved to be only marginally more robust, with the induced failure rates being, for each level of noise, comparable to the ones shown by the regular agents. Thus, for this type of attack and domain, the main benefit arising from the use of reward shaping appears to be related to a lesser ability for the agent to impact the speed of the agent when its success can't be prevented. Indeed, under a $50\%$ level of noise, the increase in time-to-success was limited to less than 2 steps, compared to the 9-step increase observed for the regular agents.

\begin{table}[h]
\centering
\resizebox{\textwidth}{!}{%
\begin{tabular}{@{}rrrr@{}}
\toprule
\textbf{Noise level} & \textbf{Avg. Failure Rate} & \textbf{Impact Score}  & \textbf{Avg. Time-to-Success} \\ \midrule
1\%         & 2.00\%           & 0.1386     & 65.5                              \\ \hd
5\%         & 8.79\%           & 0.2695     & 67.3                              \\ \hd
10\%        & 17.46\%          & 0.3482     & 69.7                             \\ \hd
20\%        & 33.34\%          & 0.4124     & 75.0                             \\ \hd
30\%        & 49.61\%          & 0.4402     & 80.3                             \\ \hd
40\%        & 64.64\%          & 0.4467     & 87.3                             \\ \hd
50\%        & 79.16\%          & 0.4449     & 96.4                             \\ \bottomrule
\end{tabular}%
}
\caption{Results obtained by the Keys World RS agents when acting under varying amounts of random hallucination noise.}
\label{tab:kwrs_randomlf}
\end{table}
The same considerations apply to the Keys World RS agent, which again showed, when compared to the regular agent, no significant difference in terms of average failure rate. In terms of time-to-success, the benefits arising from the use of reward shaping are evident, as testified by the worst value being ``only'' $48\%$ higher than the baseline. Overall, the impact arising from this type of attack was significant despite the use of reward shaping.

\begin{table}[h]
\centering
\resizebox{\textwidth}{!}{%
\begin{tabular}{@{}rrrrrr@{}}
\toprule
\textbf{Noise level} & \textbf{Avg. Failure Rate} & \textbf{Impact Score} & \textbf{Avg. Time-to-Success}  & \textbf{Avg. Time-to-Failure} & \textbf{Avg Failure Reward} \\ \midrule
1\%         & 0.08\%           & 0.0277     & 18.6                              & 19.5                  & -1                 \\ \hd
5\%         & 0.41\%           & 0.0582     & 18.8                              & 19.2                  & -1                 \\ \hd
10\%        & 0.94\%           & 0.0808     & 19.0                              & 19.0                  & -1                 \\ \hd
20\%        & 1.86\%           & 0.0974     & 19.6                              & 19.4                  & -1                 \\ \hd
30\%        & 2.67\%           & 0.1021     & 20.1                              & 20.3                  & -1                 \\ \hd
40\%        & 4.35\%           & 0.1159     & 20.8                              & 21.0                  & -1                 \\ \hd
50\%        & 7.18\%           & 0.1340     & 21.7                              & 21.7                  & -1                 \\ \bottomrule
\end{tabular}%
}
\caption{Results obtained by the Symbol World RS agents when acting under varying amounts of random hallucination noise.}
\label{tab:swrs_randomlf}
\end{table}
Finally, in the case of Symbol World, the use of reward shaping appears to have led to no difference at all, with the values for all the considered metrics being basically equal to those obtained by the regular agents. This further supports the interpretation proposed when discussing the previous results: even though the use of reward shaping can lead to better policies, this does not help the agents as their policies are already sufficiently robust to deal with random hallucination attacks, except in the rare instances where they succeed in inducing a misunderstanding the task's objective.

Overall, random hallucination attacks proved to be quite effective in both the Cookie and Keys World domains. Conversely, their performance in the Symbol World environment was poor. Nonetheless, this domain provided interesting insights relating to the hallucination attack's ability to induce an agent to pursue the wrong objectives. Therefore, if properly designed, more sophisticated attack strategies could potentially lead to more consistent results in this regard and, thus, pose serious concern even in this environment. 

\subsubsection{Blinding noise}
Tables \ref{tab:cw_randblind}, \ref{tab:kw_randblind}, and \ref{tab:sw_randblind} present the results for each domain's regular agents when attacked with increasing amounts of random blinding noise.
\begin{table}[h]
\centering
\resizebox{\textwidth}{!}{%
\begin{tabular}{@{}rrrr@{}}
\toprule
\textbf{Noise level} & \textbf{Avg. Failure Rate} & \textbf{Impact Score}  & \textbf{Avg. Time-to-Success} \\ \midrule
1\%         & 0.16\%           & 0.0392     & 35.9                              \\ \hd
5\%         & 0.80\%           & 0.0813     & 36.2                              \\ \hd
10\%        & 1.46\%           & 0.1007     & 36.7                              \\ \hd
20\%        & 3.32\%           & 0.1301     & 37.7                              \\ \hd
30\%        & 6.04\%           & 0.1536     & 39.1                              \\ \hd
40\%        & 8.54\%           & 0.1623     & 40.7                              \\ \hd
50\%        & 11.71\%          & 0.1711     & 43.3                             \\ \bottomrule
\end{tabular}%
}
\caption{Results obtained by the Cookie World agents when acting under varying amounts of random blinding noise.}
\label{tab:cw_randblind}
\end{table}
The Cookie World agents were quite robust to this type of noise, showing only a modest increase in their average failure rate, with a maximum value of about $12\%$ when faced with a $50\%$ of noisy labeling function outputs. Concurrently, for increasing amounts of noise, they also showed a minor increase in the time needed to achieve their task. Overall, the effectiveness of this type of attack on these agents was low, as reflected by the obtained impact scores.

\begin{table}[h]
\centering
\resizebox{\textwidth}{!}{%
\begin{tabular}{@{}rrrr@{}}
\toprule
\textbf{Noise level} & \textbf{Avg. Failure Rate} & \textbf{Impact Score}  & \textbf{Avg. Time-to-Success} \\ \midrule
1\%         & 0.80\%           & 0.0877     & 65.5                              \\ \hd
5\%         & 4.60\%           & 0.1950     & 67.5                              \\ \hd
10\%        & 10.30\%          & 0.2674     & 70.3                             \\ \hd
20\%        & 20.30\%          & 0.3218     & 76.8                             \\ \hd
30\%        & 31.70\%          & 0.3519     & 84.7                             \\ \hd
40\%        & 45.20\%          & 0.3735     & 95.4                             \\ \hd
50\%        & 59.10\%          & 0.3844     & 110.2                            \\ \bottomrule
\end{tabular}%
}
\caption{Results obtained by the Keys World agent when acting under varying amounts of random blinding noise.}
\label{tab:kw_randblind}
\end{table}
Conversely, in the case of the --- \emph{single} --- Keys World agent, this type of attack was significantly more effective, leading to an increase in failure rate which seems to grow slightly faster than the amount of noise required to induce it, peaking to an almost $60\%$ chance of agent failure in light of half of the labeling function outputs being noisy. In terms of time-to-success, the results are comparable to those obtained for the hallucination noise, suggesting that the decrease in agent speed is not strictly related to the injection of events that do not hold true, and could, instead, be traced back to the frequently inaccurate representation of the state of the environment provided by the noisy labeling function.

\begin{table}[h]
\centering
\resizebox{\textwidth}{!}{%
\begin{tabular}{@{}rrrr@{}}
\toprule
\textbf{Noise level} & \textbf{Avg. Failure Rate} & \textbf{Impact Score}  & \textbf{Avg. Time-to-Success} \\ \midrule
1\%         & 0.00\%          & 0.0000  & 18.6                              \\ \hd
5\%         & 0.00\%          & 0.0000  & 18.8                              \\ \hd
10\%        & 0.00\%          & 0.0000  & 19.0                              \\ \hd
20\%        & 0.00\%          & 0.0000  & 19.6                              \\ \hd
30\%        & 0.00\%          & 0.0000  & 20.2                              \\ \hd
40\%        & 0.00\%          & 0.0000  & 21.1                              \\ \hd
50\%        & 0.00\%          & 0.0000  & 22.3                              \\ \bottomrule
\end{tabular}%
}
\caption{Results obtained by the Symbol World agents when acting under varying amounts of random blinding noise.}
\label{tab:sw_randblind}
\end{table}
The results for the Symbol World domain agents show the complete ineffectiveness of the attack which, however, is not surprising in light of the results that were obtained by the hallucination attack. Indeed, these results confirm once again the intuition that was previously presented: Since the random blinding noise has no way to induce the Symbol World agent to misunderstand their task, their effect is restricted to a minor increase of the agent's speed in solving the task.

Proceeding to the RS agents, tables \ref{tab:cwrs_randblind}, \ref{tab:kwrs_randblind}, and \ref{tab:swrs_randblind} present the results that they obtained. 
\begin{table}[h]
\centering
\resizebox{\textwidth}{!}{%
\begin{tabular}{@{}rrrr@{}}
\toprule
\textbf{Noise level} & \textbf{Avg. Failure Rate} & \textbf{Impact Score}  & \textbf{Avg. Time-to-Success} \\ \midrule
1\%         & 0.17\%           & 0.0404     & 35.9                              \\ \hd
5\%         & 0.85\%           & 0.0838     & 36.0                              \\ \hd
10\%        & 1.63\%           & 0.1064     & 36.3                              \\ \hd
20\%        & 3.52\%           & 0.1340     & 36.8                              \\ \hd
30\%        & 6.10\%           & 0.1544     & 37.6                              \\ \hd
40\%        & 9.14\%           & 0.1680     & 38.7                              \\ \hd
50\%        & 11.87\%          & 0.1723     & 40.3                             \\ \bottomrule
\end{tabular}%
}
\caption{Results obtained by the Cookie World RS agents when acting under varying amounts of random blinding noise.}
\label{tab:cwrs_randblind}
\end{table}
In the Cookie World environment, the random blinding attack achieved, for each noise level, the same level of performance that was obtained in the case of regular agents, suggesting that the use of reward shaping does not provide a huge benefit in terms of robustness to this type of attack.

\begin{table}[h]
\centering
\resizebox{\textwidth}{!}{%
\begin{tabular}{@{}rrrr@{}}
\toprule
\textbf{Noise level} & \textbf{Avg. Failure Rate} & \textbf{Impact Score}  & \textbf{Avg. Time-to-Success} \\ \midrule
1\%         & 1.16\%           & 0.1056     & 65.3                              \\ \hd
5\%         & 5.40\%           & 0.2113     & 66.8                              \\ \hd
10\%        & 10.31\%          & 0.2676     & 68.6                             \\ \hd
20\%        & 21.83\%          & 0.3337     & 73.4                             \\ \hd
30\%        & 34.04\%          & 0.3646     & 79.6                             \\ \hd
40\%        & 47.67\%          & 0.3836     & 86.3                             \\ \hd
50\%        & 63.57\%          & 0.3987     & 96.0                             \\ \bottomrule
\end{tabular}%
}
\caption{Results obtained by the Keys World RS agents when acting under varying amounts of random blinding noise.}
\label{tab:kwrs_randblind}
\end{table}
Surprisingly, in the case of Keys World RS agents, the attack seemed to have a slightly higher impact, in terms of agent failure rate, than the one it had on the regular agent. However, this is likely due to chance: given that the regular agent results were obtained on a single agent, it is highly possible that the obtained results were, in that case, a lucky instance, not perfectly representative of the actual impact of the attack. In terms of time-to-success, the RS agents showed, when compared with regular agents, a decrease in speed which grew slower as the noise level increased. This is in line with what was observed until now, further consolidating the idea that reward shaping can lead to beneficial effects in this regard.

\begin{table}[h]
\centering
\resizebox{\textwidth}{!}{%
\begin{tabular}{@{}rrrrrr@{}}
\toprule
\textbf{Noise level} & \textbf{Avg. Failure Rate} & \textbf{Impact Score} & \textbf{Avg. Time-to-Success}  & \textbf{Avg. Time-to-Failure} & \textbf{Avg Failure Reward} \\ \midrule
1\%         & 0.00\%          & 0.0000  & 18.6                              & -                     & -                  \\ \hd
5\%         & 0.00\%          & 0.0000  & 18.7                              & -                     & -                  \\ \hd
10\%        & 0.01\%          & 0.0083  & 18.9                              & 18                    & -1                 \\ \hd
20\%        & 0.01\%          & 0.0071  & 19.2                              & 20                    & -1                 \\ \hd
30\%        & 0.02\%          & 0.0088  & 19.6                              & 19                    & -1                 \\ \hd
40\%        & 0.01\%          & 0.0056  & 20.2                              & 19                    & -1                 \\ \hd
50\%        & 0.01\%          & 0.0050  & 20.9                              & 20                    & -1                 \\ \bottomrule
\end{tabular}%
}
\caption{Results obtained by the Symbol World RS agents when acting under varying amounts of random blinding noise.}
\label{tab:swrs_randblind}
\end{table}
In the case of Symbol World RS agents, the random blinding attack led to basically no impact in every case. However, it is interesting to notice how, for noise levels of $10\%$ and over, there were rare instances of agent failure. Moreover, as testified by the average failure reward metric, the cause of these failures can be identified as the agents reaching the wrong symbol, which is unexpected, as blinding attacks are not able to cause a misunderstanding of the agent's task. Since the failure rates are lower than $0.05\%$, they might as well be due to some particularly unusual sequence of actions and events that the agents did not learn to handle during training. Nonetheless, the fact that these results were obtained for \emph{all} noise levels from $10\%$ on might allow for some speculation being made towards the hypothesis of the underlying cause being the attack itself. However, due to the extremely rare nature of these scenarios, a deeper investigation would be needed to accept either of the above explanations.

Overall, the random blinding attacks led to a moderate impact on the agents' performance only in the case of the Keys World environment. This, in addition to the results obtained in the case of hallucination noise, suggests that these agents are quite sensitive to any type of labeling noise. Conversely, the Cookie World agents proved themselves to be significantly more robust to blinding noise than they are to hallucination noise. Finally, the Symbol World agents suffered from no amount of performance decrease, showcasing their complete robustness to random blinding attacks.

\subsection{Event-based blinding attacks}
Following the results obtained by the random blinding attacks, observing the results obtained by event-based ones was extremely interesting, as it allowed me to investigate whether the robustness demonstrated by Cookie and Symbol World agents is an intrinsic property arising from their learned policy or, conversely, it is due to the lack of more sophisticated blinding and timing strategies. Tables \ref{tab:cw_evtblind}, \ref{tab:kw_evtblind}, \ref{tab:sw_evtblind} present the results obtained by both the atomic and compound variations of event-blinding attacks, for the three benchmark domains, using five different timing strategies. Note that the time-to-success metric is not included in any of the following results as, in \emph{every} set of them, its value was never significantly different from the baselines. This indicates that, when unable to induce the agents to fail their tasks, the event-based attacks are also unable to cause any other type of decrease in their proficiency. In other words, this type of attack either works or doesn't. 

Another interesting observation that applies to the results obtained in every domain relates to the difference in performance between atomic and compound attacks. Specifically, both of them lead to the exact same results for every set of agents, suggesting that the chosen blinding strategies did not differ between the two. For this to be possible, the blinding targets picked by the compound attacks must consist of only one event. In all three domains, the only possible labeling function outputs consisting of a single event are those generated when the agent is in an empty room: either $\ao{0}$, $\ao{1}$, $\ao{2}$, or $\ao{3}$. Therefore, in all of their instances, event-blinding attacks chose to deprive the agents of their knowledge regarding their current, empty room. Interestingly, if done frequently enough, this was sufficient to induce very high failure rates in the victims.

\begin{table}[h]
\centering
\resizebox{\textwidth}{!}{%
\begin{tabular}{@{}rrrrr@{}}
\toprule
\textbf{Attack Type} & \textbf{Timing Strategy} & \textbf{Avg. Tampering Rate} & \textbf{Avg. Failure Rate}     & \textbf{Impact Score}   \\ \midrule
atom/comp        & all                 & 88.87\%             & 80.00\%           & 0.3220                             \\ \hd
atom/comp        & first               & 73.58\%             & 70.00\%           & 0.3385                             \\ \hd
atom/comp        & trigger\_30         & 75.26\%             & 56.93\%           & 0.3012                             \\ \hd
atom/comp        & trigger\_40         & 75.55\%             & 61.80\%           & 0.3131                             \\ \hd
atom/comp        & trigger\_50         & 74.91\%             & 65.40\%           & 0.3237                             \\ \bottomrule
\end{tabular}%
}
\caption{Results obtained by the Cookie World agents when attacked with both variations of event-based blinding attacks.}
\label{tab:cw_evtblind}
\end{table}
In the case of Cookie World agents, the best impact score was obtained when using the \emph{first-stream} timing strategy, leading to a $70\%$ chance of agent failure by means of a $74\%$ tampering rate. The triggered strategies proved themselves to be worse, achieving significantly lower agent failure rates while presenting comparable tampering rates. Finally, the \emph{all-instances} timing strategy led to even higher failure rates, at the cost, however, of an almost $90\%$ tampering rate. Despite the very high number of attacker interventions, which likely limits the real-world applicability of the attack, the high induced failure rates provide interesting evidence in support of the fact that, if properly designed, blinding attacks can be an effective way of hindering an RM-based agent's performance. 

\begin{table}[h]
\centering
\resizebox{\textwidth}{!}{%
\begin{tabular}{@{}rrrrr@{}}
\toprule
\textbf{Attack Type} & \textbf{Timing Strategy} & \textbf{Avg. Tampering Rate} & \textbf{Avg. Failure Rate}     & \textbf{Impact Score}   \\ \midrule
atom/comp        & all           & 100.00\%            & 100.00\%            & 0.3334                            \\ \hd
atom/comp        & first         & 100.00\%            & 100.00\%            & 0.3334                            \\ \hd
atom/comp        & trigger\_30   & 87.04\%             & 99.10\%            & 0.3632                             \\ \hd
atom/comp        & trigger\_40   & 92.78\%             & 99.60\%            & 0.3495                             \\ \hd
atom/comp        & trigger\_50   & 96.66\%             & 99.90\%            & 0.3408                             \\ \bottomrule
\end{tabular}%
}
\caption{Results obtained by the Keys World agent when attacked with both variations of event-based blinding attacks.}
\label{tab:kw_evtblind}
\end{table}
For the --- \emph{single} --- Keys World agent, event-based attacks were able to nullify the victim's ability to achieve its task completely, at the cost of tampering with \emph{every} labeling function output. This was observed for both the \emph{all-instances} and \emph{first-stream} timing strategies. An attack showing a $100\%$ tampering rate is a very interesting fact by itself, as it implies that all the outputs produced by the labeling function contained the target and, consequently, were the same. Therefore, for them to never change, the agent must have stayed for all the 500 timesteps in its initial position, namely the hallway. This clearly indicates that the attacks were able to effectively \emph{trap} the victim into its current room. However, in terms of impact score, the best results were obtained by the \emph{triggered-stream} timing strategy, when applied with a $30\%$ trigger chance. Indeed, in this case, the victim failed about $99\%$ of time while only requiring less than $90\%$ of the labeling function outputs to be tampered with.

\begin{table}[h]
\centering
\resizebox{\textwidth}{!}{%
\begin{tabular}{@{}rrrrrrr@{}}
\toprule
\textbf{Attack Type} & \textbf{Timing Strategy} & \textbf{Avg. Tampering Rate} & \textbf{Avg. Failure Rate} & \textbf{Impact Score}  & \textbf{Avg. Time-to-Failure} & \textbf{Avg Failure Reward} \\ \midrule
atom/comp        & all           & 98.87\%             & 70.11\%           & 0.2812                             & 500                   & 0                  \\ \hd
atom/comp        & first         & 53.69\%             & 39.52\%           & 0.3031                             & 500                   & 0                  \\ \hd
atom/comp        & trigger\_30   & 83.42\%             & 31.63\%           & 0.2108                             & 500                   & 0                  \\ \hd
atom/comp        & trigger\_40   & 77.40\%             & 32.77\%           & 0.2247                             & 500                   & 0                  \\ \hd
atom/comp        & trigger\_50   & 69.45\%             & 33.27\%           & 0.2414                             & 500                   & 0                  \\ \bottomrule
\end{tabular}%
}
\caption{Results obtained by the Symbol World agents when attacked with both variations of event-based blinding attacks.}
\label{tab:sw_evtblind}
\end{table}
In the case of the Symbol World domain, the agents once again proved themselves to be the most robust to blinding attacks. The \emph{all-instances} timing strategy demonstrated the same trap-like properties that were observed for the Keys World agent. However, in this case, the agents were able to ``evade'' about $30\%$ of the time, leading to an average failure rate of only $70\%$. Interestingly, the triggered timing strategies also seemed to present some degree of trap-like effect but led to vastly lower failure rates. This suggests that proper timing might be a critical factor in allowing the attacker to reduce the chance of the victim being able to recover from the attack.

Proceeding with the RS agents, tables \ref{tab:cwrs_evtblind}, \ref{tab:kwrs_evtblind}, and \ref{tab:kwrs_evtblind} present the results that were obtained by exposing them to event-based attacks. 
\begin{table}[h]
\centering
\resizebox{\textwidth}{!}{%
\begin{tabular}{@{}rrrrr@{}}
\toprule
\textbf{Attack Type} & \textbf{Timing Strategy} & \textbf{Avg. Tampering Rate} & \textbf{Avg. Failure Rate}     & \textbf{Impact Score}   \\ \midrule
atom/comp        & all           & 58.36\%             & 20.50\%           & 0.2089                             \\ \hd
atom/comp        & first         & 15.65\%             & 0.50\%           & 0.0538                              \\ \hd
atom/comp        & trigger\_30   & 22.18\%             & 1.99\%           & 0.0977                              \\ \hd
atom/comp        & trigger\_40   & 21.01\%             & 1.56\%          & 0.0879                              \\ \hd
atom/comp        & trigger\_50   & 18.19\%             & 0.91\%           & 0.0699                              \\ \bottomrule
\end{tabular}%
}
\caption{Results obtained by Cookie World RS agents when attacked with both variations of event-based blinding attacks.}
\label{tab:cwrs_evtblind}
\end{table}
In the Cookie World domain, the RS agents showed greatly improved robustness to the attack when compared with regular ones, with as much as about a $60\%$ tampering rate being needed to induce as little as a $20\%$ chance of agent failure in the case of \emph{all-instances} timing. The other strategies proved themselves to be even less effective, never causing failure rates higher than $2\%$, in spite of tampering rates between $15\%$ and $20\%$.

\begin{table}[h]
\centering
\resizebox{\textwidth}{!}{%
\begin{tabular}{@{}rrrrr@{}}
\toprule
\textbf{Attack Type} & \textbf{Timing Strategy} & \textbf{Avg. Tampering Rate} & \textbf{Avg. Failure Rate}     & \textbf{Impact Score}   \\ \midrule
atom/comp        & all           & 86.08\%             & 95.24\%            & 0.3586                             \\ \hd
atom/comp        & first         & 4.95\%              & 0.00\%          & 0.0000                              \\ \hd
atom/comp        & trigger\_30   & 58.90\%             & 17.04\%           & 0.1895                             \\ \hd
atom/comp        & trigger\_40   & 46.46\%             & 10.43\%           & 0.1674                             \\ \hd
atom/comp        & trigger\_50   & 35.13\%             & 6.37\%           & 0.1482                              \\ \bottomrule
\end{tabular}%
}
\caption{Results obtained by Keys World RS agents when attacked with both variations of event-based blinding attacks.}
\label{tab:kwrs_evtblind}
\end{table}
Similar results were observed for the Keys World RS agents: while the \emph{all-instances} timing strategy was able to maintain most of its effectiveness, all the other strategies showed a significant decrease in performance when compared to the impact they had on regular agents. One possible explanation for the \emph{all-instances} strategy being somewhat an exception might reside in some time-sensitive property of the attack. Under this interpretation, the use of reward shaping might be seen as a tool to reduce the window of opportunity of the attacker, thus decreasing its overall performance.

\begin{table}[h]
\centering
\resizebox{\textwidth}{!}{%
\begin{tabular}{@{}rrrrrrr@{}}
\toprule
\textbf{Attack Type} & \textbf{Timing Strategy} & \textbf{Avg. Tampering Rate} & \textbf{Avg. Failure Rate} & \textbf{Impact Score}  & \textbf{Avg. Time-to-Failure} & \textbf{Avg Failure Reward} \\ \midrule
atom/comp        & all           & 98.26\%             & 47.35\%           & 0.2321                             & 500                   & 0                  \\ \hd
atom/comp        & first         & 22.88\%             & 0.00\%          & 0.0000                              & -                     & -                  \\ \hd
atom/comp        & trigger\_30   & 72.13\%             & 7.43\%           & 0.1116                              & 500                   & 0                  \\ \hd
atom/comp        & trigger\_40   & 62.81\%             & 4.67\%           & 0.0958                              & 500                   & 0                  \\ \hd
atom/comp        & trigger\_50   & 50.67\%             & 2.43\%           & 0.0774                              & 500                   & 0                  \\ \bottomrule
\end{tabular}%
}
\caption{Results obtained by Symbol World RS agents when attacked with both variations of event-based blinding attacks.}
\label{tab:swrs_evtblind}
\end{table}
Finally, in the case of Symbol World RS agents, the results were qualitatively similar to those obtained for Keys World RS agents, with each timing strategy leading to lower failure rates than the one observed for regular agents. However, in this case, the robustness-enhancing effect appeared to be even higher, with the top-performing attack only being able to induce about a $48\%$ chance of the victim failing with a $98\%$ tampering rate. 

Overall, event-blinding attacks were able to perform better than random-blinding ones, thus confirming that the use of more sophisticated blinding strategies is of paramount importance in determining the quality of the attack. However, this increase in performance was always associated with very high tampering rates. In practice, this severely limits the feasibility of carrying out such attacks in every scenario where the victim could reasonably be expected to make at least some type of effort to try and detect the attacker. In terms of robustness, the use of reward-shaping showed great promise in increasing the robustness of the agents to this type of attack, with varying outcomes depending on the domain at hand.

\subsection{Edge-based blinding attacks}
To conclude the discussion of the results of the experimental phase of my work, table \ref{tab:cw_edgblind}, \ref{tab:kw_edgblind}, and \ref{tab:sw_edgblind} present the results obtained by subjecting the agents trained for each environment to both edge-blinding and state-blinding attacks with the usual timing strategies.

\begin{table}[h]
\centering
\resizebox{\textwidth}{!}{%
\begin{tabular}{@{}rrrrr@{}}
\toprule
\textbf{Attack Type} & \textbf{Timing Strategy} & \textbf{Avg. Tampering Rate} & \textbf{Avg. Failure Rate}     & \textbf{Impact Score}   \\ \midrule
edge/state        & all           & 73.44\%             & 100.00\%        & 0.4051                               \\ \hd
edge/state        & first         & 2.68\%              & 0.00\%          & 0.0000                              \\ \hd
edge/state        & trigger\_30   & 0.84\%              & 0.00\%          & 0.0000                              \\ \hd
edge/state        & trigger\_40   & 1.12\%              & 0.00\%          & 0.0000                             \\ \hd
edge/state        & trigger\_50   & 1.39\%              & 0.00\%          & 0.0000                              \\ \bottomrule
\end{tabular}%
}
\caption{Results obtained by Cookie World agents when attacked with both variations of edge-based blinding attacks.}
\label{tab:cw_edgblind}
\end{table}
When carried out over Cookie World regular agents, the two attacks led to completely identical results, indicating that the choice of the blinding strategy was the same. For this to be possible, the state-based attacks must have resorted to a strategy targeting a state with a single entering transition. Thus, in this domain, the only possibility is that both attacks targeted the transition leading to the $u_1$ state of the agents' reward machine, which represents the fact that the button has been pressed and the cookie is present. This observation allows us to deduce how the attack impacted the agents: in the case of \emph{first-stream} and \emph{triggered-stream}, the attacker tried to prevent the agents from receiving the labeling function output $\ao{3B}$. However, recall that those timing strategies conclude the attack after the first consecutive stream of target events is exhausted. Therefore, the victims, to be able to achieve their task in spite of the attack, simply had to press the button once, move away, and press it again. While, the first time, they would not receive the output of the labeling function indicating that the button had been pressed, the second time they could revert to their usual course of action and achieve their task as usual. Indeed, this intuition is supported by the results obtained by the stream-based timing strategies, which achieved minimal tampering rates without, however, leading to any victims' failure. Conversely, in the case of the 
\emph{all-instances} strategy, the attacker prevented its victims from receiving the information on the button, regardless of how many times it was pressed. By doing so, these attacks were able to obtain the best results on the Cookie World so far, inducing a $100\%$ failure rate in its victims with  a tampering rate limited to about $74\%$.

\begin{table}[h]
\centering
\resizebox{\textwidth}{!}{%
\begin{tabular}{@{}rrrrr@{}}
\toprule
\textbf{Attack Type} & \textbf{Timing Strategy} & \textbf{Avg. Tampering Rate} & \textbf{Avg. Failure Rate}     & \textbf{Impact Score}   \\ \midrule
edge        & all           & 63.93\%             & 100.00\%           & 0.4389                               \\ \hd
edge        & first         & 0.19\%              & 66.40\%           & 0.8118                             \\ \hd
edge        & trigger\_30   & 0.13\%              & 66.40\%           & 0.8127                             \\ \hd
edge        & trigger\_40   & 0.14\%              & 66.40\%           & 0.8126                           \\ \hd
edge        & trigger\_50   & 0.14\%              & 66.40\%           & 0.8126                             \\ \HD
state       & all           & 95.69\%             & 100.00\%            & 0.3432                              \\ \hd
state       & first         & 0.28\%              & 66.40\%           & 0.8103                             \\ \hd
state       & trigger\_30   & 0.22\%              & 66.40\%           & 0.8113                         \\ \hd
state       & trigger\_40   & 0.22\%              & 66.40\%           & 0.8113                             \\ \hd
state       & trigger\_50   & 0.24\%              & 66.40\%           & 0.8110                           \\  \hd \bottomrule
\end{tabular}%
}
\caption{Results obtained by the Keys World agent when attacked with both variations of edge-based blinding attacks.}
\label{tab:kw_edgblind}
\end{table}
In the case of the Keys World agents, edge-blinding and state-blinding attacks resorted to two different blinding strategies, which, however, led to similar results. Among the two, the best-performing attack was the edge-based one. When carried out with an \emph{all-instances} timing strategy, the attack was able to completely prevent the victim from achieving its task, while only requiring slightly less than a $70\%$ tampering rate, the lowest value among every testing session for a $100\%$ failure rate attack. However, the highest impact ever was observed when the stream-based timing strategies were used: by tampering with less than $1\%$ of the labeling function outputs, both edge-based and state-based attacks were able to induce a $67\%$ failure rate in the victim. Regardless of their low statistical significance, due to them being related to only one single agent, these results pose an extremely promising path for further investigation that could lead to a deeper comprehension of the optimal strategy to carry out these type of attacks.

\begin{table}[h]
\centering
\resizebox{\textwidth}{!}{%
\begin{tabular}{@{}rrrrrrr@{}}
\toprule
\textbf{Attack Type} & \textbf{Timing Strategy} & \textbf{Avg. Tampering Rate} & \textbf{Avg. Failure Rate} & \textbf{Impact Score}  & \textbf{Avg. Time-to-Failure} & \textbf{Avg Failure Reward} \\ \midrule
edge        & all           & 73.17\%             & 11.57\%           & 0.1380                             & 500                   & 0                  \\ \hd
edge        & first         & 0.62\%              & 0.00\%          & 0.0000                              & -                     & -                  \\ \hd
edge        & trigger\_30   & 0.19\%              & 0.00\%          & 0.0000                              & -                     & -                  \\ \hd
edge        & trigger\_40   & 0.25\%              & 0.00\%          & 0.0000                              & -                     & -                  \\ \hd
edge        & trigger\_50   & 0.31\%              & 0.00\%          & 0.0000                              & -                     & -                  \\ \HD
state       & all           & 73.02\%             & 11.45\%           & 0.1375                             & 500                   & 0                  \\ \hd
state       & first         & 0.61\%              & 0.00\%          & 0.0000                              & -                     & -                  \\ \hd
state       & trigger\_30   & 0.18\%              & 0.00\%          & 0.0000                             & -                     & -                  \\ \hd
state       & trigger\_40   & 0.24\%              & 0.00\%          & 0.0000                              & -                     & -                  \\ \hd
state       & trigger\_50   & 0.31\%              & 0.00\%          & 0.0000                              & -                     & -                  \\ \bottomrule
\end{tabular}%
}
\caption{Results obtained by Symbol World agents when attacked with both variations of edge-based blinding attacks.}
\label{tab:sw_edgblind}
\end{table}
Finally, when the Symbol World agents were targeted, both the edge-based and state-based attacks led to poor results, with the \emph{all-instances} strategy being the only one that could induce any amount of failure rate in the victims. This, in hindsight, is to be expected due to the structure of this domain's reward machine: since these types of attacks target one or more transitions leading to a given RM state, the attacker had no good choice in terms of a single target that could consistently help him in hindering its victims' ability to achieve their task. This is due to the fact that every transition in the agent's reward machine is related to a specific goal among the 9 alternatives the agent can face. Thus, targeting a single transition
and using it to attack the agents, regardless of their actual goal, can only lead, on average, to a maximum decrease in performance of $1/9$, which corresponds precisely to the $11\%$ failure rate induced by the \emph{all-instances} strategies. Moreover, this consideration can also be applied to every type of blinding attack, thus providing an a-posteriori explanation for the results that were obtained by both event-blinding and random-blinding attacks in this environment.

Moving to the RS agents, tables \ref{tab:cwrs_edgblind}, \ref{tab:kwrs_edgblind}, and \ref{tab:swrs_edgblind} present the results that were obtained by subjecting them to both edge-blinding and state blinding-attacks.
\begin{table}[h]
\centering
\resizebox{\textwidth}{!}{%
\begin{tabular}{@{}rrrrr@{}}
\toprule
\textbf{Attack Type} & \textbf{Timing Strategy} & \textbf{Avg. Tampering Rate} & \textbf{Avg. Failure Rate}     & \textbf{Impact Score}   \\ \midrule
edge/state        & all           & 93.00\%             & 100.00\%            & 0.3497                           \\ \hd
edge/state        & first         & 2.71\%              & 0.00\%          & 0.0000                              \\ \hd
edge/state        & trigger\_30   & 0.84\%              & 0.00\%          & 0.0000                             \\ \hd
edge/state        & trigger\_40   & 1.12\%              & 0.00\%          & 0.0000                            \\ \hd
edge/state        & trigger\_50   & 1.39\%              & 0.00\%          & 0.0000                            \\ \bottomrule
\end{tabular}%
}
\caption{Results obtained by Cookie World RS agents when attacked with both variations of edge-based blinding attacks.}
\label{tab:cwrs_edgblind}
\end{table}
The RS agents for the Cookie World environment once again confirmed the increased robustness arising from the use of reward shaping: although the \emph{all-instances} timing strategy was still able to nullify the victim's performance completely, in the case of RS agents, this required a $20\%$ higher tampering rate. Similarly, the increase in robustness was even higher for Keys World RS agents, which suffered from neither the $100\%$ failure rate induced by the \emph{all-instances} strategy in the regular agents nor the high-impact attacks requiring less than a $1\%$ tampering rate.
Finally, the Symbol World RS agents showed the exact same level of robustness as the one observed in regular agents, which can be again motivated by the same considerations above.

\begin{table}[h]
\centering
\resizebox{\textwidth}{!}{%
\begin{tabular}{@{}rrrrr@{}}
\toprule
\textbf{Attack Type} & \textbf{Timing Strategy} & \textbf{Avg. Tampering Rate} & \textbf{Avg. Failure Rate}     & \textbf{Impact Score}   \\ \midrule
edge/state        & all           & 85.91\%             & 57.29\%           & 0.2785                             \\ \hd
edge/state        & first         & 0.82\%              & 4.34\%           & 0.2050                              \\ \hd
edge/state        & trigger\_30   & 0.52\%              & 1.46\%           & 0.1196                              \\ \hd
edge/state        & trigger\_40   & 0.59\%              & 1.96\%           & 0.1384                              \\ \hd
edge/state        & trigger\_50   & 0.64\%              & 2.33\%           & 0.1507                              \\ \bottomrule
\end{tabular}%
}
\caption{Results obtained by Keys World RS agents when attacked with both variations of edge-based blinding attacks.}
\label{tab:kwrs_edgblind}
\end{table}

\begin{table}[h]
\centering
\resizebox{\textwidth}{!}{%
\begin{tabular}{@{}rrrrrrr@{}}
\toprule
\textbf{Attack Type} & \textbf{Timing Strategy} & \textbf{Avg. Tampering Rate} & \textbf{Avg. Failure Rate} & \textbf{Impact Score}  & \textbf{Avg. Time-to-Failure} & \textbf{Avg Failure Reward} \\ \midrule
edge/state        & all           & 76.93\%             & 11.45\%          & 0.1333                             & 500                   & 0                  \\ \hd
edge/state        & first         & 0.61\%              & 0.00\%          & 0.0000                              & -                     & -                  \\ \hd
edge/state        & trigger\_30   & 0.18\%              & 0.00\%          & 0.0000                              & -                     & -                  \\ \hd
edge/state        & trigger\_40   & 0.24\%              & 0.00\%          & 0.0000                              & -                     & -                  \\ \hd
edge/state        & trigger\_50   & 0.31\%              & 0.00\%          & 0.0000                              & -                     & -                  \\ \bottomrule
\end{tabular}%
}
\caption{Results obtained by Symbol World RS agents when attacked with both variations of edge-based blinding attacks.}
\label{tab:swrs_edgblind}
\end{table}
\chapter{Conclusions}
In my thesis, I presented the first investigation of the security implications of reward machine-based reinforcement learning. Starting with the development of a novel class of attacks on DQRM agents, \textbf{blinding attacks}, I conducted an exhaustive experimental evaluation of their effectiveness for hindering the performance of optimally trained agents in three partially observable benchmark domains. In light of the results that I obtained, these attacks showed the potential for posing significant concern to any deployment of RM-based agents. 

The key issue that currently limits their applicability to real-world scenarios is related to the amount of tampering the attacker is required to carry out on the agent's labeling function to induce a high probability of agent failure. Despite the fact that this property highly depends on both the environment and the agent at hand, future work could focus on the definition of more sophisticated blinding and timing strategies, allowing for a more careful tampering process that could reduce the overall detectability of the attacks. 

Another promising direction for further investigations is represented by \textbf{hallucination attacks}, for which I only provided a definition and a limited set of results based on a random noise approach. Specifically, further work could be aimed at the development of techniques that allow for the dynamic computation of promising attack paths under a set of heuristic constraints over, for instance, the expected number of required tamperings, and probability of success. 

Finally, the study of white-box variations over the basic blinding and hallucination attack frameworks could be of great value by allowing for a deeper understanding of both the theoretical and practical conditions that need to be satisfied for the success of the attacks to be guaranteed.

\printbibliography

\end{document}